\documentclass{ieeeaccess}
\usepackage{cite}
\usepackage{amsmath,amssymb,amsfonts}
\usepackage{algorithmic}
\usepackage{graphicx}
\usepackage{textcomp}
\usepackage{subcaption}
\usepackage{rotating}
\usepackage{pdflscape}
\usepackage{tablefootnote}
\usepackage{threeparttable}
\usepackage{multirow}
\usepackage{etoolbox}
\usepackage{url} 
\usepackage{amssymb}% http://ctan.org/pkg/amssymb
\usepackage{pifont}% http://ctan.org/pkg/pifont
\newcommand{\cmark}{\ding{51}}%
\newcommand{\xmark}{\ding{55}}%

\usepackage[shortlabels]{enumitem}
\def\thevol{}
\def\myyear{2020}
\makeatletter
\patchcmd{\@oddfoot}{2020}{\myyear}{}{}
\patchcmd{\@evenfoot}{2020}{\myyear}{}{}

\makeatother
\def\BibTeX{{\rm B\kern-.05em{\sc i\kern-.025em b}\kern-.08em
        T\kern-.1667em\lower.7ex\hbox{E}\kern-.125emX}}

\begin{document}

\markboth
{Rezk \headeretal: Recurrent Neural Networks: An Embedded Computing Perspective}
{Rezk \headeretal: Recurrent Neural Networks: An Embedded Computing Perspective}

\history{Accepted for publication in IEEE open Access}
\doi{10.1109/ACCESS.2020.2982416}

\title{Recurrent Neural Networks: An Embedded Computing Perspective}

\author{\uppercase{Nesma M. Rezk}\authorrefmark{1}, 
\uppercase{Madhura Purnaprajna\authorrefmark{2},Tomas Nordström\authorrefmark{3}, and Zain Ul-Abdin\authorrefmark{1}}  
}
\address[1]{School of information technology, Halmstad University, Sweden (e-mail: nesma.rezk,zain-ul-abdin@hh.se)}
\address[2]{Amrita Vishwa Vidyapeetham, Bangalore, India (e-mail: p\_madhura@blr.amrita.edu)}
\address[3]{Umeå University (e-mail:tomas.nordstrom@umu.se)}

\tfootnote{This research is performed in the NGES (Towards Next Generation Embedded Systems: Utilizing Parallelism and Reconfigurability) Indo-Swedish project, funded by VINNOVA Strategic Innovation grant and the Department of Science and Technology (INT/SWD/VINN/p-10/2015), Government of India.}

\corresp{Corresponding author: Nesma M. Rezk (e-mail: nesma.rezk@hh.se).}

\begin{abstract}
Recurrent Neural Networks (RNNs) are a class of machine learning algorithms used for applications with time-series and sequential data. Recently, there has been a strong interest in executing RNNs on embedded devices. However, difficulties have arisen because RNN requires high computational capability and a large memory space. In this paper, we review existing implementations of RNN models on embedded platforms and discuss the methods adopted to overcome the limitations of embedded systems.

We will define the objectives of mapping RNN algorithms on embedded platforms and the challenges facing their realization. Then, we explain the components of RNN models from an implementation perspective. We also discuss the optimizations applied to RNNs to run efficiently on embedded platforms. Finally, we compare the defined objectives with the implementations and highlight some open research questions and aspects currently not addressed for embedded RNNs.

Overall, applying algorithmic optimizations to RNN models and decreasing the memory access overhead is vital to obtain high efficiency. To further increase the implementation efficiency, we point up the more promising optimizations that could be applied in future research. Additionally, this article observes that high performance has been targeted by many implementations, while flexibility has, as yet, been attempted less often. Thus, the article provides some guidelines for RNN hardware designers to support flexibility in a better manner.  
\end{abstract}


\titlepgskip=-15pt

\maketitle

\section{Introduction}
%\todo[inline]{Review}

% RNN is important... should run on embedded ... In this paper, we ......
Recurrent Neural Networks (RNNs) are a class of Neural Networks (NNs) dealing with applications that have sequential data inputs or outputs. RNNs capture the temporal relationship between input/output sequences by introducing feedback to FeedForward (FF) neural networks. Thus, many applications with sequential data such as speech recognition \cite{rnn-speech}, language translation \cite{rnn-language}, and human activity recognition \cite{rnn-activity} can benefit from RNNs.

In contrast to cloud computing, edge computing can guarantee better response time and enhance security for the running application. Augmenting edge devices with RNNs grant them the intelligence to process and respond to sequential problems. Realization on embedded platforms in edge devices imposes some optimizations to RNN applications. Embedded platforms are time-constrained systems that suffer from limited memory and power resources. To run RNN applications efficiently on embedded platforms, RNN applications need to overcome these restrictions.

% Related Surveys

\begin{figure*}[h]
   \centering
   \includegraphics[width=2\columnwidth]{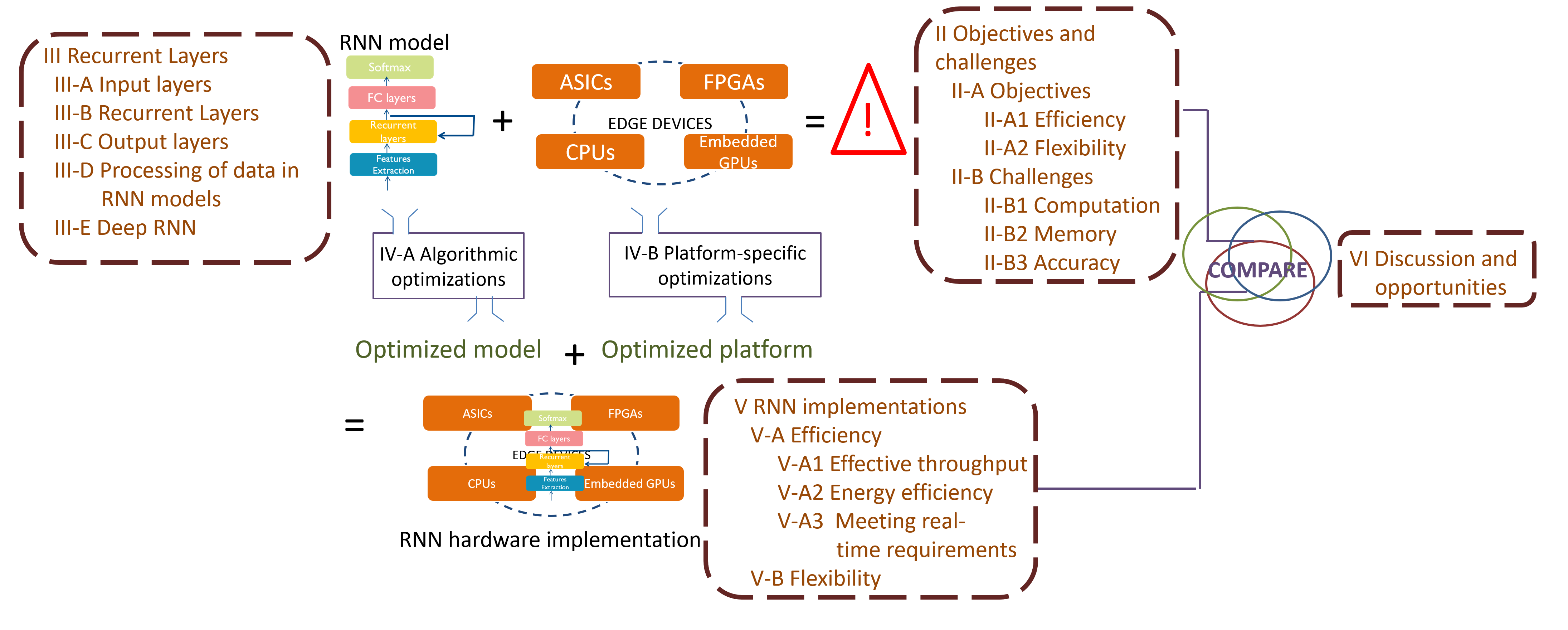}
    \caption{Structure of the survey article. RNN models should run on an embedded platform in an edge device. Section \ref{Sec:Challanges} discusses the objectives of such an implementation and the challenges facing it. Section \ref{Sec:RNN} describes the RNN models in detail. There follows a discussion of how algorithmic optimizations (Section \ref{sec:algsection}) may be applied to RNN models and how platform-specific optimizations (Section \ref{subsec:hw-optimize}) are applied to embedded platforms. The resulting implementations are discussed in Section \ref{Sec:RNNonHW} and compared to the objectives in Section \ref{Sec:Discussion}. }
    \label{fig:structure}
\end{figure*}
\subsection{Scope of the article}
 In this article, we study RNN models and specifically focus on RNN optimizations and implementations on embedded platforms. %This survey article focuses on embedded solutions for RNN models.
 The article compares recent implementations of RNN models on embedded systems found in the literature. For a research paper to be included in the comparison, it should satisfy the following conditions:
\begin{itemize}
    \item It discusses the implementation of an RNN model or the recurrent layer of an RNN model.
    \item The target platform is an embedded platform such as FPGA, ASIC, etc.
    
\end{itemize} 
To provide a complete study, the survey also addresses the methods used for optimizing the RNN models and realizing them on embedded systems.
%\subsection{Related Work}

This survey distinguishes itself from related works because no existing article includes the components of the RNN models with optimizations and implementations in a single analysis, as may be seen in Table~\ref{table:related}.
The other surveys focus on one or two aspects compared to those covered in this article. Some articles study RNNs from an algorithmic point of view~\cite{rnn-survey,RNN-survey-lipton}. While another group of survey articles looks at the hardware implementations. For example, one survey on neural networks efficient processing~\cite{CNN-survey} studied CNNs, CNN optimizations, and CNN implementations, while another CNN survey~\cite{survey-CNN-FPGA} studied CNN mappings on FPGAs. Some articles were specialized in algorithmic optimizations such as compression\cite{survey-compression}. All algorithmic optimizations for both CNNs and RNNs were surveyed in one article that also discussed their implementations~\cite{Survey-ACM-optimizations}. However, the main scope of the article was optimizations, and so RNN models and their components were not studied. Furthermore, the RNN implementations included were limited to speech recognition applications on the TIDIGITs dataset. 
\begin{table*}[htpb]
\centering
\begin{threeparttable}
\caption{Comparison with related survey articles.}

%\begin{tabular}{|p{2cm}||p{0.75cm}|p{0.75cm}|p{0.75cm}|p{1cm}|p{0.75cm}|}
\begin{tabular}{|c||c|c|c|c|c|c|}
\hline
Article &\cite{rnn-survey} \& \cite{RNN-survey-lipton}&\cite{survey-compression}&\cite{Survey-ACM-optimizations}&\cite{CNN-survey}&\cite{survey-CNN-FPGA}&This article \\ \hline

Analysis of RNN models components  &\hfil{\cmark}&\hfil{\xmark}&\hfil{\xmark}& \hfil{\xmark}&\hfil{\xmark}&\hfil{\cmark}\\ \hline

Analysis of CNN models components 
&\hfil{\xmark}&\hfil{\xmark}&\hfil{\xmark}& \hfil{\cmark}&\hfil{\xmark}&\hfil{\xmark}\\ \hline

Algorithmic Optimizations  &\hfil{\xmark}&\hfil{\cmark}&\hfil{\cmark}&\hfil{\cmark}&\hfil{\cmark}& \hfil{\cmark}\\ \hline

Platform-specific optimizations  &\hfil{\xmark}&\hfil{\xmark}&\hfil{\xmark}& \hfil{\cmark}&\hfil{\cmark}&\hfil{\cmark} \\ \hline

RNN Implementations  &\hfil{\xmark}&\hfil{\xmark}&Speech recognition only&\hfil{\xmark}&\hfil{\xmark}&\hfil{\cmark} \\ \hline

CNN Implementations  &\hfil{\xmark}&\hfil{\xmark}&\hfil{\cmark}&\hfil{\cmark}&FPGAs only&\hfil{\xmark} \\ \hline

\end{tabular}

\label{table:related}%
\end{threeparttable}
\end{table*}

\subsection{Contributions}
 %To the best of our knowledge, this is the first survey article that focuses on RNNs implementations on embedded systems.
 This survey article provides the following:
\begin{itemize}
    
    \item A detailed comparison of RNN models' components from a computer architecture perspective that addresses computational and memory requirements.
    
    \item A study of the optimizations applied to RNNs to execute them on embedded platforms.
    
    \item An application-independent comparison of recent implementations of RNNs on embedded platforms.
    
    \item Identification of possible opportunities for future research.

\end{itemize}
\subsection{Survey Structure}
This survey article is organized as shown in Figure~\ref{fig:structure}. Section~\ref{Sec:Challanges} defines the objectives of realizing RNN models on embedded platforms and the challenges faced in achieving them. 
We then define a general model for RNN applications and discuss different variations for the recurrent layers in RNN models in Section~\ref{Sec:RNN}. However, it is difficult to run RNN models in their original form efficiently on embedded platforms. Therefore, researchers have applied optimizations to both the RNN model and the target platform. The optimizations applied to the RNN model are called algorithmic optimizations and are discussed in Section~\ref{sec:algsection}; the optimizations applied to the hardware platform are called platform-specific optimizations and are discussed in Section~\ref{subsec:hw-optimize}. Then, in Section~\ref{Sec:RNNonHW}, we present an analysis of the hardware implementations of RNNs suggested in the literature. The implementations are compared against the applied optimizations and their achieved performance. In Section~\ref{Sec:Discussion}, we compare the implementations analyzed in Section~\ref{Sec:RNNonHW} with the objectives defined in Section~\ref{Sec:Challanges} to define the gap between them and propose research opportunities to fill this gap. Finally, in Section~\ref{Sec:Conclusions}, we summarize our survey.%--------------------
\section{Objectives and Challenges}
\label{Sec:Challanges}

Implementation efficiency is the primary objective in implementing RNN applications on embedded systems. Implementation efficiency requires the implementation to have high throughput, low energy consumption, and to meet real-time requirements. A secondary objective for the implementation would be flexibility. Flexibility requires the implementation to support variations in the RNN model, to allow for online training, and to meet different applications requirements. In meeting these objectives, there exist challenges in mapping these applications onto embedded systems, such as the large number of computations to be performed within the limited available memory. 
These objectives and challenges are discussed in detail below.

\subsection{\textbf{Objectives of realizing RNNs on embedded platforms}}
\label{subsec:objectives}
To realize RNN models on embedded platforms, we define some objectives that will influence the solution. These objectives are divided into implementation efficiency objectives and flexibility objectives. 

\subsubsection{\textbf{Implementation Efficiency}}

Since we target embedded platforms, we consider the online execution of the application. To satisfy the implementation efficiency objective, the implementation should have a high throughput, low energy consumption, and meet the real-time requirements of the application. The real-time requirements of the application pose additional demands for the throughput, energy consumption and the accuracy of the implementation. Accuracy indicates how correct the model is in performing recognition, classification, translation, etc. 
\begin{itemize}
    \item \textbf{High throughput}
    Throughput is a measure of performance. It measures the number of processed input/output samples per second. Application-level inputs and outputs are diverse. For image processing applications, the input can be frames and the throughput can be the number of consumed frames per second, which may also depend on the frame size. For speech/text applications, it can be the number of predicted words per second. Thus for different sizes and types of input and outputs, throughput can have different units and the throughput value may be interpreted in various ways. To compare different applications, we use the number of operations per second as a measure of throughput.  
    
    \item \textbf{Low energy consumption}
    For an implementation to be considered efficient, the energy consumption of the implementation should meet embedded platforms' energy constraints. To compare the energy consumption of different implementations, we use the number of operations per second per watt as a measure of energy efficiency.  
    %The energy constraints are defined by the application as well and should not exceed a predefined limit. The energy limit can differ from one application to the other. For instance, accepted consumed energy by a wearable device is much less than accepted consumed energy by a vehicle~\cite{energy-wearable,autonomous-performance}.
    
    \item \textbf{Real-time requirements}
    In real-time implementations, a response cannot be delayed beyond a predefined deadline, and energy consumption cannot exceed a predefined limit. The deadline is defined by the application and is affected by the frequency of sensor inputs and the system response time. Normally, the RNN execution should meet the predefined deadline.% The power limit is defined by the application. Power limit for an application running in a car is different if the same application is running on a wearable device.

    %\item meeting runt-time deadline
\end{itemize}

%at runtime, it is not accepted for the response to be delayed or the throughput to be under a predefined threshold.
%Nevertheless, the power consumption of the implementation should be low to meet embedded platforms energy constraints.

\subsubsection{\textbf{Flexibility}}
The flexibility of the solution in this context is the ability of the solution to run different models under different constraints without being restricted to one model or one configuration. For an implementation to be flexible, we define the following requirements that should be satisfied:
\begin{itemize}
    \item {\textbf{Supporting variations in RNN layer}}
The recurrent layers of RNN models can vary in the type of the layer (different types of the recurrent layer are discussed in Section~\ref{subsec:RNN-types}), the number of hidden cells, and the number of recurrent layers.% The number of recurrent layers can define the depth of the RNN model, where a model with stacked recurrent layers can be considered as a deep RNN model \cite{RNN-Deep-stacked,deep-bengio}.
 \item {\textbf{Supporting other NN layers}}
RNN models have other types of NN layers as well. A solution that supports more NN layers is considered a complete solution for RNN models, and not just a flexible solution. Convolution layers, fully connected layers, and pooling layers might be required in an RNN model. 

\item{\textbf{Supporting algorithmic optimization variations}}
Different algorithmic optimizations are applied to RNN models to implement them efficiently on embedded systems (Section~\ref{Sec:Optimizations}). Supporting at least one algorithmic optimization for the hardware solution is in many cases mandatory for a feasible execution of RNN models on an embedded system. Combinations of optimizations will lead to higher efficiency and flexibility as this gives the algorithmic designer more choices while optimizing the model for embedded execution. 

\item{\textbf{Online training}} 
Training is a process that sets parameter values within the neural network. In embedded platforms, training is performed offline, and only inference is run on the platform at run-time. For real-life problems, it is often not enough to run only inference on the embedded platforms -- some level of training is required at run-time as well. Online training allows the neural network to adapt to new data that was not encountered within the training data, and to adapt to changes in the environment. For example, online training is required for object recognition in autonomous cars to achieve lifelong learning, by continuously receiving new training data from fleets of robots and updating the model parameters~\cite{online-tracking}. Another example is in automated visual monitoring systems that continuously receive new labeled data \cite{online-finetuning}.

\item{\textbf{Meeting the requirements of different application domains}}
One aspect of flexibility is to support the requirements of different application domains. This makes the implementation attractive because the solution can support a wider range of applications. However, different application domains can have different performance criteria. Some application domains, such as autonomous vehicles~\cite{autonomous-performance}, might require very high throughput with moderate power consumption, while others, such as mobile applications~\cite{HW-mobile-deepx,survey-mobile}, require extremely low power consumption but have less stringent constraints on throughput. %For instance, in autonomous vehicles, the latency should be in nanoseconds and the power consumption does not exceed 30 watts~\cite{autonomous-performance}. On the other hand, in mobile devices, latency can be in milliseconds, while power consumption should be in milli-watts or even micro-watts~\cite{HW-mobile-deepx,survey-mobile}. In addition, different applications have different accepted accuracy ranges. For instance, if the model can be changed to enhance its throughput, accuracy loss might happen. This loss can be accepted in some applications and rejected in others.
%Being able to make changes in the solution to meet these different performance metrics would help increase the flexibility of the solution to support different application domains.
\end{itemize}

\subsection{\textbf{Challenges in mapping RNNs on embedded platforms}}
We shall now take a look at the challenges faced by hardware solutions to meet the objectives discussed above.
\subsubsection{\textbf{Computation challenges }}

The main computation bottleneck in RNNs is the matrix to vector multiplications. The LSTM layer (Explained in detail in Section~\ref{subsec:RNN-types}) has four computation blocks, each of which has one matrix to vector multiplication. For example, if the size of the vector is 1280 and the size of the matrices is 1280 $\times$ 1024, each matrix to vector multiplication requires $1280 \times 1024$ MAC (Multiply And Accumulate) operations. The total number of MAC operations in the LSTM would be $4 \times 1280 \times 1024 = 5.24$ Mega MAC, which is approximately equivalent to 10.5 MOP. The high number of computations negatively affects both the throughput of the implementation and energy consumption.

One other problem in RNNs is the recurrent structure of the RNN. In RNNs, the output is fed back as an input in such a way that each time-step computation needs to wait for the previous time-step computation to complete. This temporal dependency makes it difficult to parallelize the implementation over time-steps. 

\subsubsection{\textbf{Memory challenges}}
The memory required for the matrix to vector multiplications can be very large. The size and the access time of these matrices become a memory bottleneck. The previous example of the LSTM layer, requires four matrices, each of size $1280 \times 1024$. Consider 32-bit floating-point operations: the size of the required memory for the weights would be $32 \times 4 \times 1280 \times 1024 = 21 MB$. Also, the high number of memory accesses affects the throughput and energy consumption of the implementation~\cite{Chen:2016:ESA:3001136.3001177}.

\subsubsection{\textbf{Accuracy challenges}}

To overcome the previous two issues (computation and memory challenges), optimizations can be applied to RNN models as discussed in Section \ref{Sec:Optimizations}. These optimizations may affect accuracy. The acceptable decrease in accuracy varies with the application domain. For instance, in aircraft anomaly detection, the accepted range of data fluctuation is only $5\%$ \cite{HW-FPGA-flight}.% and i, it might be difficult to apply optimizations and keep accuracy high.

%In this section, we have discussed the objectives and challenges of mapping RNNs on embedded platforms. In the next section, we explain the RNN model components and variations and we discuss the application domains studied in the literature.

%--------------------
\section{ Recurrent Neural Networks}
\label{Sec:RNN}
\label{Sec:RNNs}
 The intelligence of humans, as well as most animals, depends on having a memory of the past. This can be short-term, as when combining sounds to make words, and long-term, for example where the word ``she'' can refer back to ``Anne'' mentioned hundreds of words earlier. This is exactly what RNN provides in neural networks. It adds feedback that enables using the outputs of previous time step while processing the current time-step input. It aims to add memory cells that function similarly to human long-term and short-term memories.

\begin{figure}[h]
   \centering
   \includegraphics[width=1\columnwidth]{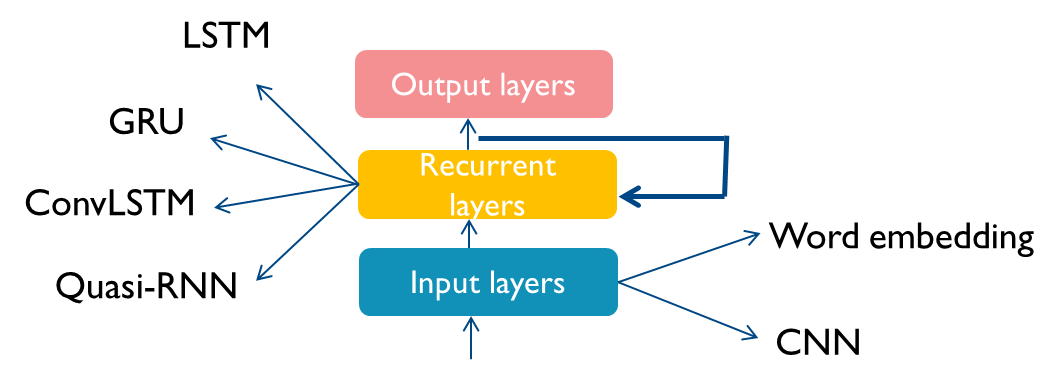}
    \caption{Generic model of RNNs with diverse recurrent layers.}
    \label{fig:general}
\end{figure}

RNNs add recurrent layers to the NN (Neural Network) model. Figure~\ref{fig:general} presents a generic model for RNNs that consists of three sets of layers (input, recurrent, and output). Input layers take the sensor output and convert it into a vector that conveys the features of the input. These are followed by the recurrent layers, which provide feedback. In most recent recurrent layer models, memory cells exist as well. Subsequently, the model completes similarly to most NN models with Fully Connected (FC) layers and an output layer that can be a softmax layer. FC layers and the output layer are grouped into the set of output layers in Figure~\ref{fig:general}. In this section, we discuss the input layers, different types of recurrent layer, output layers, RNN modes of operation, deep RNN, and RNN applications and their corresponding datasets. 

 %After explaining the RNNs models components, we discuss the RNN models variations through time-steps and explain how an RNN can be a Deep RNN. Finally, we state the applications covered by the implementations studied in this survey and their corresponding datasets.

%\subsection{\textbf{Input layers}}
%\label{subsec:RNN-features}
%\section{Feature Extractor}
%\input{source/features.tex}
\subsection{\textbf{Input layers (features extractor)} and corresponding applications and datasets}
\label{app:background}
%\section{Feature Extractor}
% Section on input layers
Input layers are needed by many implementations to prepare the sensor output for processing (these may also called feature extraction layers). 
Often, the raw sensor data, e.g., the audio samples or video frames, are in a form that is unsuitable for direct processing in the recurrent layer. Also, the RNN performance (in learning rate and accuracy) can be significantly improved if suitable features are extracted in the input layer. 

As sensor types (and numbers) change with the application, RNN models show a large variation with application types as well. Thus it is important to study which applications an RNN model is used for and their corresponding datasets. Datasets are used by researchers to demonstrate success in applying their methods and the modifications to them. Datasets differ in the size of the data samples, the values of data samples, and the total size of the dataset. The success of
NN models is measured by accuracy. Accuracy indicates how
correct the model is when carrying out recognition, classification, translation, etc. 

In this section, we discuss examples from three application domains where input layer pre-processing is used: audio, video, and text. In Table 2, we summarize these application domains and their corresponding datasets. For different datasets, different metrics are used to assess accuracy. 

\subsubsection{\textbf{Audio inputs}}
Audio feature extractors translate sound signals into feature vectors. In speech processing, we often want to extract a frequency content from the audio signal (in a similar way to the human ear)~\cite{sejdic2009time}. 
There are many ways to do this, for example, by using short-time Fourier transform (STFT), mel frequency cepstral coefficients (MFCC) and linear predictive coding (LPC) coefficients~\cite{gupta2016analysis}.

\textbf{Applications: Speech recognition}

 Speech recognition applications receive audio as input, understand it, and translate it into words. Speech recognition can be used for phonetic recognition, voice search, conversational speech recognition, and speech-to-text processing~\cite{RNN-DL-survey}.  

\subsubsection{\textbf{Video inputs}}
When the input is a video signal, that is, a sequence of images or frames, it is natural to use a convolutional neural network (CNN) as an input layer. CNN layers then extract image features from each video frame and feed the resulting feature vector to the recurrent layer. 
This use of a CNN as an input layer before a recurrent layer has been employed for many applications with video inputs, such as activity recognition, image description~\cite {rnn-activity,RNN-image-captioning}, or video description~\cite{HW-FPGA-conv-LSTM-short}.

The use of CNN as an input layer can also be found for audio signals~\cite{xu2017convolutional}. 
In this case, a short segment of audio samples is transformed into a frequency domain vector using, for example, STFT or MFCC. By combining a number of these segments into a spectrogram, we can show information about the source's frequency and amplitude against time. This visual representation is then fed into a CNN as an image. The CNN then extracts speech or audio features suitable for the recurrent layer.

\textbf{Applications: Image/Video applications}

Image/video applications cover any application that takes images as input, for example, image captioning, activity recognition, and video description.
 
% QUESTION... What embedded system will get text as input??
% Maybe  Mobile applications?! Text to speech for special needs?
\subsubsection{\textbf{Text inputs}}
When the input is in the form of text, we often want to represent words as vectors, and word embedding is one common way to do this~\cite{RNN-NLP}. The word embedding layer extracts the features of each word in relation to the rest of the vocabulary. 
The output of the word embedding is a vector. For two words with similar contexts, the distance between their two vectors is short, while it is large for two words that have different contexts.

Following word embedding in an input layer, deeper text analysis or natural language processing is performed in the recurrent layers.
%
%One example is sentiment analysis (or emotional AI) that captures the feelings behind the text and words  \cite{RNN-sentiment}.

%\subsection{\textbf{Applications and their corresponding datasets}}
%The purpose of this survey paper is to study the implementations of RNN models on embedded platforms. Thus, we try to cover the relevant applications and datasets in the literature only, rather than providing a broad coverage of all applications and datasets for RNN models.  

\textbf{Applications:}
\begin{itemize}
 \item {\textbf{Text generation}}

RNN models can be used for language-related applications such as text generation. RNN models can predict the next words in a phrase, using the previous words as inputs.

\item {\textbf{Sentiment analysis}}

Sentiment analysis is the task of understanding the underlying opinion expressed by words~\cite{sentiment,RNN-sentiment}. Since the input words comprise a sequence, RNN methods are well-suited to performing sentiment analysis. 

\end{itemize}

\begin{table*}[tb]
\centering
\begin{threeparttable}
\caption{RNN input layer types and their corresponding application domains and datasets.}
\begin{tabular}{|p{3cm}|p{3cm}||p{4.2cm}|p{5cm}|}
\hline

Input type &Applications &Dataset&Accuracy measure metric \\ 
\hline \hline
% Fixed point
%94.2,95.3 vs 94
% accuracy got from higher point in curve
\multirow{5}{*}{Audio input} &\multirow{5}{*}{Speech recognition}&TIDIGITS \cite{dataset-tidigits}&\multirow{5}{*}{\parbox{5cm}{Word Error Rate (WER) (Lower is better) \& Phone Error Rate (PER) (Lower is better) }}  %2018
 \\\cline{3-3}  

&&AN4 \cite{dataset-AN4}%WER (Lower is better)
 &\\\cline{3-3}

&&TIMIT \cite{dataset-timit}%WER (Lower is better)
&\\ \cline{3-3} 

&&\text{Wall Street} \text{Journal(WSJ)} \cite{dataset-wsj}%WER (Lower is better)
 &\\ \cline{3-3} 

&&\text{LibriSpeech ASR corpus~\cite{dataset-librispeech}}%WER (Lower is better)
&\\ \hline

\multirow{3}{*}{Video input}&\multirow{3}{*}{\parbox{3cm}{Image/video applications}} &COCO \cite{dataset-coco}&BLEU (Higher is better) \\ \cline{3-4}

&&Moving MNIST \cite{dataset-moving-mnist}&Cross entropy loss (Lower is better) \\ \cline{3-4}

&&comma.ai driving dataset \cite{dataset-comma-driving}&RMS prediction error (Lower is better)\\ \hline

\multirow{5}{*}{Text input}&\multirow{4}{*}{Text generation}&\text{Penn Treebank (PTB) \cite{dataset-ptb}}&\multirow{4}{*}{}Perplexity per word (PPW)%2018
\\ \cline{3-3}

&&wikitext \cite{dataset-wikitext}&(Lower is better)
\\ \cline{3-3}

&&Text8 \cite{dataset-text8} & \& Bilingual Evaluation Understudy (BLEU)
\\ \cline{3-3}

&&WMT'14 \cite{dataset-WMT14}&(Higher is better)
\\ \cline{2-3}

&Sentiment analysis & IMDB \cite{dataset-IMDB}& Testing accuracy (Higher is better) \\\hline

%Music generation &Nottingham~\cite{dataset-music}& Testing accuracy (Higher is better) \\ \hline
\end{tabular}
\label{tab:datasets}%
\end{threeparttable}
\end{table*}	
%\tablefootnote{*Accuracy is also affected by the compression scheme used in this work.}

%\paragraph{Model Example}:
%\end{itemize}
\subsection{\textbf{Recurrent layers}}
\label{subsec:RNN-types}
In this section, we cover the various types of recurrent layers. For each layer, we discuss the structure of the layer and the gate equations. The most popular recurrent layer is the Long Short Term Memory (LSTM)~\cite{RNN-LSTM}. Changes have been proposed to the LSTM to enhance algorithmic efficiency or improve computational complexity. Enhancing algorithmic efficiency means improving the accuracy achieved by the RNN model, which includes LSTM with peepholes and ConvLSTM, as discussed in Sections~\ref{subsub:lstmpeep} and~\ref{subsub:convlstm}. Improving computational complexity means reducing the number of computations and the amount of memory required by an LSTM to run efficiently on a hardware platform.  Techniques include LSTM with projection, GRU, and QRNN/SRU, which are discussed in Sections~\ref{subsub:lstmproj},~\ref{subsub:gru}, and~\ref{subsub:qrnnsru}, respectively. These changes can be applied to the gate equations, interconnections, or even the number of gates. 
Finally, we compare all the different layers against the number of operations and the number of parameters in Table~\ref{tab:complstm}.
\subsubsection{\textbf{LSTM}}

\begin{figure*}[] 
  \begin{subfigure}[b]{0.5\linewidth}
    \centering
    \includegraphics[width=0.75\linewidth]{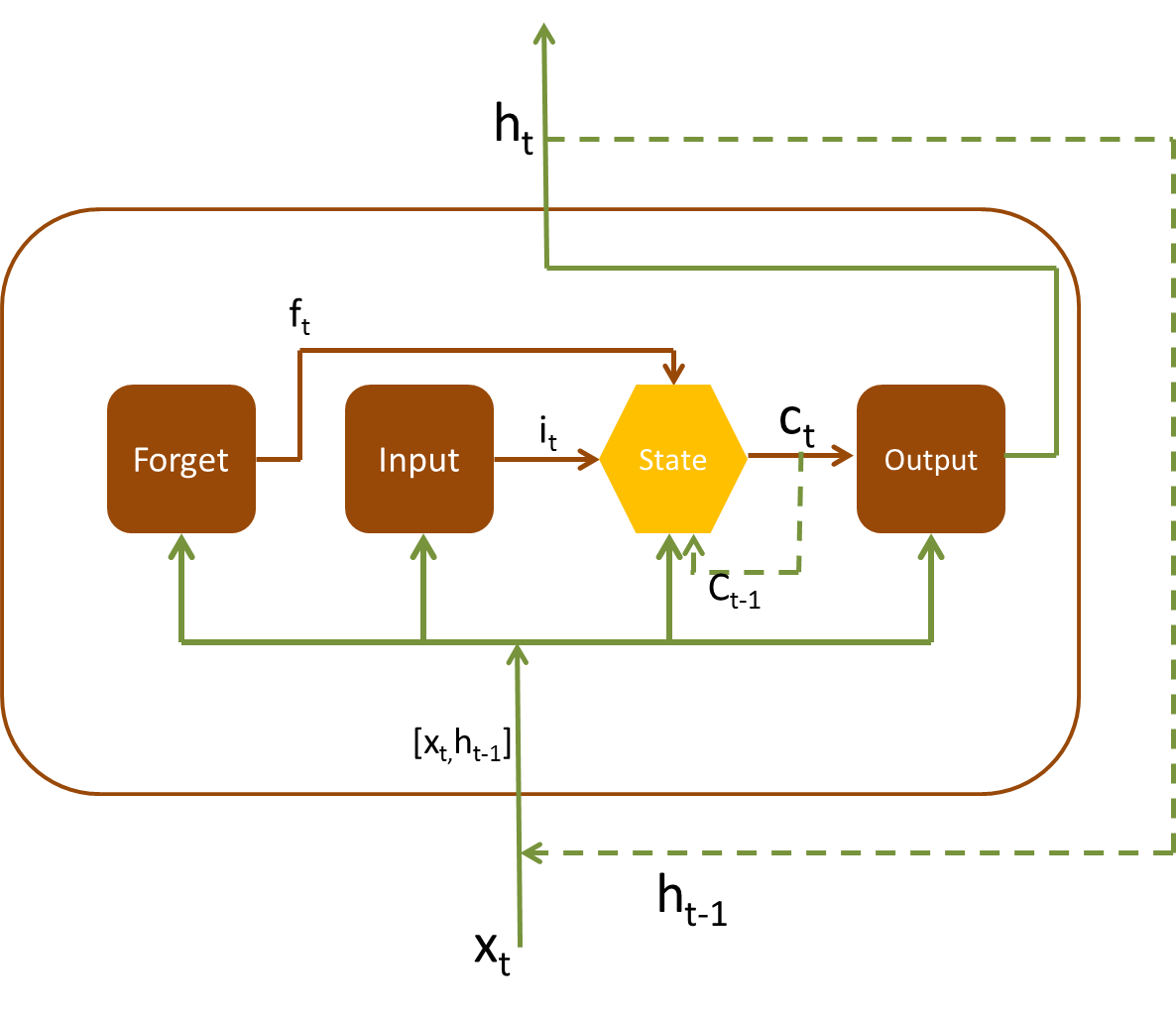} 
    \caption{Long Short Term Memory (LSTM).}
    \label{fig:lstm} 
    \vspace{4ex}
  \end{subfigure}%% 
  \begin{subfigure}[b]{0.5\linewidth}
    \centering
    \includegraphics[width=0.75\linewidth]{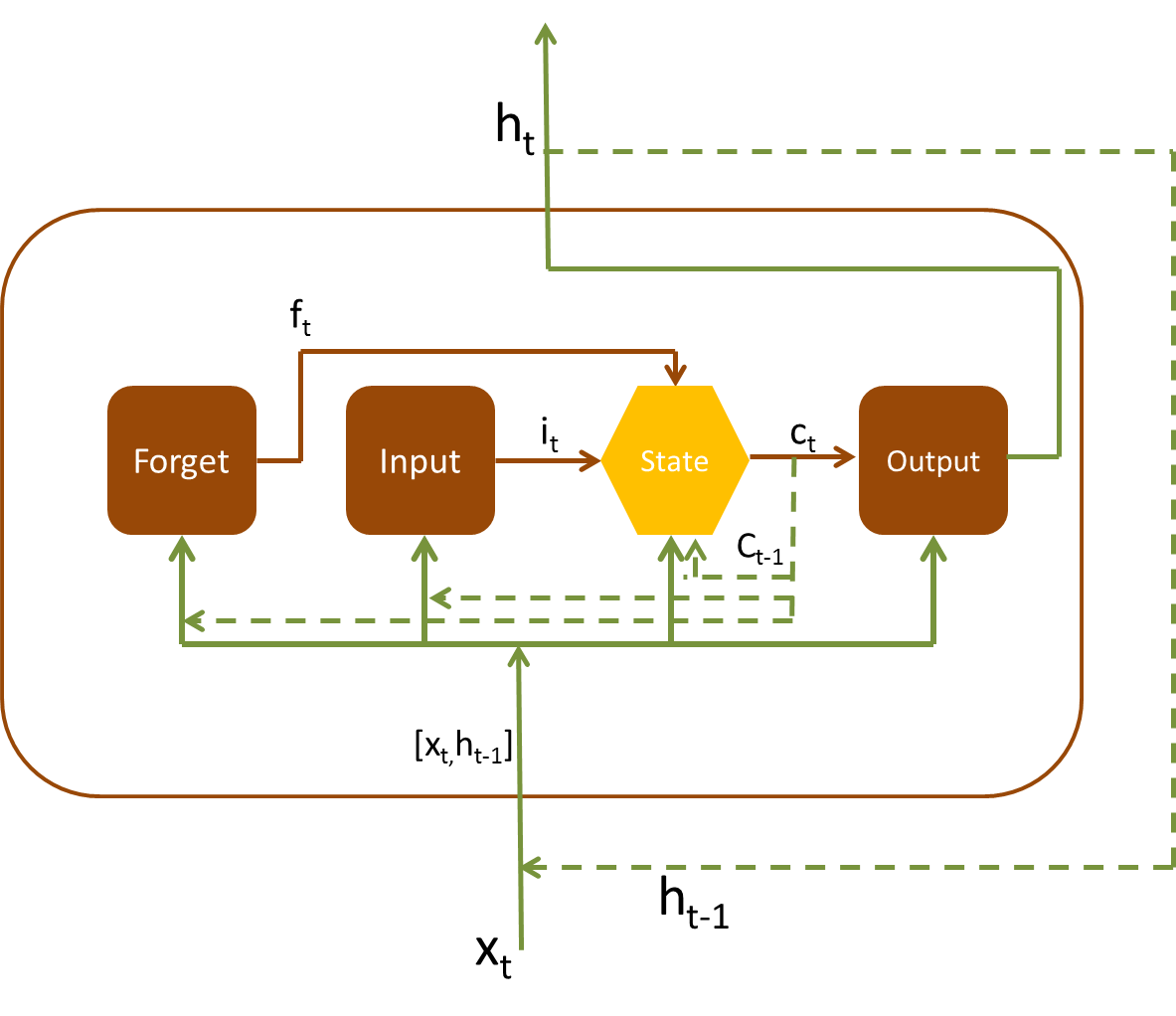} 
   \caption{LSTM with peepholes.}
    \label{fig:lstmpeep}
    \vspace{4ex}
  \end{subfigure} 
  \begin{subfigure}[b]{0.5\linewidth}
    \centering
    \includegraphics[width=0.75\linewidth]{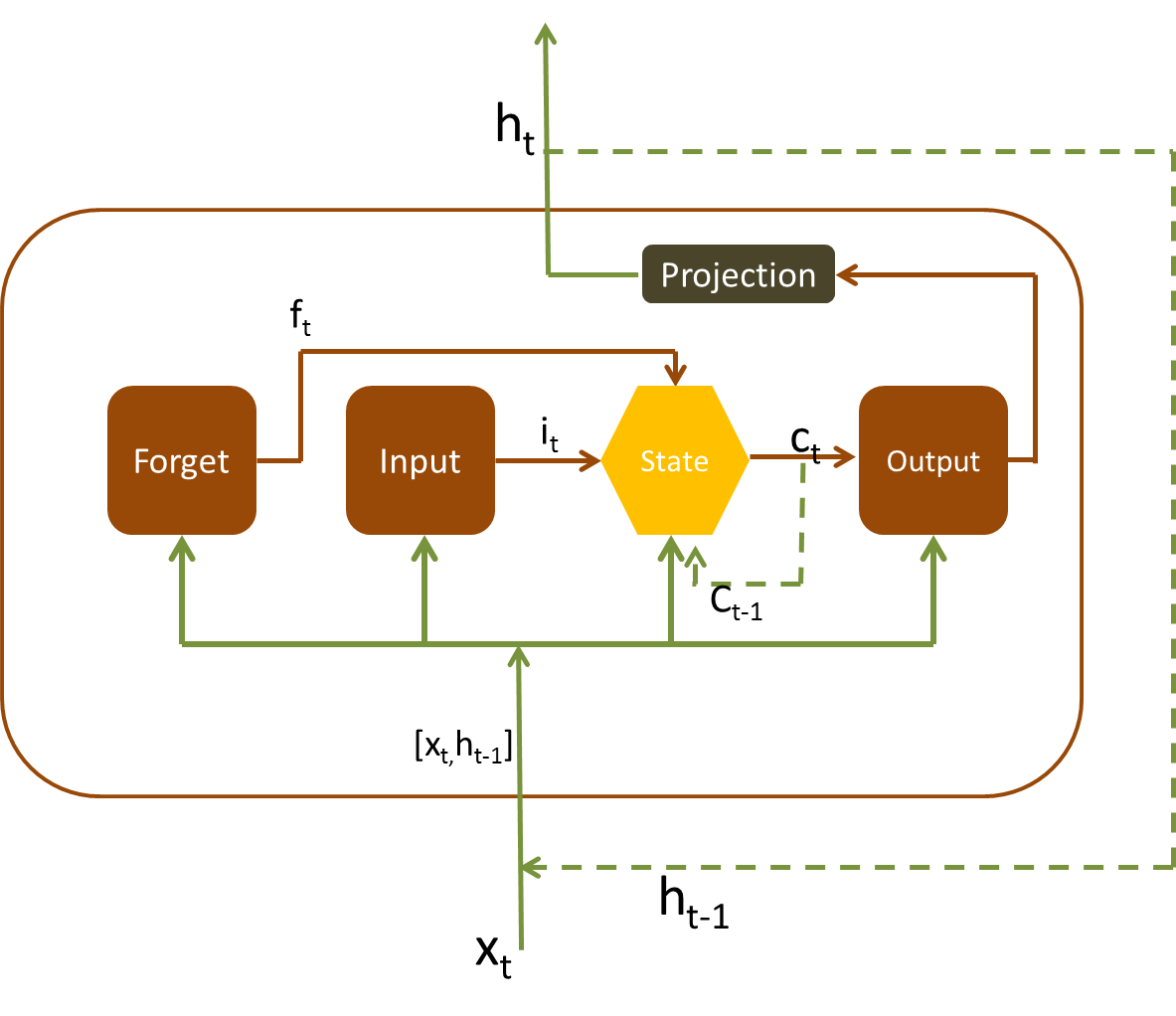} 
   \caption{LSTM with projection layer.}
    \label{fig:lstmproj}
  \end{subfigure}%%
  \begin{subfigure}[b]{0.5\linewidth}
    \centering
    \includegraphics[width=0.75\linewidth]{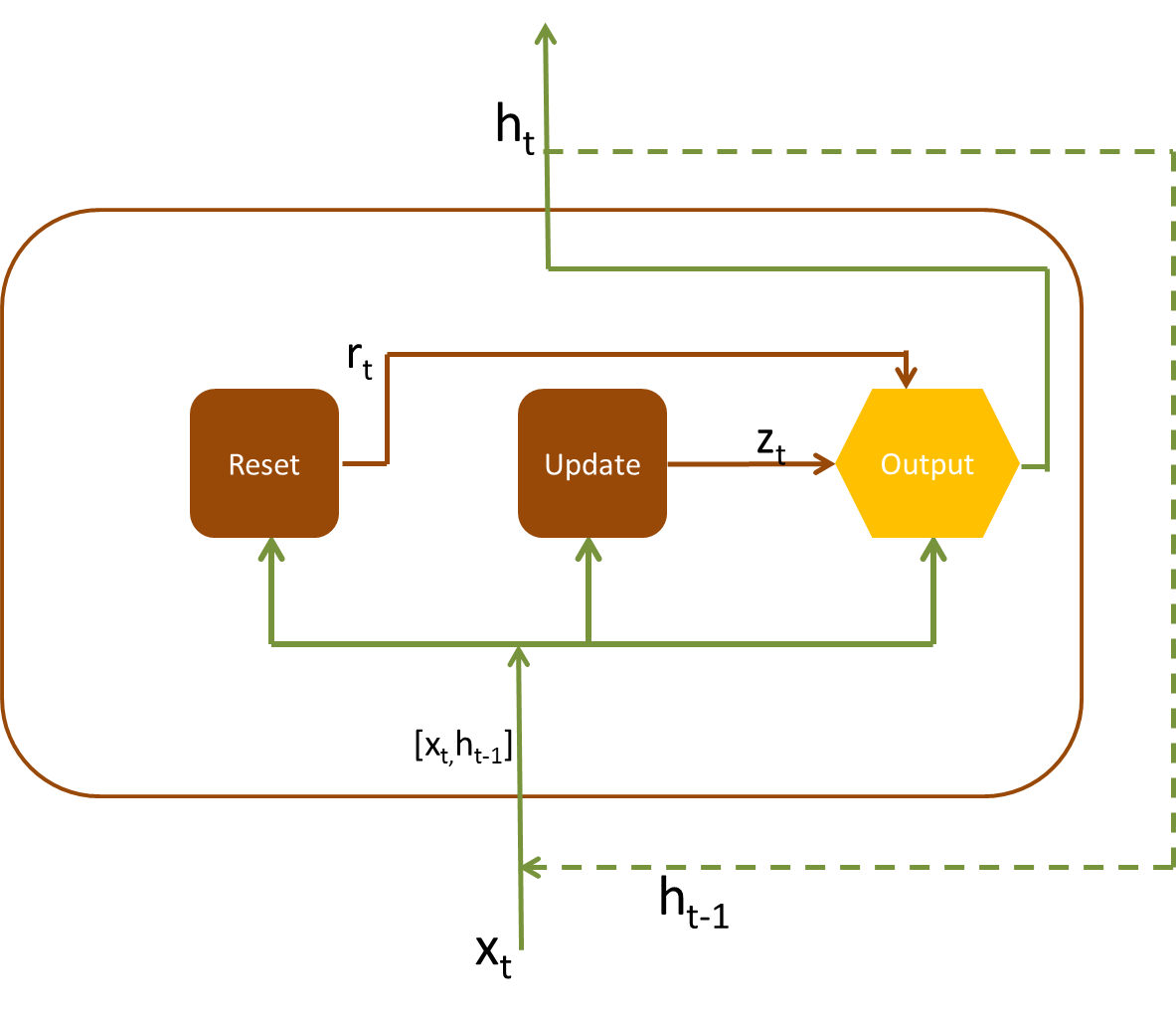} 
 \caption{Gated Recurrent Unit (GRU).}
    \label{fig:gru}
  \end{subfigure} 

    \begin{subfigure}[b]{0.5\linewidth}
    \centering
    \includegraphics[width=0.7\linewidth]{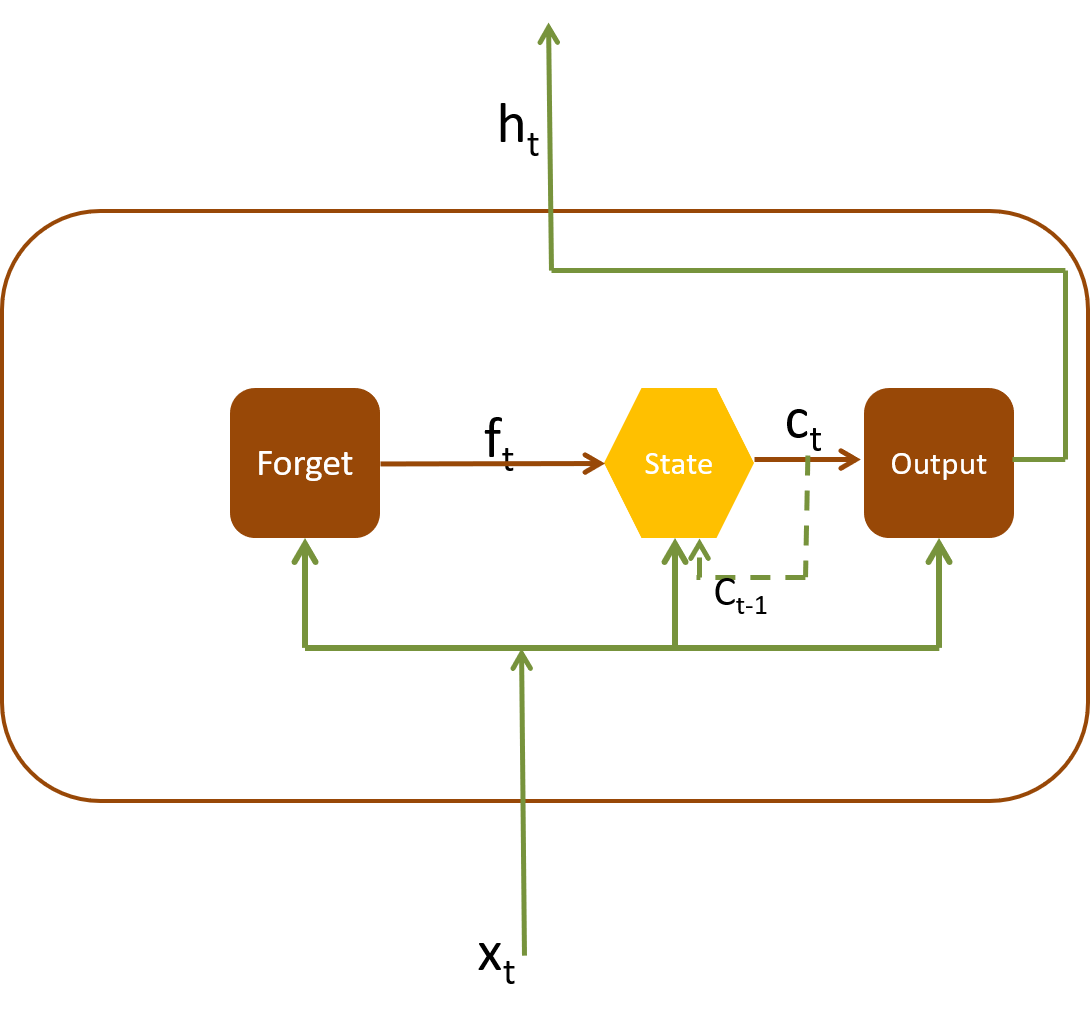} 
      \caption{Quasi-RNN (QRNN).}
    \label{fig:qrnn}
  \end{subfigure}%%
  \begin{subfigure}[b]{0.5\linewidth}
    \centering
    \includegraphics[width=0.7\linewidth]{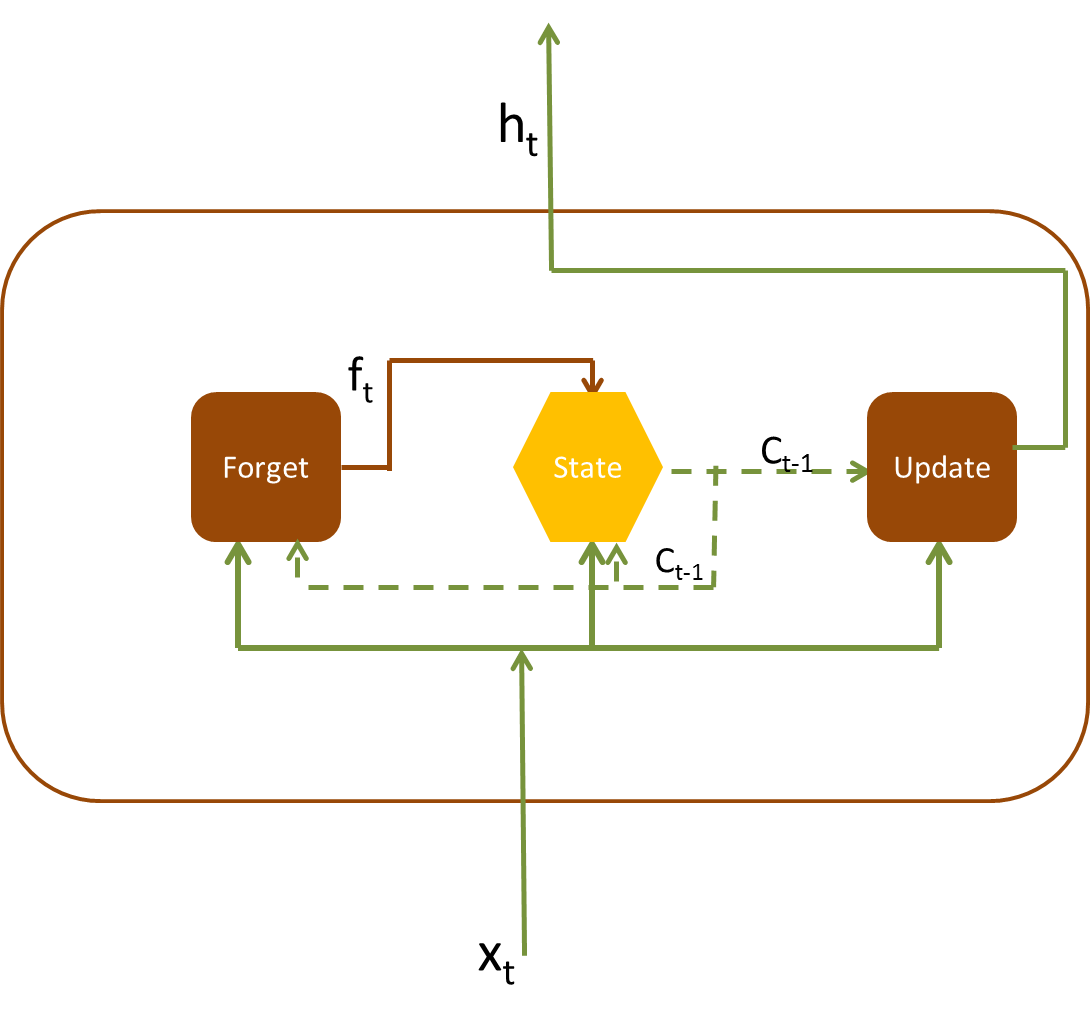} 
    \caption{Simple Recurrent Unit (SRU).}
    \label{fig:sru}
  \end{subfigure} 
  \begin{subfigure}[b]{1\linewidth}
    \centering
    \includegraphics[width=0.75\linewidth]{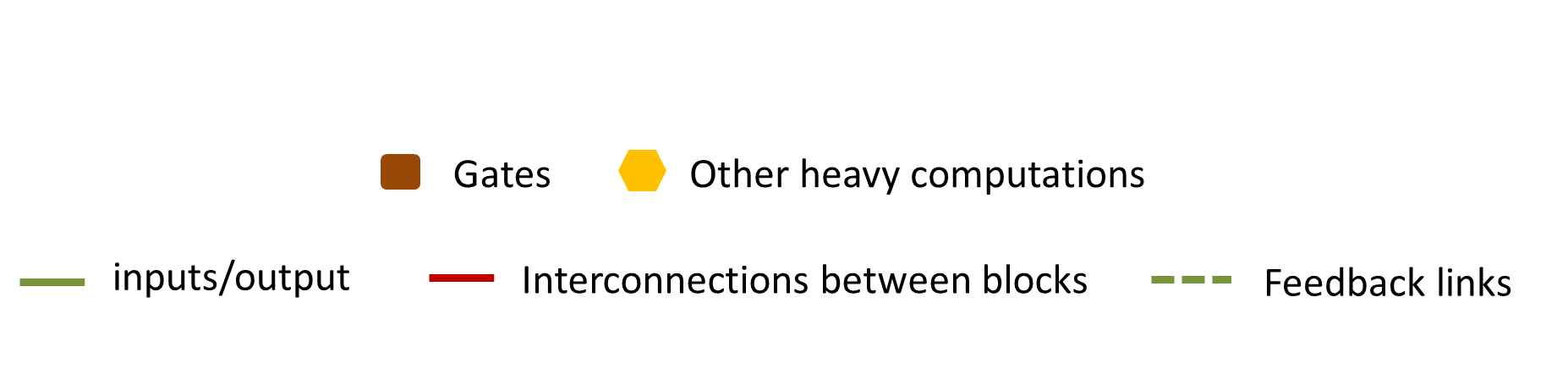} 
    %  \caption{Quasi-RNN (QRNN).}
    \label{fig:legend}
  \end{subfigure}%%
   \caption{Different variations of an RNN layer.}
  \label{fig:rnn-types}
\end{figure*}

First, we explain the LSTM (Long Short Term Memory) layer. Looking at LSTM as a black box, the input to LSTM is a vector combination of the input vector $x_t$ and the previous time-step output vector $h_{t-1}$, where the output vector at time $t$ is denoted as $h_t$. Looking at the structure of an LSTM, it has a memory cell state $C_t$ and three gates. These gates control what is to be forgotten and what is to be updated by the memory state (forget and input gates). They also control the part of the memory state that will be used as an output (output gate). Our description of the LSTM unit is based on its relationship with hardware implementations. Thus, in Figure~\ref{fig:lstm}, we show the LSTM as four blocks instead of three gates because LSTM is composed of four similar computation blocks.

%\begin{figure}[h]
 %  \centering
 %  \includegraphics[width=0.5\columnwidth]{LSTMcomp.png}
  %  \caption{Long Short Term Memory (LSTM).}
  %  \label{fig:lstm}

The computation block is the matrix to vector multiplication of the combination of $x_t$ and $h_{t-1}$ with one of the weight matrices \{$W_f, W_i, W_c, W_o$\}. This is considered the dominant computational task in LSTMs. Each block is composed of a matrix to vector multiplication followed by the addition of a bias vector $\{b_f, b_i, b_c, b_o\}$, and then the application of a nonlinear function. Each block might have element-wise multiplication operations as well. The nonlinear functions used in the LSTM are $tanh$ and $sigmoid$ functions. The four computation blocks are as follow:

\begin{itemize}
    \item \textbf{ {Forget gate}}
    The role of the forget gate is to decide which information should be forgotten. The forget gate output $f_t$ is calculated as
      \begin{equation}
    \label{eq:forget}
        f_t=\sigma(W_f[h_{t-1},x_t]+b_f),
    \end{equation}
    where $x_t$ is the input vector, $h_{t-1}$ is the hidden state output vector, $W_f$ is the weight matrix, $b_f$ is the bias vector, and $\sigma$ is the $sigmoid$ function.
    %multiplying the combination of the input  vector $x_t$ and hidden state output vector $h_{t-1}$ with the weight matrix $W_f$, adding the result to the bias vector $b_f$ and then a $sigmoid$ function is applied. 

    \item \textbf{ {Input gate}}
     The role of the input gate is to decide which information is to be renewed. The input gate output $i_t$ is computed similarly to the forget gate output as
         \begin{equation}
    \label{eq:input}
        i_t=\sigma(W_i[h_{t-1},x_t]+b_i),
    \end{equation}
     using the weight matrix $W_i$ and the bias vector $b_i$.

    \item \textbf{ {State computation}}
    The role of this computation is to compute the new memory state $C_t$ of the LSTM cell. First, it computes the possible values for the new state %$\widetilde{C}_t$
       \begin{equation}
    \label{state1}
        \widetilde{C}_t=\tanh(W_C[h_{t-1},x_t]+b_C),
    \end{equation} 
     where $x_t$ is the input vector, $h_{t-1}$ is the hidden state output vector, $W_c$ is the weight matrix, and $b_c$ is the bias vector.
    %By multiplying the combination of the input vector $x_t$ and hidden state output $h_{t-1}$ with the weight matrix $W_c$, adding bias $b_c$, and applying non-linear function $tanh$. 
   Then, the new state vector, $C_t$ is calculated by the addition of the previous state vector $C_{t-1}$ element-wise multiplied with the forget gate output vector $f_t$ and the new state candidate vector $\widetilde{C}_t$ element-wise multiplied with the input gate output vector $i_t$ as
    \begin{equation}
    \label{eq:state2}
         C_t=f_t \odot C_{t-1} + i_t \odot  \widetilde{C}_t,
    \end{equation}
   where $\odot$ is used to denote the element-wise multiplication.
    
    \item \textbf{ {Output gate}}
    The role of the output gate is to compute the LSTM output. First, the output gate vector $o_t$ is computed as
    %multiplying the combination of the input vector $x_t$ and the hidden state output $h_{t-1}$ with the weight matrix $W_o$, adding the result to the bias vector $b_o$ and applying the $sigmoid \sigma$ function.
    \begin{equation}
      \label{eq:output1}
        o_t=\sigma(W_o[h_{t-1},x_t]+b_o),
    \end{equation}
     where $x_t$ is the input  vector, $h_{t-1}$ is the hidden state output vector, $W_o$ is the weight matrix, $b_o$ is the bias vector, and $\sigma$ is the $sigmoid$ function.
    Then, the hidden state output $h_t$ is computed by applying the element-wise multiplication of the output gate vector $o_t$ (that holds the decision of which part of the state is the output) to the $tanh$ of the state vector $C_t$ as
    \begin{equation}
    \label{eq:output2}
        h_t = o_t \odot tanh(C_t).
    \end{equation}
\end{itemize}

The number of computations and parameters for LSTM are shown in Table~\ref{tab:complstm}. Matrix to vector multiplications dominate the number of computations and parameters. For each matrix to vector multiplication, the input vector $x_t$ of size $m$ and the hidden state output vector $h_{t-1}$ of size $n$ are multiplied with weight matrices of size $(m+n) \times n$. That requires $n(m+n)$ MAC operations, which is equivalent to $nm + n^2$ multiplications and $nm + n^2$ additions. The number of parameters in the weight matrices is $nm + n^2$ as well. Since this computation is repeated four times within the LSTM computation, these numbers are multiplied by four in the total number of operations and parameters for an LSTM. For the models in the studied papers, $n$ is larger than $m$. Thus, $n$ has a dominating effect on the computational complexity of the LSTM.

%\begin{itemize}

\subsubsection{\textbf{LSTM with peepholes}}
\label{subsub:lstmpeep}
Peephole connections were added to LSTMs to make them able to count and measure the time between events~\cite{RNN-peepholes}. As seen in Figure~\ref{fig:lstmpeep}, the output from the state computation is used as input for the three gates. The LSTM gate equations are changed to: 
%\begin{figure*}[h]
%   \centering
%   \includegraphics[width=.6\columnwidth]{LSTMpeep.png}
%    \caption{LSTM with peepholes.}
%    \label{lstmpeep}
%\end{figure*}
 \begin{equation}
    \label{eq:forgetp}
        f_t=\sigma(W_f[h_{t-1},x_t,C_{t-1}]+b_f),
    \end{equation}
    \begin{equation}
    \label{eq:inputp}
        i_t=\sigma(W_i[h_{t-1},x_t,C_{t-1}]+b_i),
    \end{equation} and
   \begin{equation}
      \label{eq:output1p}
        o_t=\sigma(W_o[h_{t-1},x_t,C_{t}]+b_o).
    \end{equation} where $x_t$ is the input vector, $h_{t-1}$ is the hidden state output vector, $C_{t-1}$ is the state vector at time $t-1$, $W_f$, $W_i$, $W_o$ are the weight matrices, and $b_f$, $b_i$ and $b_o$ are the bias vectors.

The number of operations and computations for an LSTM with peepholes are shown in Table~\ref{tab:complstm}. There exist two rows for an LSTM with peepholes. The first one considers the multiplication with the cell state in the three gates as a matrix to vector multiplication. The number of multiplications, additions, and weights increases by $3 \times n^2$. However, the weight matrices multiplied with the cell state can be diagonal matrices~\cite{RNN-Deep-stacked}. Thus, the matrix to vector multiplication can be considered as element-wise vector multiplication, which has become widely used for LSTM with peepholes. In this case, the number of multiplications, additions, and weights will increase by $3n$ only.

\subsubsection{\textbf{ConvLSTM}}
\label{subsub:convlstm}
ConvLSTM is an LSTM with all matrix to vector multiplications replaced with 2D convolutions~\cite{RNN-convlstm}. The idea is that if the input to the LSTM is data that holds spatial relations such as visual frames, it is better to apply 2D convolutions than matrix to vector multiplications.  Convolution is capable of extracting spatial information from the data. The vectors $x_t$, $h_t$, and $C_t$ are replaced with 3-D tensors. One can think of each element in the LSTM vectors as a 2D frame in the ConvLSTM vectors. Convolution weights need less memory than to vector matrices weights. However, using them involves more computation.

The number of operations and parameters required for a convLSTM are shown in Table~\ref{tab:complstm}. The calculated numbers are for a convLSTM without peepholes. If peepholes are added, the number of multiplications, additions, and weights will increase by $3n$. Since the main change from an LSTM is the replacement of the matrix to vector multiplications with convolutions, the change in the number of operations and parameters would be via the $nm + n^2$ factor that appears in multiplications, additions, and the number of weight equations. The number of multiplications and additions (MACs) in convolutions of input vector $x_t$ and hidden state output vector $h_{t-1}$ is $rcn m{k_i}^2 +r c n^2 \times {k_s}^2 $, where $r$ is the number of rows and $c$ is the number of columns in the frames, $n$ is the number of frames in input $x_t$, $m$ is the number of frames in output $h_t$ (or the number of hidden cells), $k_i$ is the size of the filter used with $x_t$, and $k_s$ is the size of the filter used with $h_{t-1}$.  
The number of weights is the size of the filters used for convolutions.

%\textbf{Changes applied to enhance computational complexity}

\subsubsection{\textbf{LSTM with projection layer}}
\label{subsub:lstmproj}
The LSTM is changed by adding one extra step after the last gate \cite{RNN-projection}. This step is called a projection layer. The output of the projection layer is the output of the LSTM and the feedback input to the LSTM in the next time-step, as shown in Figure~\ref{fig:lstmproj}. Simply, a projection layer is like an FC layer. The purpose of this layer is to allow an increase in the number of hidden cells while controlling the total number of parameters. This is performed by using a projection layer that has a number of units $p$ less than the number of hidden cells. The dominating factor in the number of computations and the number of weights will be $4 p  n$ instead of $4 n^2$, where $n$ is the number of hidden cells and $p$ is the size of the projection layer. Since $p<n$, $n$ can increase with a smaller effect on the size of the model and the number of computations.

In Table~\ref{tab:complstm}, we show the number of operations and parameters required for an LSTM with a projection layer. In the original paper proposing the projection layer, the authors considered the output layer of the RNN as a part of the LSTM~\cite{RNN-projection}. The output layer was an FC layer that changes the size of the output vector to $o$, where $o$ is the output size. Thus, there is an extra $p o$ term in the number of multiplications, additions, and weights. We put the extra terms between curly brackets to show that they are optional terms. The projection layer can be applied to an LSTM with peepholes as well. In Table~\ref{tab:complstm}, we show the number of operations and parameters for an LSTM with peepholes and a projection layer.

%\begin{figure*}[h]
%   \centering
%   \includegraphics[width=.6\columnwidth]{LSTMproj.png}
%    \caption{LSTM with projection layer.}
%    \label{lstmproj}
%\end{figure*}

\subsubsection{\textbf{GRU}}
\label{subsub:gru}
The Gated Recurrent Unit (GRU) was proposed in 2014~\cite{RNN-GRU}. The main purpose was to make the recurrent layer able to capture the dependencies at different time scales in an adaptive manner~\cite{RNN-GRU-study}.
However, the fact that GRU has only two gates (three computational blocks) instead of three (four computational blocks) as with the LSTM makes it more computationally efficient and more promising for high-performance hardware implementations. The three computational blocks are as follows:
\begin{itemize}
    \item  \textbf{Reset gate}
    The reset gate is used to decide whether to use the previously computed output or treat the input as the first symbol in a sequence. The reset gate output vector $r_t$ is computed as
     \begin{equation}
    \label{eq:gru-r}
        r_t=\sigma(W_r[h_{t-1},x_t]),
    \end{equation}
        where $x_t$ is the input  vector, $h_{t-1}$ is the hidden state output vector, $W_r$ is the weight matrix, and $\sigma$ is the $sigmoid$ function.

    %multiplying the combination of the input vector $x_t$ and the hidden state output vector $h_{t-1}$ with the weight matrix $W_r$ and applying $sigmoid$ function to the result.

    \item \textbf{ Update gate}
     The update gate decides how much of the output is updated. The output of the update gate $z_t$ is computed as the reset gate output $r_t$ using the  weight matrix $W_z$ as
\begin{equation}
    \label{eq:gru-z}
        z_t=\sigma(W_z[h_{t-1},x_t]).
    \end{equation}
        
\newcommand{\tabRNNLayer}{2.75cm}
\begin{landscape}
\begin{table}[p]
\centering
\begin{threeparttable}
\caption{Comparing LSTM and its variations.}
\begin{tabular}{|p{\tabRNNLayer}||p{3.5cm}|p{3.2cm}|p{2.7cm}||p{3.1cm}|p{2.2cm}|}
\hline \hline

RNN layer & \multicolumn{3}{c||}{Number of Operations}&   \multicolumn{2}{c|}{Number of Parameters} \\ 
\hline \hline

& Multiplications & Additions & Nonlinear & Weights & Biases \\ 
\hline \hline

\multirow{2}{*}{LSTM} & $4 n^2 + 4 n m + 3 n $&
$4 n^2 + 4 n m + 5 n  $&
$5 n  $&
$4 n^2 + 4 n m $& $4 n$\\ 

\cline{2-6} 
&$=LSTM_{mul}$&$             =LSTM_{add}$&$=LSTM_{nonlinear}$&$=LSTM_{weights}$&$=LSTM_{biases}$ \\ \hline \hline

\multirow{2}{*}{\parbox{\tabRNNLayer}{LSTM + peepholes}}& $7 n^2 + 4 n m + 3 n $ & 
$7n^2 + 4 n m + 5 n$ &
$5 n $ &
$7n^2 + 4 n m  $&
$4 n$\\ 
\cline{2-6} 

&$= LSTM_{mul} + 3 n^2$&$=LSTM_{add} + 3 n^2$&$=LSTM_{nonlinear}$&$=LSTM_{weights} + 3 n^2$&$ =LSTM_{biases}$
\\ \hline \hline

\multirow{2}{*}{\parbox{\tabRNNLayer}{LSTM + peepholes (diagonalized)}}& $4 n^2 + 4 n m + 6 n$ & $4n^2 + 4 n m + 8 n$ & $5 n$ &$4n^2 + 4 n m + 3 n$&$4  n$\\ 
\cline{2-6}

&$= LSTM_{mul} + 3 n$&$=LSTM_{add} + 3 n$&$=LSTM_{nonlinear}$&$=LSTM_{weights} + 3 n$&$ =LSTM_{biases}$ \\ \hline \hline

\multirow{2}{*}{\parbox{\tabRNNLayer}{LSTM + projection}}& $4 n p + 4 n m + 3 n+ n p +\{p o\}$ &  $4 n p + 4 n m + 5 n+ n p +\{p o\}$&$5  n$&$4 n p + 4 n m + n p +\{p o\}$&$4  n$\\ 
\cline{2-6}

&$=LSTMProj_{mul}$&$=LSTMProj_{add}$&$=LSTM_{nonlinear}$&$=LSTMProj_{weights}$&$= LSTM_{biases}$ \\ \hline \hline

\multirow{2}{*}{\parbox{\tabRNNLayer}{LSTM + peepholes  (diagonalized) + projection}} &$4 n p + 4 n m + 6  n +n p +\{p o\}$ &  $4 n p + 4 n m + 8 n+ n p +\{p o\}$&$5  n$&$4 n p + 4 n m +3  n + n p +\{p o\}$& $4  n$\\ 
\cline{2-6}

&$=LSTMProj_{mul} +3  n$&$=LSTMProj_{add}+ 3  n$&$=LSTM_{nonlinear}$&$=LSTMProj_{weights} + 3  n$&$ =LSTM_{biases}$ \\ \hline \hline

ConvLSTM & $4 r c n m {k_i}^2$  $+ 4 r c n^2 {k_s}^2 +3 n$ 
 & $4 r c n m {k_i}^2$ $ + 4 r c n^2 {k_s}^2 +5 n$ &$5 n$&$4 n m {k_i}^2  + 4 n^2 {k_s}^2 $& $4 n$ \\
\hline \hline

\multirow{2}{*}{\parbox{\tabRNNLayer}{GRU}} & $3 n^2 + 3 n m + 3 n$ & $3 n^2 + 3 n m + 2 n$ & $3 n $&$3 n^2 + 3 n m $&-\\ 
\cline{2-6}

&$=0.75 LSTM_{mul}  $&$=0.75 LSTM_{add}  $&$=0.6 LSTM_{nonlinear}$&$=0.75 LSTM_{weights}  $&- \\ \hline \hline

QRNN &$3 k n m + 3 n  $&$3 k n m +2 n $&$3 n$&$3k n m$&-\\ 
\hline \hline

SRU &$3 n m + 6 n$  &$3 n  m + 8 n $ &$2 n$&$3 n m + 2  n$ &$ 2 n$\\ 
\hline \hline

\end{tabular}
\label{tab:complstm}%
\begin{tablenotes}
\item[]In the table we use the following symbols: $m$ is the size of input vector $x_t$, $n$ is the number of hidden cells in $h_t$, $p$ is the size of the projection layer, $o$ is the size of the output layer, $r$ is the number of rows in a frame, $c$ is the number of columns in a frame, $k_i$ is size of the 2D filter applied to $x_t$, $k_s$ is the size of the 2D filter applied to $h_{t-1}$, and $k$ is the size of 1D convolution filter. The term $\{p o\}$ is an optional term as discussed in Section~\ref{subsub:lstmproj}. 
\end{tablenotes}
\end{threeparttable}
\end{table}	
\end{landscape}

     \item \textbf{ Output computation}
The role of this block is to compute the hidden state vector $h_t$. First, it computes the possible values for the hidden state vector $\widetilde{h}_t$ 

%multiplying the combination of the input vector $x_t$ and hidden state output $h_{t-1}$ element wise multiplied with $r_t$ with the weight matrix $W$ and applying non-linear function $tanh$. The reset gate output vector $r_t$ decides how much of $h_{t-1}$ can contribute.
\begin{equation}
    \label{eq:gru-hcap}
        \widetilde{h}_t=\tanh(W[r_t \odot h_{t-1},x_t]),
    \end{equation}
        where $x_t$ is the input  vector, $h_{t-1}$ is the hidden state output vector, and $W$ is the weight matrix.
        %The reset gate output vector $r_t$ decides how much of $h_{t-1}$ can contribute in the computation of  $\widetilde{h}_t$.
    Then, the hidden state vector $h_t$ is  computed from the old output $h_{t-1}$ and the new possible output $\widetilde{h}_t$% relying on the update gate output vector $z_t$ (that decides how much of the output will be updated) as   
    as
    \begin{equation}
    \label{eq:gru-h}
         h_t=(1 - z_t) \odot h_{t-1} + z_t \odot  \widetilde{h}_t.
    \end{equation}

\end{itemize}

 As with LSTM, we visualize a GRU in Figure~\ref{fig:gru} as three blocks, not two gates, as it has three blocks of matrix to vector multiplications. In Table~\ref{tab:complstm}, we show the number of operations and parameters required for a GRU. The number of operations and parameters is approximately 0.75 the number of operations and parameters in the LSTM.

%   \begin{figure*}[h]
%   \centering
%   \includegraphics[width=.6\columnwidth]{GRU.png}
%    \caption{Gated Recurrent Unit (GRU).}
%    \label{gru}
%\end{figure*}

 %   GRU is similar to LSTM in having the ability to remember a specific feature for long time ... complete comparison.

\subsubsection{\textbf{QRNN and SRU}}
\label{subsub:qrnnsru}
The purpose of Quasi-RNN (QRNN) \cite{RNN-quasi} and Simple Recurrent Unit (SRU)~\cite{RNN-SRU} is to make the recurrent unit friendlier for computation and parallelization. The bottleneck in an LSTM/GRU is the matrix to vector multiplications. It is difficult to parallelize this part because it depends on the previous time-step output $h_{t-1}$ and previous time-step state $C_{t-1}$. In QRNN/SRU, $h_{t-1}$ and $C_{t-1}$ are removed from all matrix to vector multiplications and appear only in element-wise operations.
QRNN has two gates and a memory state. It has three heavy computational blocks. In these blocks, only the input vector $x_t$ is used as input. It replaces the matrix to vector multiplications with 1D convolutions with inputs along the time-step dimension. For instance, if the filter dimension is two, convolution is applied on $x_t$ and $x_{t-1}$. The three computation blocks compute the forget gate vector $f_t$, candidate for new state vector $ \widetilde{C}_t$, and the output gate vector $o_t$ as

    \begin{equation}
    \label{eq:QRNNf}
        f_t=\sigma(W_f \ast x_t),
    \end{equation} 
        \begin{equation}
    \label{eq:QRNNz}
    \widetilde{C}_t=\tanh(W_c \ast x_t),
    \end{equation}
    and
    \begin{equation}
      \label{eq:QRNNo}
      o_t=\sigma(W_o \ast x_t),
    \end{equation}

where $W_f$ and $W_c$, $W_o$ are the convolution filter banks and ``$\ast$'' is to denote the convolution operation.

The state vector $C_t$ is computed as
\begin{equation}
     \label{eq:QRNNC}
         C_t= f_t \odot C_{t-1} + (1 - f_t) \odot \widetilde{C}_t
    \end{equation}
    and the hidden state vector $h_t$ is computed using equation~\ref{eq:output2}.

 Figure~\ref{fig:qrnn} is used to visualize the QRNN layer. The number of operations and parameters required for a QRNN is shown in Table~\ref{tab:complstm}, where $k$ is the size of the convolution filter.
% \begin{figure*}[h]
%   \centering
%   \includegraphics[width=.6\columnwidth]{QRNN.png}
%    \caption{Quasi-RNN (QRNN).}
%    \label{qrnn}
%\end{figure*}

The SRU has two gates and a memory state as well. The heavy computational blocks (three blocks) are matrix to vector multiplications, not convolutions. The two gates (forget and update gates) are computed using the equations
 \begin{equation}
    \label{eq:sruf}
        f_t=\sigma(W_f x_t + v_f \odot c_{t-1} + b_f )
    \end{equation}
    and
    \begin{equation}
    \label{eq:srur}
        r_t=\sigma(W_r x_t + v_r \odot c_{t-1} + b_r )
    \end{equation} respectively.
In both gate calculations, $C_{t-1}$ is used but only for element-wise multiplications. The parameter vectors $v_f$ and $v_r$ are learned with weight matrices and biases during training.

The third computational block is the state computation $C_t$
\begin{equation}
     \label{eq:sruC}
         C_t= f_t \odot C_{t-1} + (1 - f_t) \odot (W x_t),
    \end{equation} 
   
    where $C_{t-1}$ is the old state vector and $x_t$ is the input vector. The computation is controlled by the forget gate output vector $f_t$ that decides what is to be forgotten and what is to be treated as new.
   
    Finally, the SRU output $h_t$ is computed from the new state $C_t$ and the input vector $x_t$ checked by the update gate (which decides the parts of the output that are taken from the new state and the parts that are taken from input) using the equation
        \begin{equation}
     \label{eq:sruh}
         h_t= r_t \odot C_t + (1 - r_t) \odot x_t.
    \end{equation}    

Figure~\ref{fig:sru} visualizes the SRU. The output computation is performed in the same block with the update gate. It is worth observing that in neither QRNN nor SRU, $h_{t-1}$ are used in the equations -- only the old state $C_{t-1}$ is used.
The number of operations and parameters for an SRU is shown in Table~\ref{tab:complstm}.
% \begin{figure*}[h]
%   \centering
%   \includegraphics[width=.6\columnwidth]{sru.png}
%    \caption{Simple Recurrent Unit (SRU).}
%    \label{sru}
%\end{figure*}

In Table~\ref{tab:complstm}, we compare the LSTM and all of its variations against the memory requirements for the weights and the number of computations per single time-step. This comparison helps to understand the required hardware platform for each of them.
To make it easier for the reader to understand the difference between the LSTM and the other variants, we show the equations for operations and parameters in terms of LSTM operations and parameters if they are comparable.  
\subsection{\textbf{Output layers}}
\label{subsec:output}
The output layers in the RNN model are the FC layers and the output function.

\subsubsection{\textbf{FC (Fully Connected) Layers}}

The RNN model might have one or more FC layers after the recurrent layers. Non-linear functions may be applied between the FC layers as well. This is called fully connected because each neuron in the input is connected to each neuron of the output. Computationally, this is done by matrix to vector multiplication using a weight matrix of size $Input_{size} \times output_{size}$, where $Input_{size}$ is the size of the input vector and $Output_{size}$ is the size of the output vector. One purpose of the FC layer in RNN models can be to change the dimension of the hidden state output vector $h_t$ to the dimension of the RNN model output to prepare it for the output function. In this case, the FC layer might be replaced by adding a projection layer in the recurrent layer.% as discussed in Section~\ref{subsec:RNN-types}.

\subsubsection{\textbf{Output function}}

The output function is the final step in the neural networks inference. It generates the output of the neural network model. This output can be a prediction, classification, recognition, and so on. For example, in a text prediction problem, the softmax function is used as an output function. The output is a vector of probabilities that sum to one. Each probability corresponds to one word. The word with the highest probability becomes the prediction of the neural network~\cite{Bengio-softmax-predict}.

\subsection{\textbf{Processing of data in RNN models}}
There are many ways that processing of data may vary in RNN models. The first is to vary the way time steps are treated. This is influenced by the nature of the application, which may have inputs with temporal relations, outputs with temporal relations, or both. The second form of variation is related to bidirectional RNNs. We discuss below how a bidirectional RNN can process inputs both forwards and backwards in time. We also discuss what is meant by a deep RNN model.

\subsubsection{\textbf{RNN unfolding variations through time-steps}}
\label{subsub:unrolling}
RNN unfolding/unrolling is performed to reveal the repetition in the recurrent layer and to show the number of time steps required to complete a task. Unfolding the RNN illustrates the different types of RNN models one can meet.
\begin{itemize}
\item{\textbf{One to many}}
A one to many model generates a sequence of outputs for a single input, as shown in Figure~\ref{fig:onetomany}. Image captioning is one example~\cite{rnn-activity}. The model takes one image as input and generates a sentence as an output. The words of the sentence compose a sequence of temporally related data. In this case, the temporal sequence is only in the output.

 \begin{figure}[ht] 
  \begin{subfigure}[b]{0.5\linewidth}
    \centering
    \includegraphics[width=0.9\linewidth]{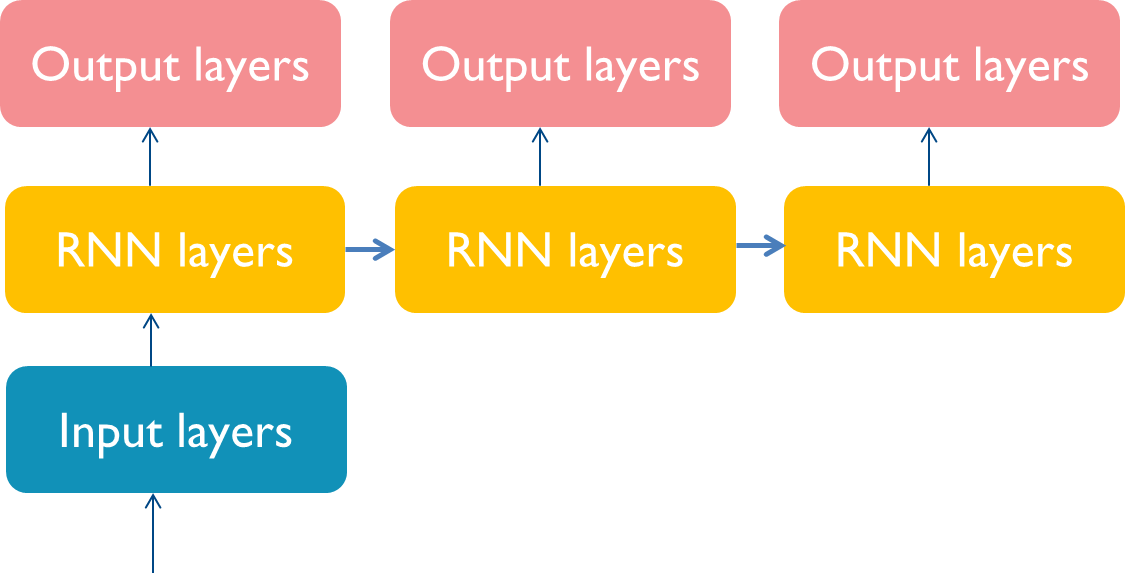} 
    \caption{One to Many RNN.}
    \label{fig:onetomany} 
    \vspace{4ex}
  \end{subfigure}%% 
  \begin{subfigure}[b]{0.5\linewidth}
    \centering
    \includegraphics[width=0.9\linewidth]{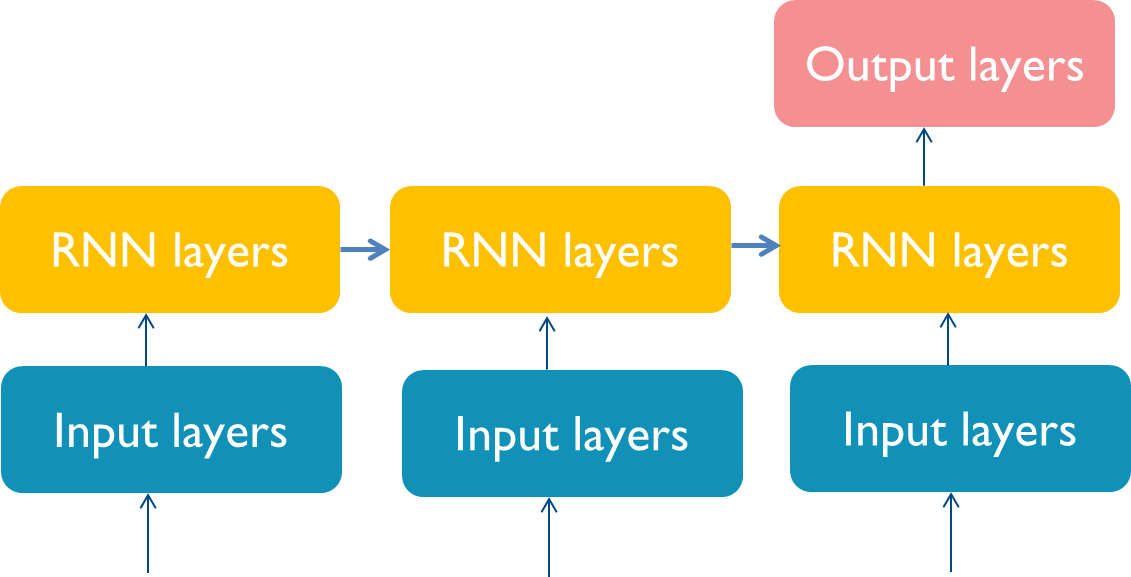} 
    \caption{Many to One RNN.}
    \label{fig:manytoone}
    \vspace{4ex}
  \end{subfigure} 
  \begin{subfigure}[]{\linewidth}
    \centering
    \includegraphics[width=0.6\linewidth]{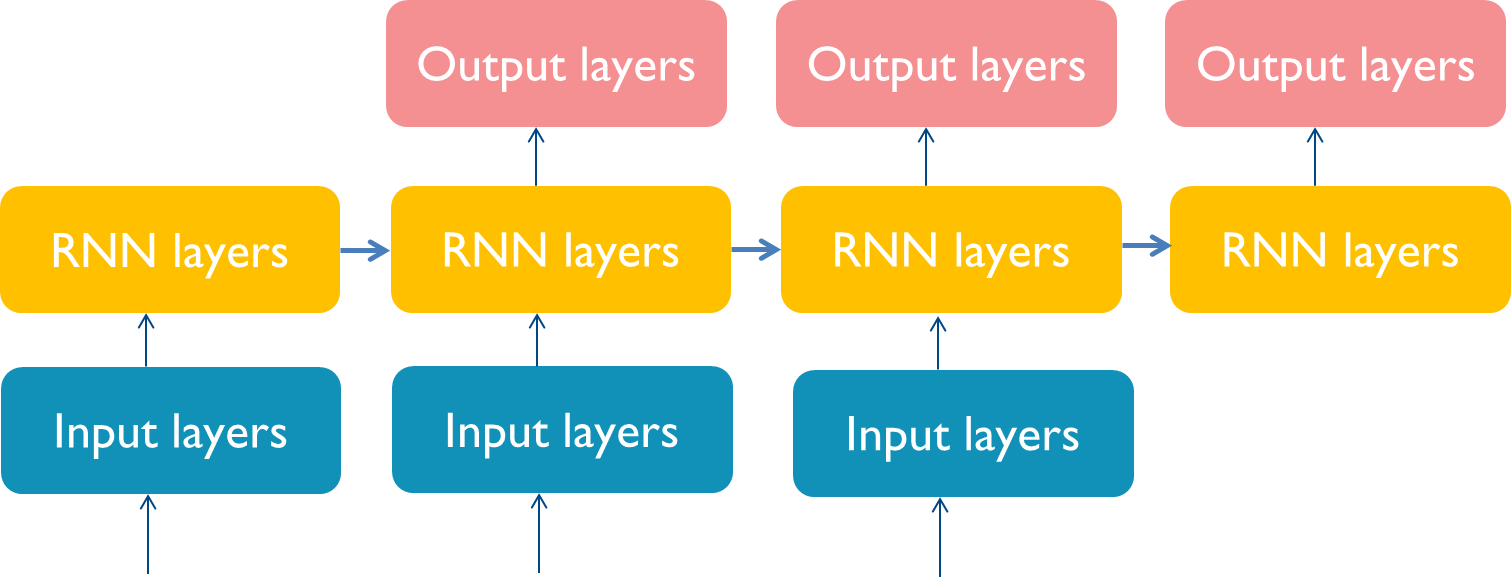} 
    \caption{Many to Many RNN.}
    \label{fig:manytomany}
    \vspace{4ex}
  \end{subfigure} 
  \caption{Unfolding RNN models through multiple time steps.}
  \label{fig:rnn-types}
\end{figure}

%\begin{figure*}[h]
%   \centering
%   \includegraphics[width=0.35\columnwidth]{onetomany.png}
 %   \caption{One to Many RNN.(may remove FC)}
 %   \label{onetomany}
%\end{figure*}

\item{\textbf{Many to one}}
A many to one model combines a sequence of inputs to generate a single output, as shown in Figure~\ref{fig:manytoone}. Activity recognition~\cite{rnn-activity} and sentiment analysis~\cite{rnn-sentiment-manytoone} are two examples. In activity recognition, the model takes a sequence of images as input and determines the activity taking place in the images. In sentiment analysis, the model takes a sequence of words (sentence) as input and generates a single emotion at the end. In this case, the temporal sequence sequence is only in the input.

%\begin{figure*}[h]
%   \centering
%   \includegraphics[width=0.4\columnwidth]{manytoone.png}
%    \caption{Many to One RNN.(may remove FC)}
%    \label{manytoone}
%\end{figure*}

\item{\textbf{Many to many}}
A many to many model has a sequence in the input and a sequence in the output, as shown in Figure~\ref{fig:manytomany}. Language translation~\cite{rnn-language} and video description~\cite{rnn-activity} are two examples. In language translation, the model has a sequence of words (sentence) as an input and a sequence of words (sentence) as an output. In video description applications, the model has a sequence of image frames as input and a sequence of words (sentence) as output.
\par

%\begin{figure*}[h]
%   \centering
%   \includegraphics[width=0.4\columnwidth]{manytomany.png}
%    \caption{Many to Many RNN.(may remove FC)}
%    \label{manytomany}
%\end{figure*}

\item{\textbf{One to one} }
There is no RNN model with one to one unrolling. One to one simply means that there is no temporal relation contained in the inputs or the outputs (a feedforward neural network).

\end{itemize}
\subsubsection{\textbf{Bidirectional RNN}}
\label{subsubsec:bi-rnn}
In Bidirectional RNN, input can be fed into the recurrent layer from two directions: past to future and future to past. That requires a duplication of the recurrent layer, so that two recurrent layers work simultaneously, each processing input in a different temporal direction. This can help the network to better understand context by obtaining data from the past and the future at the same time. This concept can be applied to different variations of recurrent layers such as BiLSTM~\cite{rnn-bi-lstm} and BiGRU~\cite{rnn-bi-gru}.

\subsection{\textbf{Deep Recurrent Neural Networks (DRNN)}}
\label{subsubsec:DRNN}
\begin{figure}[h]
   \centering
   \includegraphics[width=0.4\columnwidth]{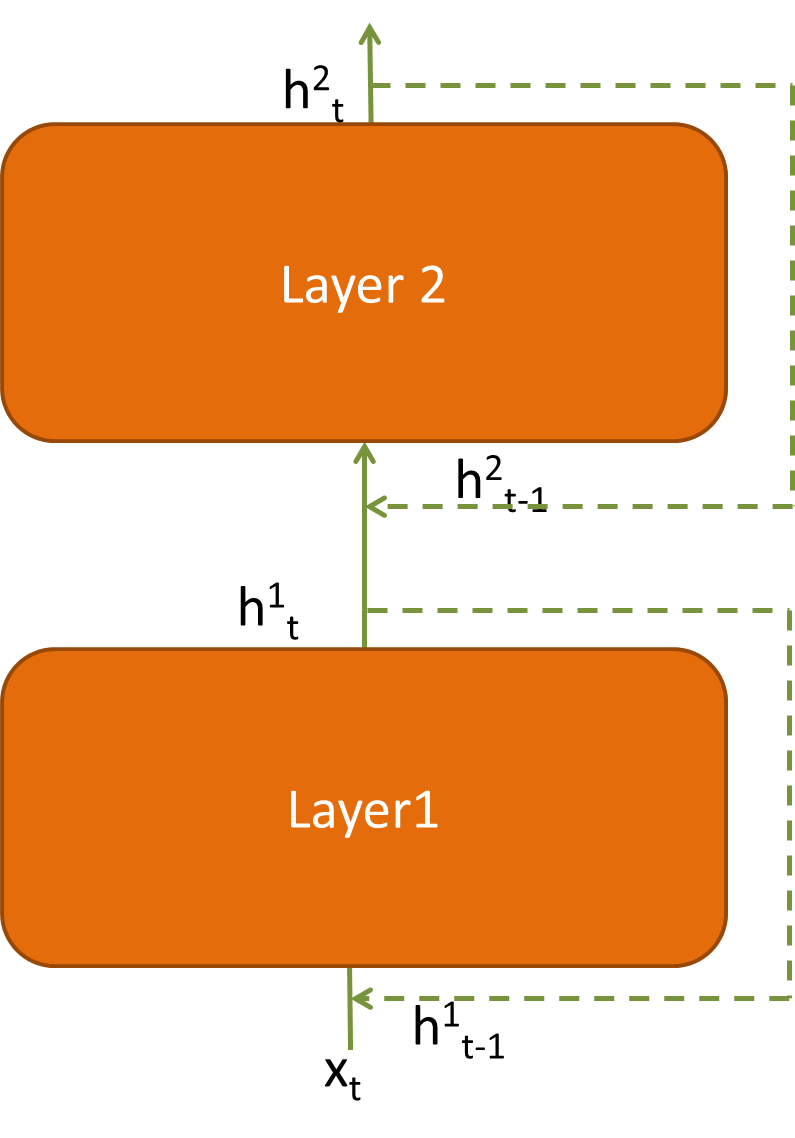}
   \caption{Stacked RNN. The first layer output is $h{^1_t}$ and the second layer output is $h{^2_t}$.  }
    \label{fig:stacked}
\end{figure}

Making a neural network a deep neural network is achieved by adding non-linear layers between the input layer and the output layer~\cite{deep-bengio}. This is straightforward in feedforward NNs. However, in RNNs, there are different approaches that can be adopted. Similarly to feedforward NNs, there can be a stack of recurrent layers (stacked RNN)~\cite{RNN-Deep-stacked} as shown in Figure~\ref{fig:stacked}, where we have a stack of two recurrent layers. The output of the first layer is considered as the input for the second layer. Alternately, the extra non-linear layers can be within the recurrent layer computations~\cite{RNN-deep}. Extra non-linear layers can be embedded within the hidden layer vector $h_{t}$ calculation, where the $x_t$ and $h_{t-1}$ vectors used to calculate $h_t$, pass through additional non-linear layers. This model is called the deep transition RNN model. The extra non-linear layers can also be added in computing the output from the hidden state vector; this model is called the deep output RNN model. It is possible to have an RNN model that is both a deep transition and a deep output RNN model~\cite{Dai:2018}. One other way to have extra non-linear functions within the recurrent layer is to have them within the gate calculations -- a method called H-LSTM (Hidden LSTM).  
%\subsection{\textbf{Output layers}}
%\label{subsec:output}
%\input{source/output.tex}

%\input{source/models.tex}

%In this section, we have discussed the RNN models, recurrent layer variations and their impact on implementation efficiency, and applications and their corresponding dataset. In the next section, we discuss the optimizations applied to RNN to implement them efficiently on embedded platforms.
%\input{source/RNNshort.tex}
%\input{source/types.tex}
%--------------------

\section{Optimizations for RNNs}
\label{Sec:Optimizations}
\label{sec:optimizations}
\begin{figure*}[h]
   \centering
   \includegraphics[width=1.5\columnwidth]{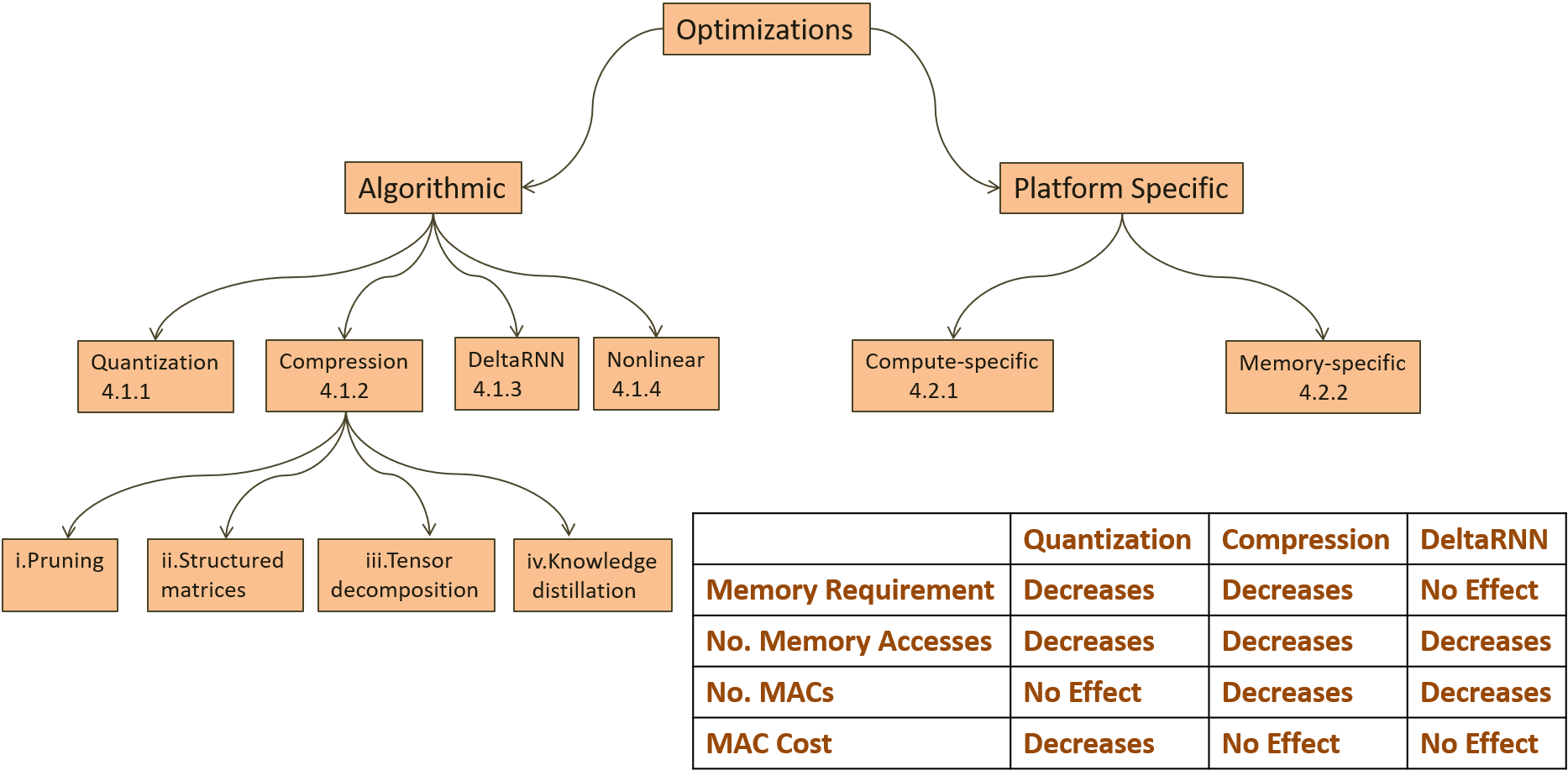}
    \caption{Optimizations applied to RNN applications with section numbers indicated, comparing the effect of different algorithmic optimizations on memory and computation requirements.}
    \label{fig:optimizations}
\end{figure*}

As with all neural network applications, RNN applications are based upon intensive operations performed on high precision values. They therefore require high computation power, large memory bandwidth, and high energy consumption. Because of the resource constraints of embedded platforms, there is a need to decrease the computation and memory requirements of RNN applications. In this section, we present optimizations that have been applied to RNNs to realize them on embedded systems. In Section~\ref{Sec:RNNonHW} which follows, we discuss hardware implementations of RNNs on embedded platforms and how they relate to the optimizations presented here. Researchers have been working on two types of optimizations. The first type is related to the RNN algorithms themselves, where RNN algorithms are modified to decrease computation and memory requirements. The modification should have no effect or only a limited effect on accuracy. The second type of optimization is related to the embedded platform, where hardware improvements are applied to increase the parallelization of the application and decrease the overhead of memory accesses. Figure~\ref{fig:optimizations} illustrates these two types of optimizations.
%\resizebox{\linewidth}{!}{%
%\Tree[.Optimizations [.{Algorithmic} [.{Pruning}  ][.Quantization ][{Delta RNN} ] [ .Others ]]
 %         [.{Platform Specific} [.Parallelization [.{Unrolling+Pipelining} [{Inner loop} ][{Over timesteps} ] ]
  %              [.Tiling ]][ .Memory [.{Where to store?} [.ip ][.Off-chip ][.Hybrid ]]
   %       [.{How to shrink?} ]]]]}
\subsection{\textbf{Algorithmic optimizations}}
\label{sec:algsection}
In this section, we discuss the different algorithmic optimizations that may be performed on the recurrent layer of an RNN application to decrease its computation and memory requirements. We discuss how these optimizations are carried out, and how they affect accuracy. Applying optimizations directly to inference can have unacceptable effects on accuracy. Thus, training the network would be
required to enhance the accuracy. optimizations may
be applied during the model main training or after the model is trained and
then the model is retrained for some epochs (training cycles).

Different datasets measure accuracy using different units. For some units higher values are better, while for others lower values are better. To provide a unified measure of the change in accuracy, we calculate the percentage change in accuracy from the original value to the value after applying the optimization method as
   \begin{equation}
    \label{eq:accuracy}
         a_\Delta= (-1)^{\alpha} \frac{V_a-V_b}{V_b} \times 100,
    \end{equation}
    where $ a_\Delta$ is the effect of the optimization method on accuracy as a percentage of the original accuracy value, $V_b$ is the value of accuracy before optimization, $V_a$ is the value of accuracy after optimization, and $\alpha$ is an indicator that has a value of $0$ if higher accuracy values are better and $1$ if lower accuracy values are better.
    Thus, if the baseline accuracy achieved by the original model without optimizations is 96$\%$ and the accuracy after optimization is  94$\%$, the effect of optimization on accuracy is $-2.1\%$. If the accuracy after optimization is $98\%$, the effect of optimization on accuracy is $+2.1\%$. If the optimization has no effect on accuracy, then the effect on accuracy is $0\%$.

 As shown in Figure~\ref{fig:optimizations}, the algorithmic optimizations are quantization, compression, deltaRNN, and nonlinear. The first three optimizations are applied to the matrix to vector multiplications operations and the last is applied to computation of non-linear functions. The table in Figure~\ref{fig:optimizations} compares quantization, compression, and deltaRNN with their effect on memory requirements, number of memory accesses, number of computations, and MAC operation cost. MAC operation cost can be decreased by decreasing the precision of operands.

\subsubsection{\textbf{Quantization}}
\label{subsub:quantize}
Quantization is a reduction in the precision of the operands. Quantization can be applied to the network parameters only, or to the activations and inputs as well. While discussing quantization, there are three important factors to consider. First, the number of bits used for weights, biases, activations, and inputs. Second, the quantization method. The quantization method defines how to store the full precision values in a lower number of bits. Third, discussing whether quantization was applied with training from the outset or the model was re-trained after applying quantization. These three factors all affect accuracy. However, they are not the only factors affecting accuracy, which may also be affected by model architecture, 
dataset, and other factors. Yet, these three factors have more relevance when applying quantization to the RNN model. 
%\begin{itemize}
%\item{16-bit fixed-point}

In discussing quantization methods, we cover fixed-point quantization, multiple binary codes quantizations, and exponential quantization. We also study whether the selection of the quantized value is deterministic or stochastic. In deterministic methods, the selection is based on static thresholds. In contrast, selection in stochastic methods relies on probabilities and random numbers. Relying on random numbers is more difficult for hardware.  

\paragraph{\textbf{Quantized values representation}}
There are different methods for representing quantized values. In the following, we explain three commonly used methods.
\begin{enumerate}
    
    \item{\textbf{Fixed-point quantization}}
In this quantization method, the 32-bit floating-point values are quantized into a fixed-point representation notated as $Q_{m,f}$, where $m$ is the number of integer bits, and $f$ is the number of fractional bits. The total number of bits required is $k$. The sign bit may be included in the number of integer bits~\cite{quant-Bengio} or added as an extra bit added to $m$ and $f$~\cite{HW-FPGA-FINN-L-Xilinx}. For example, in the first case~\cite{quant-Bengio}, $Q_{1.1}$ is used to represent 2 bits fixed-point that has three values \{$-$0.5,0,0.5\}. This quantization method is also called Pow2-ternarization~\cite{RNN-low-pow2ter}. Usually, fixed-point quantization is deterministic, in that for each floating-point value, there is one quantized fixed-point value defined by an equation (i.e. it is rule-based). Fixed-point quantization is performed by clipping the floating-point value between minimum and the maximum boundaries, and then rounding it.% as performed in the Eq.(~\ref{eq:pow2ter1}) and Eq.(~\ref{eq:pow2ter2}).

% by applying the equations
%\begin{equation}
 %       x_{clip}=
  %      \left\{ \begin{array}{ll}
   %         -2^m & \text{if } x\leq -2^m \\
    %         x & \text{if }   -2^m < x < 2^m \\
     %         2^m & \text{if } x\geq 2^m \\
           
      %  \end{array} \right.
      %  \label{eq:pow2ter1}
    %\end{equation}
%    and
 %   \begin{equation}
  %    x_{quantized}=round(2^fx_{clip})2^{-f}.
   %     \label{eq:pow2ter2}
%    \end{equation}

\item{\textbf{Exponential quantization}}
Exponential quantization quantizes a value into an integer power of two. Exponential quantization is very beneficial for the hardware as multiplying with exponentially quantized value is equivalent to shift operations if the second operand is a fixed-point value, and addition to exponent if the second operand is a floating-point value~\cite{quant-Bengio,HW-ASIC-circular}. Exponential quantization can be both deterministic and stochastic. 

\item{\textbf{Binary and multi-bit codes quantization}}
The lowest precision in RNNs is binary precision~\cite{lowp-binarized-runtime}. Each full precision value is quantized into one of two values. The most common two values are \{$-$1, $+$1\}, but it can also be \{0, $+$1\}, \{$-$0.5, 0\}, \{$-$0.5, $+$0.5\}, or any combination of two values~\cite{quant-Bengio}.
Binarization can be deterministic or stochastic. For deterministic binarization, a sign function can be used for binarization. For stochastic binarization, selection thresholds depend on probabilities to compute the quantized value   
\begin{equation}
        x^b= 
        \left\{ \begin{array}{ll}
            +1 & \text{with probability }p = \sigma_h(x)\text{,} \\
            -1 &\text{with probability }1-p\text{,} 
        \end{array} \right.
        \label{eq:sign}
    \end{equation}
 where $\sigma_h$ is the ``hard sigmoid'' function defined as
 \begin{equation}
 \sigma_h(x) = clip(\frac{x+1}{2}, 0, 1)= max(0,min(1,\frac{x+1}{2})).
 \end{equation}
 
Binarization has great value for hardware computation as it turns multiplication into addition and subtraction. The greatest value comes with full binarization, where both the weights and the activations have binary precision. In this case, it is possible to concatenate weights and activations into 32-bit operands and do multiple MAC operations using XNOR and bit-count operations. Full binarization can reduce memory requirements by a factor of 32 and decrease computation time considerably~\cite{CNN-binary-xnor}.

Adding one more value to binary precision is called \textbf{ternarization}. Weights in ternarized NN are restricted to three values. These three values can be \{$-$1, 0, 1\}~\cite{CNN-ternary-China}. Power two ternarization is discussed above as a form of fixed-point quantization, and is an example of ternarization with three different values \{$-$0.5, 0, 0.5\}. Both deterministic and stochastic ternarization have been applied to RNNs~\cite{quant-Bengio}. 

Having four possible quantization values is called \textbf{Quaternarization}. In quaternarization, the possible values can be \{$-$1, $-$0.5, +0.5, +1\}~\cite{RNN-lowp-Taha}. 
In order to benefit from the high computational benefit of having binary weights and activations while using a higher number of bits, \textbf{multiple binary codes} \{$-$1,+1\} has been used for quantization~\cite{RNN-lowp-china}. For example, two bit quantization has four possible values \{\{$-$1,$-$1\}, \{$-$1,1\}, \{1,$-$1\}, \{1,1\}\}. 

The most common method for deterministic quantization is uniform quantization. Uniform quantization may not be the best quantization method as it can change the distribution of the original data, especially for non-uniform data, which can affect accuracy. One solution is balanced quantization~\cite{lowp-balanced}. In balanced quantization, data is divided into groups of the same amount of data before quantization to ensure a balanced distribution of data following quantization. Other suggested solutions treat quantization as an optimization problem, and include greedy quantization, refined greedy quantization, and alternating multi-bit quantization \cite{lowp-greedy,RNN-lowp-china}.

%\item{\textbf{Vector quantization}}
%\todo[inline]{To be added: Think of postponing after reviews, not common on RNNs}
%\cite{HW-ASIC-EIE}
\end{enumerate}
\paragraph{\textbf{Training/Retraining}}
As mentioned earlier, there are three options to minimize accuracy loss due to quantization. The first is to apply quantization with training~\cite{lowp-binary-connect}, where quantized weights are used during the forward and backward propagation only. Full precision weights are used for the parameters update step in the (Stochastic Gradient Descent) SGD. Copies for both quantized and full precision weights are kept to decide at inference time which one to use~\cite{quant-Bengio}. In the second approach, quantization is applied to pretrained parameters and the RNN model is retrained to decrease the accuracy loss. Also, binarization of LSTM gate outputs during training have been applied by using the GumbelSoftmax estimator~\cite{optimize-binarize-train-Li}. 
Authors in one RNN implementation~\cite{HW-FPGA-FINN-L-Xilinx} adopted a mix of training and retraining approaches, where only the activations were not quantized from the beginning. Activations were quantized after training and then the model was retrained for 40 epochs. 
The third approach is to use quantized parameters without training/retraining. This is very commonly used with 16-bit fixed-point quantization. Usually, training happens at training servers and quantization is applied at the inference platform without having the opportunity to retrain the model. It is very common as well to use 16-bit fixed-point quantization with other optimization techniques such as circulant matrices compression~\cite{HW-FPGA-CLSTM}, pruning~\cite{HW-FPGA-baltimore}, and deltaRNN (discussed later in Section~\ref{subsubsec:DRNN})~\cite{HW-FPGA-delta}.  
%Either applying quantization with training, re-training pre-trained model for a number of epochs after applying quantization or doing nothing regarding training and use the quantized parameters for inference directly.
%\begin{itemize}
%\item{\textbf{Quantization with training}}

%In this approach, the parameters are quantized during training. To train the RNN using quantized parameters, the following approach is adopted \cite{lowp-binary-connect}. 

%\item{\textbf{Retraining quantized models}}

%In this approach,

%\item{\textbf{Using quantized parameters without training}}

%\end{itemize}

\begin{table*}[tb]
\centering
\begin{threeparttable}
\caption{Effect of quantization methods on accuracy.}
\begin{tabular}{|p{1.75cm}||p{1.7cm}|p{2.25cm}|p{2.2cm}|p{2cm}|p{1.2cm}|p{0.75cm}|}
\hline

Method&W/A&RNN type&Dataset&Training&Accuracy&Paper\\ 
\hline \hline
% Fixed point
%94.2,95.3 vs 94
% accuracy got from higher point in curve
\multirow{2}{*}{Fixed Point}&2/2
&1*BiLSTM*128 &OCR dataset &With training&%Higher
$+$0.7\% &\cite{HW-FPGA-FINN-L-Xilinx} %2018
\\ \cline{2-7} 

%&4/8
%&1*BiLSTM*128 &OCR dataset &With training&%Higher
%$+$1.4\%&\cite{HW-FPGA-FINN-L-Xilinx} %2018
%\\ \cline{2-7}

&P2T/real &4*BiLSTM*250&WSJ&With training&%WER: 11.16-$>$  10.49
$+$6\% &\cite{quant-Bengio}% 2017
\\ \hline \hline 

%99.18 vs98.1
%&P2T/real &1*GRU*200&TIDIGITS&With training&%Higher 
%$+$1.1\%&\cite{quant-Bengio}% 2017
%\\ \hline \hline
 
 %99.1 vs 98.1
 %Exp
Exponential&EQ/real&1*GRU*200&TIDIGITS&With training&%Higher
$+$1\%&\cite{quant-Bengio}%2017 
\\ \hline \hline
%Mixed
%accuracy got from last row in the tables
%Mixed%&EQ+fixed6 /fixed8 &2*LSTM*128&PTB&Retraining&%Negligible
%$-$0.6\%\tnote{1}&\cite{HW-ASIC-circular}% 2017
%\\ \cline{2-7}

Mixed&EQ+ fixed6/8 &3*BiLSTM*512&AN4&Retraining&$+$10.7\%\tnote{1}%*WER: 26.46/$>$ 23.64
&\cite{HW-ASIC-circular}% 2017
\\ \hline\
% Binary
\multirow{4}{*}{Binary}
& B/real&1*GRU*128&IMDB&With training&%80.35-$>$76.12
$-$5.3\%&\cite{RNN-lowp-Taha} %2018
\\ \cline{2-7} 

%& B/real&1*LSTM*128&IMDB&With training&%82.87-$>$76.25
%$-$8\%&\cite{RNN-lowp-Taha} %2018
%\\ \cline{2-7} 

&B/real&ConvLSTM&Moving MNIST&With training&$-100\%$\tnote{2}&\cite{RNN-lowp-Taha} %2018
\\ \cline{2-7} 

%accuracy put for activation 1 and 8 bits
&B/1&1*BiLSTM*128 &OCR dataset &With training&%Lower
$-$3.7\%&\cite{HW-FPGA-FINN-L-Xilinx} %2018
\\ \cline{2-7} 
&B/4&1*BiLSTM*128 &OCR dataset &With training&%Lower
$+$1\%&\cite{HW-FPGA-FINN-L-Xilinx} %2018
\\ \cline{2-7} 

&B/real&1*GRU*200/400&TDIGITS&With training&%Failed
$-$80.9\%&\cite{quant-Bengio}% 2017
\\ \hline \hline

%Ternary

\multirow{5}{*}{Ternary}& T/real&1*GRU*128&IMDB&With training&%80.35-$>$77.12
$-$4\%&\cite{RNN-lowp-Taha} %2018
\\ \cline{2-7} 

%& T/real&1*LSTM*128&IMDB&With training&%82.87-$>$76.86
%$-$7.3\%&\cite{RNN-lowp-Taha} %2018
%\\ \cline{2-7} 
%1.5MSE
&T/real&ConvLSTM&Moving MNIST&With training&$-50\%$\tnote{2}&\cite{RNN-lowp-Taha}% 2018
\\ \cline{2-7} 

&T/real&1*GRU*200&TDIGITS&With training&%98.1-$>$ 96/96.6
$-$1.6\%&\cite{quant-Bengio} %2017
\\ \hline \hline

%Quaternary
\multirow{2}{*}{Quaternary}
& Q/real&1*GRU*128&IMDB&With training&%80.35-$>$78.96
$-$1.7\%&\cite{RNN-lowp-Taha} %2018
\\ \cline{2-7} 

%& Q/real&1*LSTM*128&IMDB&With training&%82.87-$>$79.64
%$-$3.9\%&\cite{RNN-lowp-Taha} %2018
%\\ \cline{2-7} 

%1.75MSE
&Q/real&ConvLSTM&Moving MNIST&With training&$-75\%$\tnote{2}&\cite{RNN-lowp-Taha} %2018
\\ \hline \hline

%Multiple Binary
% original ppw was 110.1 for lstm and 106.7 for GRU
%ppw 98.7 vs 100.1
\multirow{3}{*}{Multi-Binary}&3/3&
1*LSTM*512%/1*LSTM*1024
&%PTB/
WikiText2%/ Text8
&Retraining&$+1.4\%$& \cite{RNN-lowp-china} %2018
\\  \cline{2-7} 

%1*LSTM*300 /
%ppw 106.4 vs 106.7
%&3/3&%1*GRU*300 /
%1*GRU*512 %/1*GRU*1024
%&WikiText2&Retraining&$+0.3\%$& %\cite{RNN-lowp-china} %2018
%\\  \cline{2-7} 
%ppw 106.1 vs 100.1
&2/2&%1*LSTM*300 /
1*LSTM*512 %/1*LSTM*1024
&WikiText2&Retraining&$-6\%$& \cite{RNN-lowp-china} %2018 
\\  \cline{2-7} 

&1/4&2*LSTM*256&PTB&With training&$-7.8\%$&\cite{HW-ASIC-rram-xnor}
\\  \hline

%ppw 113.7 vs 106.7
%& 2/2&%1*GRU*300 /
%1*GRU*512% /1*GRU*1024
%& WikiText2&Retraining&$-6.6\%$& \cite{RNN-lowp-china} %2018
%\\  \hline 

\end{tabular}
\label{tab:acc-lowP}%
\begin{tablenotes}
\item[1] Accuracy is also affected by the compression scheme and nonlinear functions approximation used in this work.
\item[2] We calculate the error at the tenth frame (third predicted frame).
\item In the table we have used the symbols: W/A for number of bits for weights/number of bits for activations, P2T for power two ternarization, EQ for exponential quantization, B for binary quantization, T for ternary quantization, and Q for quaternary quantization.
\end{tablenotes}
\end{threeparttable}
\end{table*}	
%\tablefootnote{*Accuracy is also affected by the compression scheme used in this work.}\paragraph{\textbf{Effect on accuracy}}

\paragraph{\textbf{Effect on accuracy}}
In Table~\ref{tab:acc-lowP}, we list studies that included experiments on the quantization of RNN models. Not all of the studies have a hardware implementation, as the purpose is to show that quantization can be performed while keeping accuracy high. In the table, we put the three factors affecting the accuracy discussed earlier (number of bits, quantization method, and training) with an addition of the type of recurrent layer (LSTM, GRU...) and the dataset. Then, we show the effect of quantization on accuracy computed with respect to the accuracy achieved by full precision parameters and activation using Eq.~\ref{eq:accuracy}. For the number of bits, we use W/A where W is the number of bits used for weights and A is the number of bits used for activations. For the RNN type, we put the recurrent layers used in the experiments. All recurrent layers are explained in Section~\ref{Sec:RNNs}. We use x*y*z, where x is the number of layers, y is the type of the layers, and z is the number of hidden cells in each layer. For training, if quantization was applied with training from the beginning, we write ``With training''. If quantization was applied after training and the model was later retrained, we write ``Retraining''. Positive values for accuracy means that quantization enhanced the accuracy and negative values for accuracy means that quantization caused the model to be less accurate.

Each experiment in Table~\ref{tab:acc-lowP} is applied to a different model, different dataset, and may also have used different training methods. Thus, conclusions about accuracy from Table~\ref{tab:acc-lowP} cannot be generalized. Still, we can make some observations:
\begin{itemize}
    \item Fixed point quantization, exponential quantization and mixed quantization have no negative effect on accuracy. Accuracy increased after applying these quantization methods. Quantized models can surpass baseline models in accuracy as weight quantization has a regularization effect that overcomes over-fitting~\cite{HW-FPGA-FINN-L-Xilinx}.
    \item Regarding binary quantization, the negative effect on accuracy varied within small ranges in some experiments~\cite{RNN-lowp-Taha,HW-FPGA-FINN-L-Xilinx}. Experiments showed that using more bits for activations may enhance the accuracy~\cite{HW-FPGA-FINN-L-Xilinx}. Using binary weights with convLSTM is not solely responsible for the poor accuracy obtained, as Ternary and Quaternary quantization resulted in poor accuracy with convLSTM as well~\cite{RNN-lowp-Taha}. However, these quantization methods were successful when applied on LSTM and GRU in the same work~\cite{RNN-lowp-Taha}.
\end{itemize}
\subsubsection {\textbf {Compression}}
Compression decreases the model size by decreasing the number of parameters or connections. As the number of parameters is reduced, memory requirements and the number of computations decrease. Table~\ref{tab:prune} compares different compression methods. The compression ratio shows the ratio between the number of parameters of models before and after applying compression methods. Accuracy degradation is computed using Eq.~\ref{eq:accuracy}. 
\begin{enumerate}[(i)]
    
\item\textbf{Pruning} Pruning is the process of eliminating redundancy. Computations in RNNs are mainly dense matrix operations. To improve computation time, dense matrices are transformed into sparse matrices, which affects accuracy. However, careful choice of the method used to transform a dense matrix to a sparse matrix may result in only a limited impact on accuracy while providing significant gains in computation time. Reduction in memory footprint along with computation optimization is essential for making RNNs viable. However, pruning results in two undesirable effects. The first is a loss in the regularity of memory organization due to sparsification of the dense matrix, and the second is a loss of accuracy on account of the removal of weights and nodes from the model under consideration. The transformation from a regular matrix computation to an irregular application often results in the use of additional hardware and computation time to manage data. To compensate for the loss of accuracy caused by pruning, various methods, including retraining, have been applied. The following sections describe methods of pruning and compensation techniques found in the literature.  Table~\ref{tab:prune} summarizes the methods of pruning and its impact on sparsity and accuracy. Sparsity in this context refers to the number of empty entries in the matrices. In Table~\ref{tab:prune}, sparsity indicates the impact on the number of entries eliminated because of the method of pruning used. Within RNNs, pruning can be classified as either magnitude pruning for weight matrix sparsification, or structure-based pruning.

\textbf{Magnitude pruning} Magnitude pruning relies on eliminating all weight values below a certain threshold. In this method, the choice of threshold is crucial to minimize the negative impact on accuracy. Magnitude pruning is primarily based on identifying the correct threshold for pruning weights. 

\begin{itemize}
\item{\textbf{Weight Sub-groups}} For weight matrix sparsification, the RNN model is trained to eliminate redundant weights and only retain weights that are necessary. There are three categories to create weight subgroups to select the pruning threshold~\cite{See:2016}. These three categories are class-blind, class-uniform, and class-distribution. In class-blind, $x\%$ of weights with the lowest magnitude are pruned, regardless (blind) of the class.  In class-uniform, lower pruning $x\%$ of weights is uniformly performed in all classes. In class-distribution, weights within the standard deviation of that class are pruned.

\item{\textbf{Hard thresholding}}~\cite{HW-ASIC-DATE, HW-FPGA-ESE} identifies the correct threshold value that preserves accuracy. ESE~\cite{HW-FPGA-ESE} uses hard thresholding during training to learn which weights contribute to prediction accuracy.

\item{\textbf{Gradual thresholding}} This method~\cite{Narang:2017} uses a set of weight masks and a monotonically increasing threshold.  Each weight is multiplied with its corresponding mask. This process is iterative, and the masks are updated by setting all parameters that are lower than the threshold to zero. As a result, this technique gradually prunes weights introduced within the training process, in contrast to hard thresholding.

\item{\textbf{Block Pruning}}
In block pruning~\cite{narangblock:2017}, magnitude thresholding is applied to blocks of a matrix instead of individual weights during training. The weight with the maximum magnitude is used as a representative for the entire block. If the representative weight is below the current threshold, all the elements in the blocks are set to zero. As a result, block sparsification mitigates the indexing overhead, irregular memory accesses, and incompatibility with array-data-paths that characterises unstructured random pruning.

\item{\textbf{Grow and prune}}
Grow and prune~\cite{Dai:2018} combines gradient-based growth~\cite{Dai:2017} and magnitude-based pruning~\cite{HW-FPGA-ESE} of connections. The training starts with a randomly initialized seed architecture. Next, in the growth phase, new connections, neurons, and feature maps are added based on the average gradient over the entire training set. Once the required accuracy has been reached, redundant connections and neurons are eliminated based on magnitude pruning. 
\end{itemize}

\textbf{Structure pruning} Modifying the structure of the network by eliminating nodes or connections is termed structure pruning. Connections that may be important are learned in the training phase or pruned using probability-based techniques.  

\begin{itemize}

\item{\textbf{Network sparsification}} 
Pruning through network sparsification~\cite{HW-FPGA-timeconstr} introduces sparsity for the connections at every neuron output, such that each output has the same number of inputs. Furthermore, an optimization strategy is formulated that replaces non-zero elements in each row with the highest absolute value. This step avoids any retraining, which may be compute-intensive and difficult in privacy critical applications. However, the impact of this method of pruning on accuracy has not been directly measured. Design space exploration over different levels of sparsity measures the quality of output and gives an indication of the relationship between the level of approximation and the application-level accuracy.

\item{\textbf{Drop-out}} DeepIoT~\cite{Yao:2017} compresses neural network structures into smaller dense matrices by finding the minimum number of non-redundant hidden elements without affecting the performance of the network. For LSTM networks, Bernoulli random probabilities are used for dropping out hidden dimensions used within the LSTM blocks.

\end{itemize}

\textbf{Retaining accuracy levels} Pruning alongside training and retraining has been employed to retain the accuracy levels of the pruned models. Retraining works on the pruned weights and/or pruned model until convergence to a specified level of accuracy is achieved. Pruning has shown a regularization effect on the retraining phase~\cite{See:2016}. The regularization effect might be the reason for outperforming baseline model accuracy. Another benefit for pruning which might be the reason for outperforming the baseline accuracy is that pruning allows the finding of a better local minimum. Pruning increases the loss function immediately, which results in further gradient descent.

%\renewcommand{\thefootnote}{\fnsymbol{footnote}}
%\footnote[1]{Dataset not mentioned}

\textbf{Handling irregularity in pruned matrices}
Pruning to maximize sparsity results in a loss in regularity (or structure) of memory organization due to sparsification of the original dense matrix. Pruning techniques that are architecture agnostic, mainly result in unstructured irregular sparse matrices. Methods such as \textbf{load balancing-aware pruning}~\cite{HW-FPGA-ESE} and \textbf{block pruning} (explained earlier within magnitude pruning)~\cite{narangblock:2017} have been applied to minimize these effects. Load balancing-aware pruning~\cite{HW-FPGA-ESE} works towards ensuring the same sparsity ratio among all the pruned sub-matrices, thereby achieving an even distribution of non-zero weights. These techniques introduce regularity in the sparse matrix to improve performance and avoid index tracking.

%\begin{table}[!h]
%\centering
%\caption{Fixed Point 16 bits}
%\begin{tabular}{|p{4cm}|p{4cm}|p{4cm}|}
%\hline
%Paper&Saving in memory&train/retrain/no-train\\ 
%\hline
%HW-FPGA-conv-LSTM-short \cite{HW-FPGA-conv-LSTM-short}&2x&retrain \\ \hline
%HW-FPGA-CLSTM \cite{HW-FPGA-CLSTM}\newline HW-FPGA-structured \cite{HW-FPGA-structured}&2x &no-train\\ \hline
%HW-FPGA-delta \cite{HW-FPGA-delta} &2x&no-train \\ \hline
%HW-FPGA-baltimore  \cite{HW-FPGA-baltimore}&2x&no-train \\ \hline
%HW-tool-FPGA-CA \cite{HW-tool-FPGA-CA}&2x& not their responsibility \\ \hline
%& &\\ \hline

%\end{tabular}
%\label{HW-compare}%
%\end{table}
%\item{Low precision}
%~\ref{acc-lowP}

%\end{itemize}

%\subsubsection{\textbf{Other optimizations}} In addition to quantization and pruning, other optimization techniques that reduce the memory and computation have been explored. For each optimization technique, we study the effect of this technique on accuracy and the benefit of using it in Table~\ref{tab:Alg-others} following the same format used in Table~\ref{tab:acc-lowP} and Table~\ref{tab:prune}. 

\item \textbf{Structured matrices}

\textbf{Circulant matrices}
A circulant matrix is a matrix in which each column (row) is a cyclic shift of the preceeding column (row)~\cite{HW-ASIC-circular}. It is considered as a special case of Toeplitz-like
matrices. The weight matrices are reorganized into circular matrices. The redundancy of values in the matrices reduces the space complexity of the weights matrices. For large matrices, circulant matrices can use nearly 4$\times$ less memory space.
The back-propagation algorithm is modified to allow training of the weights in the form of circulant matrices.
\textbf{Block-circulant matrices}
Instead of transforming the weight matrix into a circulant matrix, it is transformed into a set of circulant sub-matrices~\cite{HW-FPGA-CLSTM,HW-FPGA-ERNN}. Figure~\ref{fig:block circular} shows a weight matrix that has 32 parameters. The block size of the circular sub-matrices is 4. The weight matrix has transformed into two circulant sub-matrices with 8 parameters (4 parameters each). The compression ratio is 4$\times$, where 4 is the block size. Thus, having larger block sizes will result in a higher reduction in model size. However, a high compression
ratio may degrade the prediction accuracy.
 In addition, the Fast Fourier Transform (FFT) algorithm can be used to speed up the computations. Consequently, the computational complexity decreases by a factor of $\mathcal{O}(\frac{k}{\log{}k})$.
 
\begin{figure}[h]
   \centering
   \includegraphics[width=0.6\columnwidth]{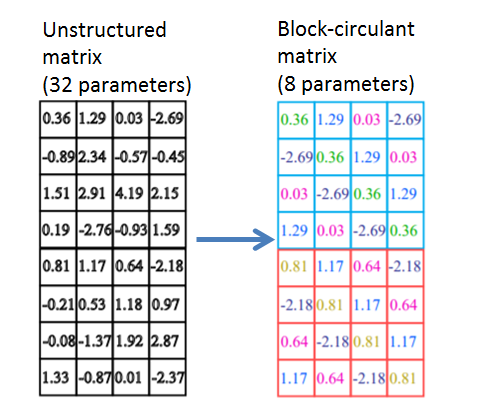}
    \caption{Regular weight matrix transformed into block-circulant sub-matrices of block size 4~\cite{HW-FPGA-CLSTM}.}
    \label{fig:block circular}
\end{figure}
%\begin{figure}[h]
 %  \centering
 %  \includegraphics[width=0.4\columnwidth]{clstm.png}
 %   \caption{Regular weight matrix transformed into block-circulant sub-matrices of block size 4~\cite{HW-FPGA-CLSTM}.}
  %  \label{fig:block circular}
%\end{figure}
%The modelThe model compression ratio is determined by the
%block size of the circulant submatrices: larger block size leads to
%higher compression ratio and vice versa. 
%Using block circulant matrices as a method of retaining regularity in the pruned matrix, the weight matrices are reorganized into blocks of circular matrices. Within each block, the use of a circular representation reuses a single row of the matrix with right shifted values for the rest of the matrix. Reusing a single row for the entire matrix reduces the storage needed for the weights.  \cite{HW-FPGA-CLSTM}.

\item{\textbf{Tensor decomposition}}
Tensors are multidimensional arrays. A vector is a tensor of rank one, and a 2-D matrix is a tensor of rank two and so on. Tensors can be decomposed into lower ranks tensors, and tensor operations can be approximated using these decompositions in order to decrease the number of parameters in the NN model. Canonical polyadic (CP) decomposition, Tucker decomposition, and tensor train decomposition are some of the techniques used to apply tensor decomposition~\cite{tensor-RNN}.
Tensor decomposition techniques can be applied to the FC layers~\cite{tensor-FC}, convolution layers~\cite{tensor-Conv}, and recurrent layers~\cite{tensor-RNN}. In Table~\ref{tab:prune}, we show an example of applying tensor decomposition on a GRU layer using the CP technique. In another example, Adam's algorithm has been used as an optimizer for the training process~\cite{train-Adam}. Tensor decomposition techniques can achieve a high compression ratio compared to other compression methods. 

%\item {\textbf{Weight sharing}}
%Weight sharing replaces each weight with an approximate obtained through k-means clustering. For instance, deep compression~\cite{HanMD:2015} uses Huffman coding with weight sharing to reduce the length of the weight indices. Huffman coding relies on using the occurrence probability of used weights, more common symbols are encoded with fewer bits. However, we did not find any work applying weight-sharing on RNNs.
\begin{landscape}
\begin{table}[!h]
\centering
\begin{threeparttable}
\caption{Effect of compression methods on accuracy. }
\begin{tabular}{|p{1.5cm}||p{3.25cm} |p{2.2cm}|p{1.8cm}|p{3cm}|p{2 cm}|p{1.8 cm}| p{0.75cm}|}
\hline\hline

Method  	                        & Technique	        & RNN Type			&Dataset    &Compression ratio \newline(Sparsity for pruning )		&Training      & Accuracy & Paper\\\hline\hline

\multirow{5}{*}{\parbox{1.7cm}{Magnitude pruning}}  & Weight subgroups& 4*LSTM*1024 + 4*LSTM*1024 & WMT'14 &5$\times$ (80\%)-10$\times (90\%)$& Retraining &+2.1\%-$-1.7$\% &\cite{See:2016}  \\ \cline{2-8}

& Hard thresholding     & 2*LSTM*512         &TIMIT           &1.1$\times$ (10\%) -1.3$\times$(24\%) &None          & 0\%   &\cite{HW-ASIC-DATE} \\\cline{2-8}
            
& Gradual pruning &2*LSTM*1500& PTB&20$\times$ ( 90\%)    &With training & $-2.3$\% &\cite{Zhu:2017} \\\cline{2-8}

& Block pruning & 7*BiLSTM*2560  & Speech Data\tnote{2}  &12.5$\times$ (92\%) & With training & $-12$\% &\cite{Narang:2017}\\ \cline{2-8}  

&Grow\&Prune&1*H-LSTM*512 \tnote{1}&COCO&8$\times$ (87.5\%) -19$\times$ (95\%)&With training&0\%-$-2.2$\%&\cite{Dai:2018}\\ \hline \hline

 %& & 1*GRU*3568   &  Speech Data\tnote{1}        & 2$\times (50\%)  $        & With training      & -2.2\% &  \\                                \hline \hline

\multirow{3}{*}{\parbox{1.7cm}{Structured pruning}}          &Network sparsification    &2*LSTM*512           & COCO & 2$\times$ (50\%) &None& 0\%&\cite{HW-FPGA-timeconstr}\\    \cline{2-8}  

&Drop-out & 5*BiLSTM*512& LibriSpeech ASR corpus     & 10$\times$ (90\%)    & None    & 0\%                         &\cite{Yao:2017}  \\ \hline \hline

%revised
\multirow{3}{*}{\parbox{1.7cm}{Structured matrices}}& Circulant  &3*BiLSTM*512&AN4\newline&nearly 4$\times$&With training&$+$10.7\%\tnote{3}&\cite{HW-ASIC-circular} \\ \cline{2-8}  

%revised
&Block-circulant  %&2*LSTM*2048&TIMIT&7.6$\times$&None& +0.3\%&\cite{HW-FPGA-CLSTM} \\ \cline{3-8} 
&2*LSTM*1024&TIMIT& 15.9$\times$&With training&-5.5\%&\cite{HW-FPGA-CLSTM} \\ \hline

Tensor decomp. & CP &1*GRU*512&Nottingham&101$\times$ - 481$\times$&With training&$-1$\% - $-5$\%&\cite{tensor-RNN} \\ \hline

\multirow{2}{*}{\parbox{1.7cm}{Knowledge \newline distillation}} &Plain&4*LSTM*1000&WMT'14&3$\times$&With training&$-1$\%&\cite{RNN-knowledgedistillation}\\    \cline{2-7} &+Pruning&4*LSTM*1000&WMT'14&26$\times$&With training + Retraining&$-5.1$\%& \\ \hline 

\end{tabular}
\label{tab:prune}%
\begin{tablenotes}
%\item[1] Sparsity is added for pruning only.
\item[1]H-LSTM is hidden LSTM. Non-linear layers are added in gate computations (Explained in Section \ref{Sec:RNNs}).
\item[2]Dataset name is not mentioned in the paper.
\item[3] Accuracy is also affected by quantization (Table~\ref{tab:acc-lowP}) and nonlinear functions approximation used in this work.

\end{tablenotes}
\end{threeparttable}
\end{table}

\begin{table}[!h]
\centering
\caption{Effect of DeltaRNN method on accuracy}
\begin{tabular}{|p{3cm}|p{2.5cm}|p{2.3cm}|p{1.5cm}|p{1.5cm}|p{0.75cm}|}
\hline
RNN model& Dataset &Training &Accuracy&Speedup& paper\\ 
\hline

1*GRU*512&TIDIGITs&With training&$-1.6\%$&5.7$\times$ &\cite{HW-FPGA-delta} \\ \hline

CNN+ 1*GRU*512&Open-driving&With training&0\%&100$\times$ &\cite{RNN-delta-method} \\ \hline

\end{tabular}
\label{tab:Alg-others}
%\begin{tablenotes}
%\end{tablenotes}
\end{table}	

\end{landscape}
\item{\textbf{Knowledge distillation}}
Knowledge distillation is a method that replaces a large model with a smaller model that should behave like a large model. Starting from a large model (teacher) with trained parameters and a dataset, the small model (student) is trained to behave like the large model~\cite{RNN-knowledgedistillation}. In addition to knowledge distillation, pruning can be applied to the resulted model to increase the compression ratio, as shown in Table~\ref{tab:prune}.

\end{enumerate}

\subsubsection{\textbf{DeltaRNN}}
\label{subsubsec:DRNN}
Delta Recurrent Neural Networks (DeltaRNN)~\cite{RNN-delta-method} makes use of the temporal relation between input sequences. For two consecutive input vectors $x_t$ and $x_{t-1}$, the difference between corresponding values in the two vectors may be zero or close to zero. The same holds for the hidden state output vector. The idea is to skip computations for input/hidden state values that when compared to input/hidden state values of the previous time step, have a difference that is less than a pre-defined threshold called delta ($\Theta$). Improvement comes from decreasing the number of computations and the number of memory accesses required by the recurrent unit. However, memory requirements will not decrease because we still need to store all the weights as we cannot predict which computations will be skipped.

%\begin{figure}[h]
 %  \centering
 %  \includegraphics[width=0.5\columnwidth]{ delta.PNG}
 %   \caption{Delta RNN computation saving in matrix to vector multiplication \cite{HW-FPGA-delta}. The vector $\Delta x $ is the difference between the two vectors $x_t$ and $x_{t-1}$. Matrix to vector multiplications are skipped for $\Delta x $ zero values. }
  %  \label{fig:deltafig}
%\end{figure}

The value of delta threshold affects both accuracy and speedup. In Table~\ref{tab:Alg-others}, we summarize the effect of DeltaRNN on accuracy for two different datasets. In some occasions, it was required to train the RNN using a delta algorithm before inference to obtain better accuracy at inference time.
Furthermore, the speedup gained by the delta algorithm at one delta value is not static. It depends on the relation between the input sequences. The highest speedup could be reached using video frames (open driving dataset) as input data, as seen in Table~\ref{tab:Alg-others}. However, the time-consuming CNN before the recurrent layer negated the speedup gained by deltaRNN. Thus, the 100x speedup in GRU execution dropped down to a non-significant speedup for the model as a whole.
On the other hand, CNN-Delta~\cite{CNN-Delta} applied a similar delta algorithm on CNNs. Applying delta algorithms to both recurrent layers and CNN layers might prove beneficial.

%\begin{table}[!h]
%\centering
%\caption{Other optimization techniques and impact on memory.}
%\begin{tabular}{p{0.8cm}|p{3.5cm}|p{1.4cm}|p{2cm}|p{2cm}|p{2cm}}
%\hline\hline
%Paper					&optimization Technique		&Dataset & RNN Model	&Compression						&Training   \\\hline\hline
%\cite{HW-FPGA-CLSTM} 	&Block circulant matrices &TIMIT	& 2*LSTM*2048 		& 4$\times$ 			&No retraining\\\hline
%\cite{HW-FPGA-structured} &Block circulant matrices&TIMIT&2*LSTM*1024 & upto 15.9$\times$&No retraining\\\hline
%\cite{Dai:2018}           & Grow and Prune & COCO& 3*LSTM*256  &38.7$\times$  & During Training  \\\hline
%\hline
%\end{tabular}
%\label{tab:other-opt}%
%\end{table}

\subsubsection{\textbf{Non-linear function approximation}}
Non-linear functions are the second most used operations in the RNN after matrix to vector multiplications, as may be seen in Table~\ref{tab:complstm}. The non-linear functions used in the recurrent layers are tanh and sigmoid, respectively. Both functions require floating-point division and exponential operations, which are expensive in terms of hardware resources. In order to have an efficient implementation for an RNN, non-linear function approximations are implemented in hardware. This approximation should satisfy a balance between high accuracy and low hardware cost. In what follows, we present the approximations used in the implementations under study.
  
  \textbf{Look-up tables (LUTs)}:
    Replacement of non-linear function computation with look-up tables is the fastest method~\cite{NN-nonlinear-LUT}. The input range is divided into segments with constant output values. However, to achieve high accuracy, large LUTs are required and that consumes a large area of silicon, which is not practical. Several methods have been proposed to decrease the LUTs size while preserving high accuracy.
    
    \textbf{ Piecewise linear approximation}:
    This approximation method is done by dividing the non-linear function curve into a number of line segments. Any line segment can be represented by only two values: the slope and the bias. Thus, for each segment, only two values are stored in the LUTs. The choice of the number of segments affects both accuracy and the size of LUTs. Thus, the choice of the number of segments must be made wisely to keep the accuracy high while keeping the LUTs as small as possible. The computational complexity of the non-linear function changes to be a single comparison, multiplication and addition, which may be implemented using shifts and additions. Compared to using look-up tables, piecewise linear approximation requires fewer LUTs and more computations.
        
  \textbf{Hard tanh / Hard sigmoid}:
Hard tanh and hard sigmoid are two examples of piecewise linear approximation with three segments. The first segment is saturation to zero or $-1$ (zero in case of sigmoid and $-1$ in case of tanh), the last segment is saturation to one, and the middle segment is a line segment that joins the two horizontal lines. %For, instance, the hardtanh approximation of tanh function is computed as

%\begin{equation}
 %       hardtanh(x)=
  %      \left\{ \begin{array}{ll}
   %         -1 & \text{if } x < -1 \\
    %         x & \text{if }   -1 \leq x \leq 1 \\
     %        1 & \text{if } x > 1 \text{.}\\ 
           
     %   \end{array} \right. 
     %   \label{eq:hardtanh}
    %\end{equation}
   There is a variant of piecewise linear approximation called piecewise non-linear approximation. The line segments are replaced by non-linear segments and the use of multipliers cannot be avoided as they can in the linear version. This made the linear approximation preferable in hardware design.
    
   \textbf{RALUT}
    One other method to reduce the size of the LUTs is to use RALUT (Range Addressable Look Up Tables)~\cite{ANN-nonlinear-ralut}. In RALUTs, each group of inputs is mapped into a single output.

\subsection{\textbf{Platform specific optimizations}}
\label{subsec:hw-optimize}

In this section, we discuss the optimizations performed on the hardware level to run an RNN model efficiently. These optimizations may be related to computation or memory. For computation-related optimizations, techniques are applied to speedup the computations and obtain higher throughput. For memory-related optimizations, techniques are applied to carry out memory usage and accesses with reduced memory overhead.

\subsubsection{\textbf{Compute-specific}}
The bottleneck in RNN computations is the matrix to vector multiplications. It is difficult to fully parallelize matrix to vector multiplications over time-steps as the RNN model includes a feedback part. Each time-step computation waits for the preceding time-step computations to complete so it can use the hidden state output as an input for the new time step computation.
\begin{itemize}
\item{\textbf{Loop unrolling}}
Loop unrolling is a parallelization technique that creates multiple instances of the looped operations to gain speedup at the expense of resources. There are two kinds of loop unrolling used in RNN implementations. The first is \textbf{inner loop unrolling}, where the inner loop of the matrix to vector multiplication is unrolled~\cite{HW-FPGA-cong,HW-FPGA-conv-LSTM-short}. The second kind is \textbf{unrolling over time-steps}. RNN needs to run for multiple time-steps for each task to be completed. The computation of the recurrent unit can be unrolled over time-steps~\cite{HW-FPGA-BiLSTM-OCR}. However, this cannot be fully parallelized, as discussed earlier. Only computations that rely on inputs can be parallelized, while computations relying on hidden state outputs are performed in sequence. One solution can be to use QRNN or SRU, as discussed in Section~\ref{subsec:RNN-types}. In QRNN and SRU, the matrix to vector multiplications do not operate on the hidden state output and thus can be fully parallelized over unrolled time steps \cite{HW-CPU-stream}.  
\item{\textbf{Systolic arrays}}
2D Systolic arrays are a good candidate for matrix to vector multiplication~\cite{HW-ASIC-fusion,HW-ASIC-MAERI} and convolution units~\cite{Chen:2016:ESA:3001136.3001177}. Systolic arrays are efficient as multiplications operands move locally between neighbor PEs (processing elements)~\cite{scale-sim}. Thus, systolic arrays require less area, less energy, and less control logic. Well designed systolic arrays can guarantee that PEs remain busy to maximize throughput. 

\item{\textbf{Pipelining}}
Pipelining is an implementation technique that can increase throughput. Pipelining has been used in RNN implementations in various ways. Coarse-grained pipelining (\textbf{CGPipe}) is used to tailor the LSTM/variants data dependency~\cite{HW-FPGA-ERNN,HW-FPGA-CLSTM}. LSTM computation is performed in three stages, with double buffers in between. The first stage is for weight matrices multiplications with inputs and hidden cells vectors, the second stage is for non matrix to vector operations, and the third stage is for projection layer computations. Fine-Grained Pipelining (\textbf{FGPipe}) can be used to schedule the operations within the CGPipe stages.
The design of the pipelining scheduler is a critical task due to the data dependency in LSTM/variants~\cite{HW-FPGA-ESE}. Some operations need to be performed sequentially, while some operations can be done concurrently. Having sparse weight matrices (due to applying pruning) increases the complexity of the scheduler design.

%\tem{\textbf{Systolic arrays}} \todo[inline]{New text}
%Eyeriss, TPU, MAERI
%Since most of NN operations are matrix to vector multiplications, 2D spatial arrays of processing elements are widely used in FPGAs and ASIC accelerators. 
%SIMD (Spatially distributed) 96000 MAC unit \cite{HW-FPGA-brainwave}.
\item{\textbf{Tiling}}
Tiling consists of dividing one matrix to vector multiplication into multiple matrix to vector multiplications. Usually, tiling is used when a hardware solution has built-in support for matrix to vector multiplication of a specific size in one clock cycle. When the input vector or the weight matrix size is larger than the size of the vector or the matrix supported by the hardware, tiling is used to divide the matrix to vector multiplication to be performed on the hardware in multiple cycles~\cite{HW-ASIC-circular,HW-FPGA-cong}. Thus, tiling can be combined with \textbf{Inner-loop unrolling} or \textbf{systolic arrays}. Figure~\ref{fig:tiling} shows a vector that is broken into three vectors and a matrix that is broken into nine matrices. Thus, one matrix to vector multiplication is broken into nine matrix to vector multiplications. Each vector is multiplied with the matrices having a similar color. The output vector is built from three vectors, where each of the three output vectors are accumulated together to form one vector in the output. This computation requires nine cycles to be completed, assuming that new weights can be loaded into the hardware multiplication unit within the cycle time.
\begin{figure}[h]
   \centering
   \includegraphics[width=0.6\columnwidth]{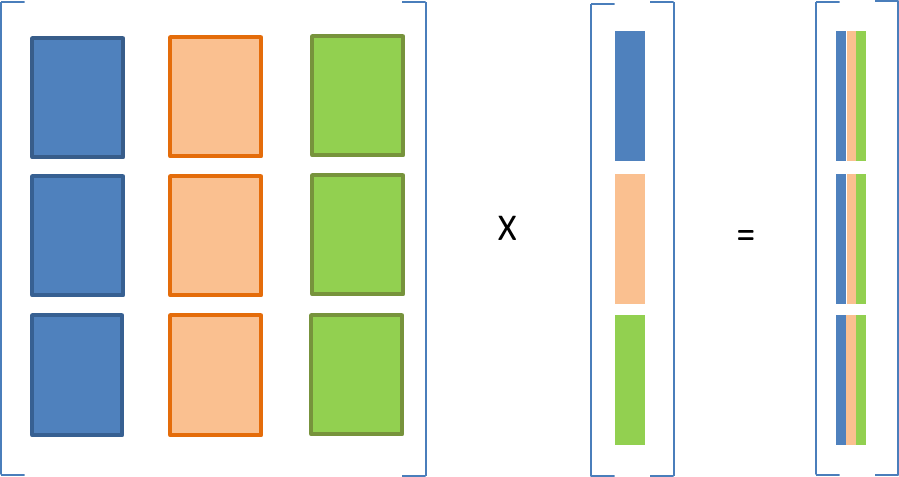}
    \caption{Tiling: converting one matrix to vector multiplication into nine matrix to vector multiplications.}
    \label{fig:tiling}
\end{figure}

%\todo[inline]{put a title or remove}
%Han \textit{\textit{et al.}} studied the possibility of concurrent execution of operations within one time step in their ESE architecture \cite{HW-FPGA-ESE}. For the matrix to vector operations and other operations that have no data dependency, the ESE controller schedules them to be executed concurrently. While, for operations that have data dependency or resource conflict, the scheduler executed them sequentially.
%\tem{\textbf {Pipelining}}
%\todo[inline]{New text}

\item{\textbf{Hardware sharing}}
In the GRU recurrent layer, the execution of $r_t$ and $\widetilde{h}_t$ has to be in sequence as $\widetilde{h}_t$ computation depends on $r_t$ as shown in Eq.~\ref{eq:gru-hcap}. Thus, the computation of $r_t$ and $\widetilde{h}_t$ is the critical path in the GRU computation. While $z_t$ can be computed in parallel as it is independent of $\widetilde{h}_t$ and $r_t$. The same hardware can be shared for computing $r_t$ and $z_t$ to save hardware resources~\cite{HW-ASIC-ocean}.

\item{\textbf{Load balancing}}
In the case of sparse weight matrices (resulting from pruning), load balancing techniques might be needed during parallelization of the matrix to vector multiplication over processing elements~\cite{HW-FPGA-ESE,HW-ASIC-DATE}.

\item{\textbf{Analog computing}}
Analog computing is a good candidate for neural network accelerators~\cite{HW-ASIC-analog}. Analog neural networks~\cite{NN-HW-analog} and analog CNNs~\cite{HW-CNN-analog} have been studied recently. Interestingly, RNN implementations using analog computing have started to attract attention from researchers~\cite{HW-ASIC-analog,HW-puma}. 
Analog computing brings significant benefits, especially for the critical matrix-vector computation, by making it both faster and more energy-efficient. This is true for the non-linear functions that normally are calculated between the NN layers as well. Analog computing also allows for more efficient communication as a wire can represent many values instead of only a binary value. The performance of an analog computer will however, critically depend on the digital to analog and analog to digital converters, for both speed and energy consumption.

\end{itemize}
\subsubsection{\textbf{Memory considerations}}
For the processing of an RNN algorithm, memory is needed to store weight matrices, biases, inputs, and activations, where the weight matrices have the highest memory requirement. The first decision related to memory is the location of weights storage. If all the weights are stored in the off-chip memory, accessing the weights comprises the largest cost with respect to both latency and energy~\cite{HW-FPGA-cong,HW-ASIC-EIE}.  

\textbf{On-chip memory}
 After applying the algorithmic optimizations introduced in Section~\ref{sec:algsection}, the memory requirements of the RNN layer are reduced, which increases the possibility of storing the weights in the on-chip memory. However, this results in a restriction on the largest model size that can run on the embedded platform. On-chip memory has been used for storing the weights by many implementations \cite{HW-FPGA-FINN-L-Xilinx,HW-FPGA-CLSTM,HW-FPGA-delta,HW-ASIC-circular,HW-FPGA-lowSpeech}.

\textbf{Hybrid memory}
Storing all the weights in the on-chip memory restricts the size of the model executed on the embedded solution. Storing some of the weights in on-chip memory with the remainder in off-chip memory might provide a solution~\cite{HW-FPGA-baltimore}.

In addition to maximizing the use of on-chip memory, some researchers use techniques to reduce the number and the cost of memory accesses.
\begin{itemize}
\item{\textbf{Multi time-step parallelization}}

The fact that QRNN and SRU remove the hidden state output from the matrix to vector multiplications can be leveraged to allow multi time-step parallelization~\cite{HW-CPU-stream}.
Multi time-step parallelization is performed by converting multiple matrix to vector multiplication into a fewer matrix to matrix multiplications. This method decreases the number of memory accesses by
reusing the weights for computations involving multiple time-steps.% To decrease, the cost of memory accesses Han \textit{\textit{et al.}} overlapped the computation and memory accesses \cite{HW-FPGA-ESE}. This can hide the memory access latency partially or totally.  

\item{\textbf{Reordering weights}} 
Reordering weights so they occupy memory in the same order as computation helps decrease the memory access time~\cite{HW-FPGA-cong}. Reordering the parameters in memory is carried out in a way that ensures memory accesses will be sequential.

\item{\textbf{Compute/load overlap}}
In order to compute matrix to vector multiplications, weights need to be accessed and loaded from memory and then used for computations. The total time is the sum of the access time and computation time. To decrease this time, memory access and computations can be overlapped. This overlap can be achieved by fetching the weights for the next time-step while performing the computation of the current time-step. The overlap would require the existence of extra buffers for storing the weights of the next time-step while using the weights of the current time-step~\cite{HW-FPGA-ESE}. %In some implementations, weights are stored in the on-chip memory and compute/load overlap is used while loading the input vectors~\cite{HW-FPGA-CLSTM}. 

\item{\textbf{Doubling memory fetching}}
In this method, twice the number of required weights for computation are fetched~\cite{HW-ASIC-laika}. Half of the weights are consumed in the current time step $t$ computations and the rest are buffered for the following time step $t+1$. Doubling memory fetching can reduce memory bandwidth by half.
\end{itemize}

\textbf{Domain-wall memory (DWM)}
DWM is a new technology for non-volatile memories proposed by Parkin \textit{\textit{et al.}} from IBM in 2008~\cite{Mem-DWM}. DWM technology is based on a magnetic spin~\cite{OPT-mem-RNNFast,DWM-1,DWM-2,DWM-3}. Information is stored by setting the spin orientation of magnetic domains in a nanoscopic permalloy wire. Multiple magnetic domains can occupy one wire which is called race-tracking. Race-tracking allows the representation of up to 64 bits. DWM density is hoped to improve SRAM by 30x and DRAM by 10x~\cite{DWM-4}. Using DWM in RNN accelerator can achieve better performance and lower energy consumption~\cite{OPT-mem-RNNFast}. 

\textbf{Processing In Memory (PIM)}
PIM gets rid of the data fetching problem by allowing computation to take place in memory, eliminating memory access overhead. In such an architecture, a memory bank is divided into three sub-array segments: memory sub-arrays, buffer sub-arrays, and processing sub-arrays, which are used as conventional memory, data buffer, and processing sub-arrays respectively. \textbf{ReRAM}-based PIM arrays is one approach used to accelerate CNNs~\cite{CNN-PIM-Prime,CNN-PIM-Pipelayer,CNN-PIM-3} and RNNs~\cite{HW-ASIC-PIM-1}. ReRAM that supports XNOR and bit counting operations will only be sufficient for RNN implementation if binary or multi-bit code (Section~\ref{subsub:quantize}) quantization has been applied~\cite{HW-ASIC-rram-xnor}. \textbf{Memristors} crossbar arrays have successfully been used as an analog dot product engine to accelerate both CNNs~\cite{HW-ISAAC} and RNNs~\cite{HW-puma}.

%\item{\textbf{Handling Sparse Matrix Parallelisation}}
%\cite{HW-ASIC-DATE} CBSR technique
%\todo[inline]{Shall this be removed to hw optimizations under title load balancing for sparse matrices?!}

%Flexibility and scalability: accelerators vs full model implementation 
%implementations restricted be on-chip memory size

%Activation functions are either implemented using LUTs or piece wise linear approximation. 

%--------------------
\section{RNN implementations on hardware}
\label{Sec:RNNonHW}
\label{HWsection}

%\todo[inline]{Review}
In the previous section, we discussed the optimizations applied to decrease the computation and memory requirements of RNN models. In this section, we study recent implementations of RNN applications on embedded platforms. The implementations are divided into FPGA, ASIC, and other implementations. We analyze these implementations and study the effects of the applied optimizations. However, the effect of each optimization is not shown separately. Instead, the outcomes of applying the mix of optimizations are discussed with respect to the objectives presented in Section~\ref{Sec:Challanges}.
 First, with regard to efficiency, the implementations are compared in terms of throughput, energy consumption, and meeting real-time requirements. Then, for flexibility, we discuss implementations that support variations in the models, online training, or different application domains.

\begin{table*}[!h]
\centering
\caption{Detailed information about papers under study}
\begin{tabular}{|p{1.25cm}||p{1.8cm}|p{1.5cm}|p{6.5cm}|p{1cm}|}
\hline
Index &Authors&Name&Affiliation&Year \\ \hline

F2 ~\cite{HW-FPGA-ERNN}&Li \textit{\textit{et al.}}&E-RNN&Syracuse University, Northeastern University, \newline Florida International University,\newline Mellon University, \newline Carnegie
University of Southern California,\newline SUNY University&2019 \\ \hline

F4~\cite{HW-FPGA-CLSTM}&Wang \textit{\textit{et al.}}&C-LSTM&Peking University, Syracuse University, \newline  City University of New York & 2018\\ \hline

F1~\cite{HW-FPGA-FINN-L-Xilinx} &Rybalkin  \textit{\textit{et al.}}&FINN-L&University of Kaiserslautern,\newline Xilinix Research Lab&2018 \\ \hline

F6~\cite{HW-FPGA-ESE} &Han \textit{\textit{et al.}}&ESE&Stanford University, DeePhi Tech, \newline Tsinghua University, NVIDIA&2017  \\ \hline

F3~\cite{HW-FPGA-delta} &Gao \textit{\textit{et al.}}&DeltaRNN&University of Zurich \& ETH Zurich&2018   \\ \hline

F5 ~\cite{HW-FPGA-BiLSTM-OCR}  &Rybalkin \textit{\textit{et al.}}&-&University of Kaiserslautern, \newline German Research Center for Artificial Intelligence&2017\\ \hline

F7 ~\cite{HW-FPGA-conv-LSTM-short}&Zhang \textit{\textit{et al.}}&& University of Illinois, Inspirit IoT Inc,\newline Tsinghua University, Beihang University&2017\\ \hline

F8 ~\cite{HW-FPGA-lowSpeech}&Lee \textit{\textit{et al.}}&-& Seoul National University&2016   \\ \hline

F9 ~\cite{HW-FPGA-flight} &Sun \textit{\textit{et al.}}&-& Shanghai Jiao Tong University,\newline Chinese Academy of Sciences,\newline University of Cambridge, Imperial College   &2018  \\ \hline

F10 ~\cite{HW-FPGA-cong}&Guan \textit{\textit{et al.}}&-&Peking University, University of California\newline PKU/UCLA Joint Research Institute in Science and Engineering &2017\\ \hline

F11 ~\cite{HW-FPGA-timeconstr}&Rizakis \textit{\textit{et al.}}&-&Imperial College London&2018  \\ \hline

F12 ~\cite{HW-FPGA-baltimore} &Chang \textit{\textit{et al.}}&DeepRnn&Purdue University&2017  \\ \hline

%F13 ~\cite{HW-FPGA-brainwave}& Fowers \textit{\textit{et al.}}&Brain wave& Microsoft& 2018 \\ \hline
A1~\cite{HW-puma}&Ankit \textit{\textit{et al.}}&PUMA& Purdue University, Hewlett Packard Enterprise,\newline University of Illinois at
Urbana-Champaign& 2019 \\ \hline

A2 ~\cite{HW-ASIC-circular}&Wang \textit{\textit{et al.}}&-&Nanjing University& 2017  \\ \hline

A3~\cite{HW-ASIC-analog}&Zhao \textit{\textit{et al.}}&-&Louisiana State University&2019 \\ \hline

%A2 ~\cite{HW-ASIC-EIE}&Han \textit{\textit{et al.}}&EIE&Stanford University, NVIDIA &2016\\ \hline
A8 ~\cite{HW-ASIC-PIM-1}&Long \textit{\textit{et al.}}&-&Georgia Institute of Technology, Atlanta&2018 \\ \hline

A4 ~\cite{HW-ASIC-ocean}&Chen \textit{\textit{et al.}}&Ocean&Fudan University, Zhejiang University, \newline  University of Washington&2017   \\ \hline

A5 ~\cite{HW-ASIC-DATE}&Park \textit{\textit{et al.}}&-&Pohang University of Science and Technology&2018   \\ \hline

A6 ~\cite{HW-ASIC-laika}&Giraldo \textit{\textit{et al.}}&Laika&KU Leuven&2018\\ \hline

A7 ~\cite{HW-ASIC-rram-xnor}&Yin \textit{\textit{et al.}}&-&Arizona State University&2018 \\ \hline
A9 ~\cite{HW-ASIC-MAERI}
 &Kwon \textit{\textit{et al.}}&MAERI&Goergia Institute of Technology&2018 \\ \hline

A10  ~\cite{HW-ASIC-fusion} &Sharma \textit{\textit{et al.}}&Bit Fusion&Goergia Institute of Technology, Arm Inc.  \newline University of California (San Diego)  &2018 \\ \hline

C1 ~\cite{HW-CPU-stream} \newline C2 ~\cite{HW-CPU-stream}&Sung \textit{\textit{et al.}}&-&Seoul National University&2018 \\ \hline

C3 ~\cite{HW-mobile-GPU-mobirnn}&Cao \textit{\textit{et al.}}&MobiRNN&Stony Brook University&2017 \\ \hline

\end{tabular}
\label{tab:ref-info}%
\end{table*}	

Table~\ref{tab:ref-info} shows the details of the implementations studied here. Authors names are shown, along with the name of the architecture; if named; the affiliation, and the year of publication. Table~\ref{tab:HW-compare-fpga} and Table~\ref{tab:HW-compare-others} present the implementations under study.  Table~\ref{tab:HW-compare-fpga} shows implementations performed on FPGAs, while Table~\ref{tab:HW-compare-others} shows implementations performed on other platforms. Each implementation has an index. The index starts with ``F'' for FPGA implementations, ``A'' for ASIC implementations, and ``C'' for other implementations.
For each implementation, the tables show the platform, the RNN model, the applied optimizations, and the runtime performance.

In most cases, only the recurrent layers of the RNN model are shown, as most of the papers provided the implementation for these layers only. The recurrent layers are written in the format x*y*z, where x is the number of recurrent layers, y is the type of recurrent layers (e.g. LSTM, GRU, ..), and z is the number of hidden cells in each layer. If the model has different modules (e.g. two different LSTM models or LSTM + CNN), we give the number of executed time-steps of the RNN model. Both algorithmic and platform optimizations are shown in the tables. 
All the optimizations found in the tables are explained above in Section~\ref{Sec:Optimizations} using the same keywords as in the tables. 
For quantized models, ``Quantization X'' is written in the optimizations column, where X is the number of bits used to store the weights. 
The effective throughput and the energy efficiency given in the tables are discussed in detail in the sub-section below.

\subsection{\textbf{Implementation efficiency}}
To study the efficiency of the implementations understudy, we focus on three aspects: throughput, energy consumption, and meeting real-time requirements. 
\subsubsection{\textbf{Effective Throughput}}
\label{subsub:throughput}
To compare the throughput of different implementations, we use the number of operations per second (OP/s) as a measure. Some of the papers surveyed did not directly state the throughput. For these papers, we have tried to deduce the throughput from other information given. 
One other aspect to consider is that compression optimization results in decreasing the number of operations in the model before running it. Consequently, the number of operations per second is not a fair indicator of the implementation efficiency. In this case, the throughput is calculated using the number of operations in the dense RNN model, not the compressed model. We call this an Effective Throughput. Below, we list the methods used to deduce the throughput values for the different papers.

\begin{itemize}
    \item Case q1: Effective throughput is given in the paper.
    
    \item Case q2: Number of operations in the dense model and computation time are given. By dividing number of operations $n_{op}$ by time, we get the effective throughput $Q_{eff}$ as shown in Eq.~\ref{eq:throughput}. In some papers, the number of operations and the computation time $time_{comp}$ were given for multiple time steps (multiple inputs), which would require running the LSTM $n_{steps}$ times. 
    
    \begin{equation}
    \label{eq:throughput}
     Q_{eff}= \frac{n_{op} \times n_{steps}}{t_{comp}}
    \end{equation}
    
    \item Case q3: The implemented RNN model information is provided in the paper. Thus, we calculate the number of operations from the model information and then divide it by computation time to get the throughput as in case q2. To compute the number of operations, the number of operations in the matrix to vector multiplications is counted as they have the dominant effect on the performance.
  If the paper does not give enough information to calculate the number of operations, the number of operations can be approximately calculated by multiplying the number of parameters by two.
    %For instance, if the model is built using LSTM layers, we use the equation
    %\begin{equation}
    %\label{eq:ops}
    % n_{op}= 2 \times 4 \times(n \times m + n^2),
%    \end{equation}
 %   where $n_{op}$ is the number of operations in an LSTM layer, the term between the brackets is the number of the matrix to vector multiplications in one gate ($m$ is the input vector size and $n$ is the number of hidden cells). This term is multiplied by four as the LSTM has four matrix to vector multiplications and multiplied by two to convert the matrix to vector multiplications into operations as each MAC operation in the matrix to vector multiplication is multiply then add (two operations). 
%    If the LSTM has a projection layer, the number of operations is calculated as
 %   \begin{equation}
 %   \label{eq:ops-proj}
 %    n_{op}= 2 \times 4 \times(n \times m + n \times p),
 %   \end{equation}

 %where the term $n^2$ is replaced by the term $n \times p$ ($p$ is the size of the projection layer).
 %Furthermore, if the recurrent layer is bidirectional, the number of operations is multiplied by two. 

    \item Case q4: The energy efficiency is given in terms of OP/s/watt and the power consumption is given in watt. By multiplying the two values, throughput is calculated.
    \item Case q5: Effective throughput could not be computed.

\end{itemize}

For a fair comparison between the ASIC implementations, we have applied scaling to 65~nm technology at 1.1~V using the general scaling equations in Rabaey~\cite{tech-scaling} and scaling estimate equations for 45~nm and smaller technologies~\cite{tech-stillmaker}. If the voltage value is not mentioned in the paper, we assume the standard voltage for the implementation technology. For example, since A7~was implemented on 65~nm, we assume the voltage value to be 1.1~V. 

To analyze Table~\ref{tab:HW-compare-fpga} and Table~\ref{tab:HW-compare-others} and understand the effect of different optimizations on throughput, the tables entries are ordered in descending order, starting with the highest throughput implementation. There exist two optimization groups that appear more frequently among the high throughput implementations. The first group is related to decreasing memory access time. Memory access time is decreased either by using on-chip memory for all weights or overlapping the computation time and the weights loading time.

The second group is related to algorithmic optimizations. Algorithmic optimizations present in all high throughput implementations are compression (pruning, block-circulant matrices, etc.), deltaRNN, and low precision quantization. Non-linear function approximations and 16-bit quantization are not within the groups of high effect optimizations. Quantization with 16-bit is present in many implementations that do not aim for lower precision, and it does not have a great effect on computation cost. Thus, it is not a differentiating factor. Non-linear function approximations do not contribute to the most used operations (matrix to vector multiplications). 

Finally, the throughput values are plotted against the implementations in Figure~\ref{fig:throughput}. The scaled effective throughput values for the ASIC implementations are used. Implementations that have memory access optimizations and/or algorithmic optimizations are highlighted by putting them inside a square and/or circle. It can be observed from Figure~\ref{fig:throughput} that all of the implementations with high throughput have some algorithmic optimization and applied memory access optimization. For example, F1~\cite{HW-FPGA-FINN-L-Xilinx} applied low precision quantization and placed all the weights on the on-chip memory.
F2~\cite{HW-FPGA-ERNN}, 
F3~\cite{HW-FPGA-delta},  F4~\cite{HW-FPGA-CLSTM}, and A2~\cite{HW-ASIC-circular}, all applied both on-chip memory optimization and algorithmic optimizations. In F6~\cite{HW-FPGA-ESE}, the architecture had a scheduler that overlapped computation with memory accesses. All the weights required for computation are fetched before the computation starts. Thus, they managed to eliminate the off-chip memory access overhead by having an efficient compute/load overlap. 

Both F2~and F4~applied block-circulant matrices optimization. In addition, A2~applied circulant matrices optimization. This indicates that restructuring weight matrices into circulant matrices and sub-matrices is one of the most fruitful optimizations. The reason could be that circulant matrices optimization does not cause the irregularity of weight matrices seen in pruning~\cite{HW-FPGA-CLSTM}. Additionally, circulant/block-circulant matrices can be accompanied by low precision quantization without a harsh effect on accuracy as in A2~(6-bit) and F2~(12-bit). It is observed in Table~\ref{tab:HW-compare-fpga} that F2~and F4~optimizations are almost identical, but their performance is different. F2~and F4~have differences in the hardware architecture and F2~applied lower precision than F4,~but the most important reason is that F2~used a better approach in training the compressed RNN model. F2~was able to reach the same accuracy level reached by F4~with block size 8 while using block size 16. Thus, the RNN model size in F2~is approximately 2x less than F4.

Nevertheless, it is noticed that the only compute-optimization in F2~and F4~is pipelining. In these two implementations, pipelining served in two roles. The first is coarse-grained pipelining between LSTM stages, and the second, fine-grained pipelining within each stage.
It worth knowing that F1~is based on the same architecture as F5. F1~achieved higher throughput than F6~ by applying higher frequency and using lower precision. Assuming linear frequency scaling, the ratio between the two implementations' throughput is close to the ratio between the precisions used for storing the weights by the two implementations.

The lack of algorithmic optimizations in A1~\cite{HW-puma} was compensated by the use of analog crossbar-based matrix to vector multiplication units. Analog crossbar units allowed low latency matrix to vector multiplications. Implementations that used analog computing are marked with an ``A'' sign in Figures~\ref{fig:throughput} and~\ref{fig:power}. Comparing A1 to A3, both were using analog crossbars. A1 surpassed A3 by applying PIM (Processing In Memory), which removes memory access overhead.  Therefore, in Figures~\ref{fig:throughput} and \ref{fig:power}, we consider PIM as a memory access optimization.

One implementation that stands out is A6~\cite{HW-ASIC-laika}, which has a very low throughput for an ASIC implementation while applying on-chip and algorithmic optimizations. This particular implementation was meant to meet a latency deadline of 16ms while consuming low power -- at the micro-watt level. Thus, high throughput was not the objective from the beginning. Implementations that defined real-time requirements are marked by an ``RT'' sign in Figures~\ref{fig:throughput} and~\ref{fig:power}. Another implementation that rewards close inspection is F8. Despite applying the two mentioned optimizations, it could not reach as high performance as expected. The conclusion here is that applying memory access optimization and algorithmic optimization is necessary but not sufficient for high performance.
\begin{landscape}
\begin{table}[]
\centering
\begin{threeparttable}
\caption{Comparison of RNNs implementations on FPGAs.}
\begin{tabular}{|p{1cm}||p{3.35cm}|p{2.8cm}|p{3.2cm}|p{3.5cm}|p{2cm}|p{1.75cm}|}
 \hline 
Index & Platform &  Model & Algorithmic\newline  Optimizations & Platform \newline Optimizations  & $Q_{eff}$\tnote{1} \newline GOP/s& $E_{eff}$\tnote{2} GOP/s/watt  \\ \hline \hline 

% PE unrolling in the paper is equivalent to unrolling over timesteps as each instance for LSTM takes different input vector
%Raybalkin \textit{\textit{et al.}}
 F1~\cite{HW-FPGA-FINN-L-Xilinx}&Zync XCZU7EV @266 MHz &1*BiLSTM*128&Quantization 1&On-chip,Pipelining \newline Inner-loop-unrolling \newline  Unrolling-timesteps, Tiling&$<q1>$ 3435 &$<e4>$ - \\ \hline
 
%Zhang \textit{\textit{et al.}} 
%\multirow{10}{*}{FPGA}&
F2 ~\cite{HW-FPGA-ERNN} & \text{Alpha Data ADM-7V3} @200MHz &2*LSTM*1024&Block-circulant  16  \newline Piecewise approx. \newline  Quantization 12 &On-chip \newline Pipelining %\newline Compute/load overlap  
%\newline Operator scheduling
&$<q3>$ 2485  &$<e1>$ 99.4 \\ \hline 

%Gao \textit{\textit{et al.}}
 F3~\cite{HW-FPGA-delta}&  Zync 7100 @125  MHz&1*GRU*256&DeltaRNN, RALUT  \newline Quantization 16  & On-chip \newline Pipelining&$<q1>$ 1198.3&$<e2>$164 \\ \hline

F4~\cite{HW-FPGA-CLSTM} & \text{Alpha Data ADM-7V3} @200MHz &2*LSTM*1024&Block-circulant  8 \newline Piecewise approx. \newline  Quantization 16 &On-chip \newline Pipelining %\newline Compute/load overlap %(not weights) %\newline Operator scheduling
&$<q3>$ 1167.3 &$<e1>$ 53 \\ \hline

%Wang \textit{\textit{et al.}} 

%Rybalkin \textit{\textit{et al.}}
F5 ~\cite{HW-FPGA-BiLSTM-OCR}&  Zynq-
7000 XC7Z045  @142 MHz&1*BiLSTM*100 &Quantization 5 &On-chip, Pipelining\newline  Inner-loop-unrolling  &$<q1>$ 308 &$<e3>$ 44  \\ \hline

%Han \textit{\textit{et al.}} 
F6~\cite{HW-FPGA-ESE}  &  XCKU060 @200 MHz&2*LSTM*1024&Pruning, Quantization 12\newline %lin. interpolation
&  Compute/load overlap \newline Pipelining, Load balancing \newline Inner-loop-unrolling%(weights) 
&$<q3>$ 78.6 \tnote{3} &$<e2>$ 1.9\\ \hline

F7 ~\cite{HW-FPGA-conv-LSTM-short} &   Virtex-7 VC709 @100 MHz&\text{AlexNet +} 15steps:1*LSTM*256 & Quantization 16 &   Inner-loop-unrolling\newline Reordering weights &$<q2>$ 36.25\tnote{4}&$<e4>$ -    \\ \hline

%Lee \textit{\textit{et al.}}
F8 ~\cite{HW-FPGA-lowSpeech}&  XC7Z045 @ 100 MHz&100steps:3*LSTM*256 \newline 3840steps:2*LSTM*256 &Quantization 6&On-chip \newline Inner-loop-unrolling&$<q3>$ 30\tnote{5}&$<e2>$ 5.4 \\ \hline

F9 ~\cite{HW-FPGA-flight}&VC707 @150 MHz&1*LSTM*10 + FC &Hard Sigmoid&Tiling,Inner-loop-unrolling\newline Pipelining&$<q1>$ 13.5&$<e4>$ - \\ \hline

%Guan \textit{\textit{et al.}} 
F10 ~\cite{HW-FPGA-cong}&  VC707 @150 MHz&3*LSTM*250&Piecewise approx.& Tiling, Inner-loop-unrolling\newline Pipelining, Reordering weights\newline  Compute/load overlap&$<q1>$ 7.3&$<e4>$ - \\ \hline

%Rizakis \textit{\textit{et al.}}
% ~\cite{HW-FPGA-timeconstr}&Xilinx Zynq ZC706 @100 MHz& 2 * LSTM*512&iterative low-rank compression \newline pruning&Inter-gate and Intra-gate Parallelism.&33&- \\ \hline
%Rizakis \textit{\textit{et al.}}
F11 ~\cite{HW-FPGA-timeconstr}& Zynq ZC706 @100 MHz& 2 * LSTM*512&Pruning & Tiling, Inner-loop-unrolling&$<q3>$ 1.55&$<e4>$ - \\ \hline

%Chang \textit{\textit{et al.}}
F12 ~\cite{HW-FPGA-baltimore}&\text{Zynq-7000 XC7Z045} @142MHz&2*LSTM*128&Quantization 16 \newline Piecewise approx.& Hybrid memory\newline Compute/load overlap&$<q4>$ 0.2&$<e1>$ 0.11 \\ \hline

%Wang \textit{\textit{et al.}} 

%F13 ~\cite{HW-FPGA-brainwave}&Startix 10 280 @250MHz &1*GRU*2816&Quantization 8&On-chip&$<q1>35920$&$<e4>-$ \\ \hline

% they could not apply more optmizations like optimizing tanh as the application demand for accuracy was high not like in speech recognition for instance. Also, this application requires small models.

\end{tabular}
\label{tab:HW-compare-fpga}%
\begin{tablenotes}
\item[1] The cases q1-q4 are explained in Section \ref{subsub:throughput}.
\item[2] The cases e1-e4 are explained in Section \ref{subsub:energy}.
\item[3]The effective throughput in the paper was computed on a bit basis. For a fair comparison, we recalculated the number of operations on an operand basis.
\item[4] The throughput is for running CNN and LSTM combined together.
\item[5] The number of time steps the model should run per second to reach real-time behavior is given. We computed the number of operations in the model and multiplied by the number of time steps in one second, then multiplied by the speedup gained over the real-time threshold to get the implementation throughput.
\end{tablenotes}
\end{threeparttable}

\end{table}	

\end{landscape}
\begin{landscape}
\begin{table}[]
\centering
\begin{threeparttable}
\caption{Comparison of RNNs implementations on ASIC and other platforms.}
\begin{tabular}{|p{1.25cm}|p{1cm}||p{2.5cm}|p{1.5cm}|p{2.5cm}|p{3.4cm}|p{2.5cm}|p{2.3cm}|}
\hline  
Category &Index & Platform & Model & Algorithmic\newline  Optimizations & Platform \newline Optimizations &  $Q_{eff}$\tnote{1} GOP/s\newline (original/scaled)\tnote{3}& $E_{eff}$\tnote{2} GOP/s/watt (original/scaled)\tnote{3}\\ \hline \hline

%power and area &62.5 &90.6
%Wang \textit{\textit{et al.}}
\multirow{9}{*}{ASIC}
&A1~\cite{HW-puma}& CMOS 32nm @1GHz&LSTM \tnote{8}&Quantization 16& Memristor PIM\newline Analog computing, Pipelining&$<q1>$52300/16000&$<e1>$837/250 \\ \cline{2-8}

%&1.01&30.77
&A2 ~\cite{HW-ASIC-circular}&TSMC 90nm\newline @600MHz \&1v & 1*LSTM*512 &Quantization 6\newline Circulant matrices\newline Piecewise approx.&On-chip\newline Tiling \newline Inner-loop-unrolling&$<q1>$ 2460/3406&$<e2>$ 2436/2787  \\ \cline{2-8}
%Han \textit{\textit{et al.}} 
 %&A2 ~\cite{HW-ASIC-EIE} &CMOS 45nm\newline @800MHz&1*LSTM*2400&Pruning \newline Weight sharing \newline Quantization 4&On-chip&$<q1>$ 3000 \tnote{4}&$<e2>$ 5000 
 
 %1.8v&0.46&
 &A3 ~\cite{HW-ASIC-analog}& CMOS 180nm &1*LSTM*16&Quantization 4& Analog computing \newline On-chip&$<q4>$473.3/1211&$<e1>$950/7044 \\ \cline{2-8}

%1.2v&0.156&
% no details about the GRU model found in paper
&A4 ~\cite{HW-ASIC-ocean}&CMOS 65nm\newline @400 MHz &GRU\tnote{7}&Quantization 16 \newline Piecewise approx.&On-chip \newline Hardware sharing, Pipelining &$<q1>$ 311.6&$<e1>$ 2000/2380\\ \cline{2-8}
%Park \textit{\textit{et al.}}

%2.4&-&
& A5~\cite{HW-ASIC-DATE}&CMOS 65nm \newline@200MHz  & 2*LSTM*512&Pruning & Load balancing \newline Inner-loop-unrolling&$<q3>$ 295&$<e3>$ 122.9\\ \cline{2-8}

%0.575v&$4.9 \mu$&
%Giraldo and Verhelst
% has various modles, we chose one model with 2 hidden layers not only one
&A6 ~\cite{HW-ASIC-laika}&CMOS 65nm \newline@239 KHz &2*LSTM*32&Quantization 4 \newline Piecewise approx. &On-chip\newline Doubling memory fetching&$<q2>$ 0.002 \tnote{4}&$<e2>$ 469.3/128 \\ \cline{2-8}  

%&A9 ~\cite{HW-ASIC-MAERI} &&&&&& \\ \cline{2-8} 
%&A10 ~\cite{HW-ASIC-fusion} &CMOS 45nm \newline @500 MHz&&&&& \\ \cline{2-8}

%&-&0.65
&A7 ~\cite{HW-ASIC-rram-xnor}&CMOS 65nm&1*LSTM*256&Quantization 1\tnote{6}&ReRAM PIM \newline Analog computing&$<q5>$ -&$<e1>$27000 % \\ \cline{2-8}

%&A8 ~\cite{HW-ASIC-PIM-1}&CMOS 28nm&1*LSTM*128 &Quantization 8&ReRAM PIM&$<q2>$525.6&$<e2>$876 

% it was not mentioned in the paper that they use 65 nm. I got this from their prior work paper
\\ \hline
%Shin \textit{\textit{et al.}}
%& ~\cite{HW-ASIC-DNPU-CNN-RNN}&&&& \\ \cline{2-8}
%Chen \textit{\textit{et al.}}

%Lane \textit{\textit{et al.}}
\multirow{3}{*}{Others}
&C1 ~\cite{HW-CPU-stream}&%-3930K 3.2GHz
 ARMv8 \newline@ 2GHz% 64-bit 2.0GHz
&1*SRU*1024& SRU& Unrolling-timesteps
 &$<q3>$ 22.3&$<e4>$ - \\  \cline{2-8}

%Sung \textit{\textit{et al.}}
 &C2 ~\cite{HW-CPU-stream}&Intel Core i7 \newline@ 3.2GHz%-3930K 3.2GHz
% 64-bit 2.0GHz
&1*SRU*1024&SRU& Unrolling-timesteps &$<q3>$19.2&$<e4>$ - \\  \cline{2-8}

 &C3 ~\cite{HW-mobile-GPU-mobirnn}&Adreno 330 GPU \newline@ 450 MHz&2*LSTM*32&-&RenderScript\tnote{5}&$<q3>$ 0.0011&$<e4>$ -\\ \hline

\end{tabular}
\label{tab:HW-compare-others}%
\begin{tablenotes}
\item[1] The cases q1-q4 are explained in Section \ref{subsub:throughput}.
\item[2] The cases e1-e4 are explained in Section \ref{subsub:energy}.
\item[3] Scaled to 65~nm at 1.1~volt using general scaling~\cite{tech-scaling} and scaling estimates for 45~nm and smaller technologies~\cite{tech-stillmaker}.
\item[4] The throughput is not high as the purpose was to reach very low power consumption while performing inference within 16ms.
%\item[4] The shown numbers are for running FC layers of a CNN as it reproduces throughput numbers for the LSTM layer experimented in the paper.
\item[5] RenderScript is a mobile-specific parallelization framework~\cite{link-render}.
\item[6] Quantization used 1 bit for weights and 4 bits for activations.
\item[7] A4~proposed a GRU core without providing specific model details.
\item[8] A1~did not specify which model achieved the provided throughput and energy efficiency.
\end{tablenotes}
\end{threeparttable}

\end{table}	

\end{landscape}
In addition, Figure~\ref{fig:throughput} shows that most of the ASIC implementations were not exceeding FPGA implementations in terms of throughput. We think the reason is that the ASIC implementations under study did not use the latest ASIC technologies, as shown in Table~\ref{tab:HW-compare-others}. 

For the low throughput implementations, Figure~\ref{fig:throughput} shows that some implementations did not apply either of the two optimizations (memory access and algorithmic), such as F9~\cite{HW-FPGA-flight} that had a strict accuracy constraint bounding the use of algorithmic optimizations and C3~\cite{HW-mobile-GPU-mobirnn}. In addition, some implementations applied only one of the two optimizations, including F11~\cite{HW-FPGA-timeconstr} and F12~\cite{HW-FPGA-baltimore}.

\subsubsection{\textbf{Energy efficiency}}
\label{subsub:energy}
To compare the implementations from the energy consumption perspective, we use the number of operations per second per watt as a measure. The last columns in Table~\ref{tab:HW-compare-fpga} and Table~\ref{tab:HW-compare-others} show the energy efficiency. Energy efficiency is calculated based on the dense model, not the sparse model, as for effective throughput. However, it was not possible to obtain values for energy efficiency for all implementations.
In some cases, the power consumption was not mentioned in the paper, while in others, the consumed power was not provided in a precise manner. For instance, the power of the whole FPGA board may be provided, which does not indicate how much power is used by the implementation with respect to the peripherals \cite{HW-FPGA-conv-LSTM-short,HW-FPGA-cong}.

\begin{figure*}[h]
   \centering
   \includegraphics[width=1.5\columnwidth]{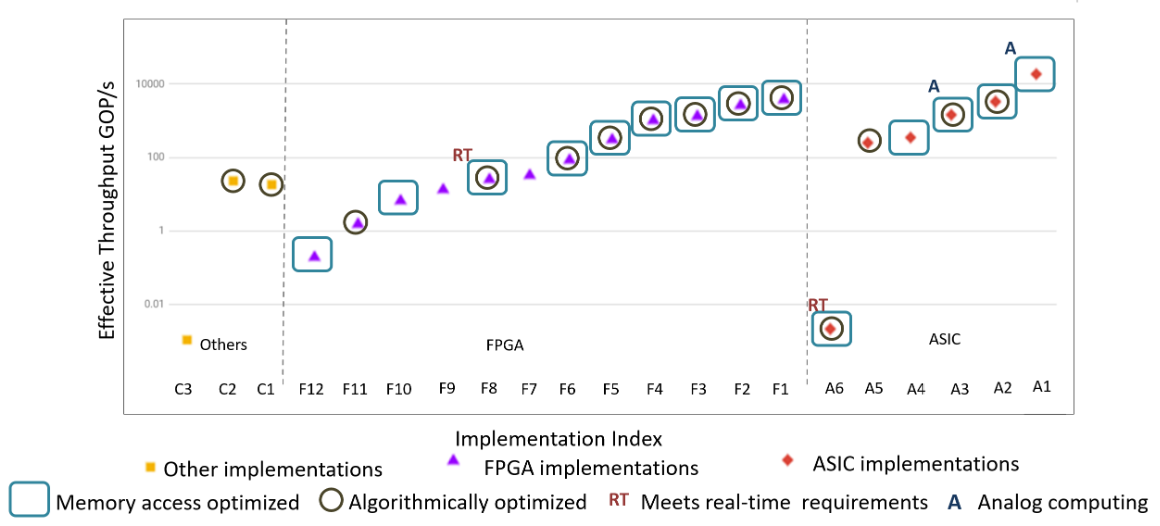}
    \caption{Effective throughput of different implementations and the key optimizations affecting them.}
    \label{fig:throughput}
\end{figure*}
\begin{figure*}[h]
   \centering
   \includegraphics[width=1.5\columnwidth]{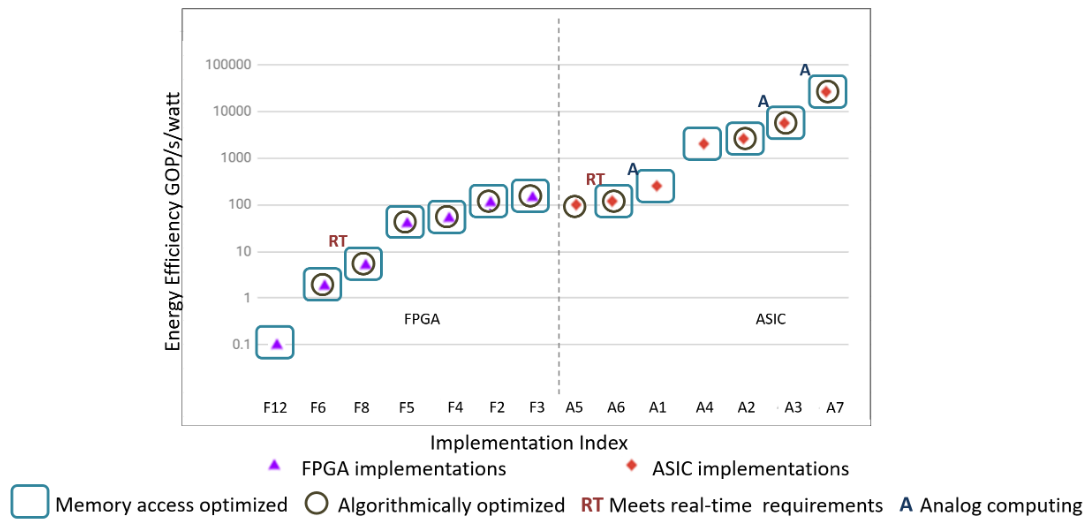}
    \caption{Energy efficiency of different implementations and the key optimizations used.}
    \label{fig:power}
\end{figure*}

Here, we list the cases used for computing
the energy efficiency in Table~\ref{tab:HW-compare-fpga} and Table~\ref{tab:HW-compare-others}.
The case number appears in triangular brackets, <>, before the numeric value
\begin{itemize}
    \item Case e1: The $E_{eff}$
    energy efficiency is given in the paper. 
    \item Case e2: The power consumption is given in the paper.  To compute the energy efficiency $E_{eff}$, the effective throughput $Q_{eff}$ (OP/s) is divided by the power $P$ (watt) as

    \begin{equation*}
    \label{eq:power}
     E_{eff}= \frac{Q_{eff}}{P}.
    \end{equation*}
    
    \item Case e3: Energy and computation time are provided. First, we divide energy by time to get power. Then, we divide effective throughput $Q_{eff}$ by the power to get energy efficiency, as in case e2.
    \item Case e4: energy efficiency could not be computed.

\end{itemize}

 Figure~\ref{fig:power} is a plot of the energy efficiency found or deduced for the implementations under study against the implementation index. Implementations are sorted in the plot according to energy efficiency and the scaled values for the ASIC implementations are used. Again, to show the effect of optimizations, we chose the two most effective optimizations from Table~\ref{tab:HW-compare-fpga} and Table~\ref{tab:HW-compare-others} to include in the figure. They are the same as in Figure~\ref{fig:throughput}: memory access optimization and algorithmic optimizations. % The observations from Figure~\ref{fig:power} agree with the observations from Figure~\ref{fig:throughput}. Algorithmic optimizations are applied in most of the efficient implementations and on-chip memory has been used for weights storage for most of the efficient implementations. 
Comparing the effective throughput and energy efficiency of FPGA and ASIC implementations, it is observed that FPGA and ASIC have close values for effective throughput while ASIC implementations are more energy efficient. The credit should go to ASIC technology.  

It can be observed that the highest energy efficiency was achieved by A7~\cite{HW-ASIC-reram} and A3~\cite{HW-ASIC-analog}. Both implementations used analog crossbar based matrix to vector multiplications. A7~managed to save memory access time by computing in memory. The quantization method used was multi-bit code quantization (1-bit for weights and 4-bit for activations). Multi-bit code quantization enables replacing the MAC operation with XNOR and bit-counting operations, as discussed in Section~\ref{subsub:quantize}. It was sufficient to use an XNOR-RRAM based architecture to implement the RNN.    

Both A1 (applying PIM and analog computing) and A6 (Applying memory and algorithmic optimizations) were less energy efficient than expected. They were both less energy efficient than A4 (Applying only memory optimization). A1 had a quite high clock frequency of 1~GHz. This high frequency helped the implementation to achieve high throughput. However, we suspect that this high frequency is the main reason for the energy efficiency degradation compared to the other implementations. A6 had the least power consumption among all implementations ($\leq 5~\mu W$). The low throughput of A6 affected the overall energy efficiency value.

\subsubsection{\textbf{Meeting real-time requirements}}
In some of the implementations, real-time requirements for throughput and power have been determined. For example, in F8~\cite{HW-FPGA-lowSpeech}, the speech recognition system had two RNN models. One model for acoustic modeling and the other for character-level language modeling. The real-time requirement was to run the first model 100 times per second and the second model 3,840 times per second. While in A6~\cite{HW-ASIC-laika}, an LSTM accelerator for an always-on Keyword Spotting System (KWS), the real-time response demanded that a new input vector should be consumed every 16 ms and the power consumption should not exceed $10~\mu W$.
\subsection{\textbf{Flexibility}}
Flexibility, as defined in Section~\ref{Sec:Challanges} is the ability of the solution to support different models and configurations. The flexibility of the solution can be met by supporting variations in the model. Models can vary in the number of layers, the number of hidden units per layer, optimizations applied on the model, and more. Flexibility can be met by supporting online training or meeting different application domain requirements.%

 Flexibility is not quantitative, like throughput. Thus, we use a subjective measure for flexibility to reach a flexibility score for each implementation. Table~\ref{tab:variations} shows the flexibility aspects supported by each implementation, as discussed in the papers and the flexibility score for each implementation. Papers that do not discuss any flexibility aspects are omitted from Table~\ref{tab:variations}. In A4~\cite{HW-ASIC-ocean}, the architecture should be able to support various models, but the number of cells and layers the architecture can support are not mentioned in the paper. Hence, we cannot deduce how the implementation could support variations in the RNN model. Also, the variations should be supported on the hardware platform and not only by the method before fabrication. In A2~\cite{HW-ASIC-circular}, the design method can support two different RNN layers. However, the fabricated chip supports only one of them. Thus, we do not consider A2~\cite{HW-ASIC-circular} to meet the flexibility objective. 
\begin{table}[htpb]
\centering
\caption{Flexibility score of implementations under study.}
\begin{tabular}{|p{1cm}||p{5cm}|p{1cm}|}
\hline
Index &Flexibility aspects in papers & Score  \\ \hline 
F2 ~\cite{HW-FPGA-ERNN}&Varying layer (LSTM/GRU)\newline  Varying number of cells\newline Varying block size (block circulant matrices) & \checkmark \checkmark \checkmark \\ \hline

F4~\cite{HW-FPGA-CLSTM}& Varying layer  (LSTM/BiLSTM)\newline Varying number of layers \newline Varying number of cells & \checkmark \checkmark \checkmark\\ \hline

F1~\cite{HW-FPGA-FINN-L-Xilinx} &Varying layer  (LSTM/BiLSTM)\newline Varying precision\newline FC supported  & \checkmark \checkmark \checkmark\\ \hline

F5 ~\cite{HW-FPGA-BiLSTM-OCR}  & Varying layer  (LSTM/BiLSTM)\newline FC supported & \checkmark \checkmark \\ \hline

F7 ~\cite{HW-FPGA-conv-LSTM-short}&  Convolution supported\newline FC supported %newline Pooling supported &  \checkmark 
&\checkmark \checkmark \\ \hline
F8~\cite{HW-FPGA-lowSpeech} &Varying number of layers\newline Varying number of cells \newline Input layer& \checkmark \checkmark \checkmark \\ \hline
F10 ~\cite{HW-FPGA-cong}& Varying number of layers\newline Varying number of cells& \checkmark \checkmark\\ \hline

%F12~\cite{HW-FPGA-baltimore}&Hybrid memory&\checkmark \\ \hline
%A1 ~\cite{HW-ASIC-circular}& Varying number of cells  & \\ \hline

%F13 ~\cite{HW-FPGA-brainwave} &Varying precision \newline CNN supported \newline Varying number of cells \newline Varying layer type & \checkmark \checkmark \checkmark \checkmark \\ \hline

%A2 ~\cite{HW-ASIC-EIE}& CNN supported &\checkmark \\ \hline

A4~\cite{HW-ASIC-ocean}& Online training&\checkmark \\ \hline
A5~\cite{HW-ASIC-DATE} &Varying number of cells\newline FC supported & \checkmark \checkmark \\ \hline
A6~\cite{HW-ASIC-laika}& Varying number of layers\newline Varying number of cells \newline Linear/nonlinear quantization\newline FC supported & \checkmark \checkmark \checkmark \checkmark\\ \hline
A8~\cite{HW-ASIC-PIM-1}&Varying type of layer(LSTM/GRU)\newline Convolution supported\newline FC supported 
&\checkmark \checkmark \checkmark  \\ \hline

A9~\cite{HW-ASIC-MAERI} & Varying number of cells\newline Varying number of layers \newline Dense/Sparse\newline  Convolution supported &\checkmark \checkmark \checkmark \checkmark \\ \hline

A10~\cite{HW-ASIC-fusion} &Varying number of cells\newline Varying number of layers \newline Convolution supported\newline Varying precision& \checkmark \checkmark \checkmark \checkmark \\ \hline

A1~\cite{HW-puma}& Varying number of cells \newline Varying number of layers \newline Varying type of layers \newline Convolution supported \newline FC supported & \checkmark \checkmark \checkmark \checkmark \checkmark\\ \hline

C2~\cite{HW-CPU-stream}& Varying layer  (LSTM/SRU/QRNN)\newline Varying number of cells & \checkmark \checkmark\\ \hline

C3 ~\cite{HW-mobile-GPU-mobirnn}&Varying  number of layers\newline Varying number of cells & \checkmark \checkmark\\ \hline

\end{tabular}
\label{tab:variations}%
\end{table}	

\begin{figure*}[h]
   \centering
   \includegraphics[width=1.25\columnwidth]{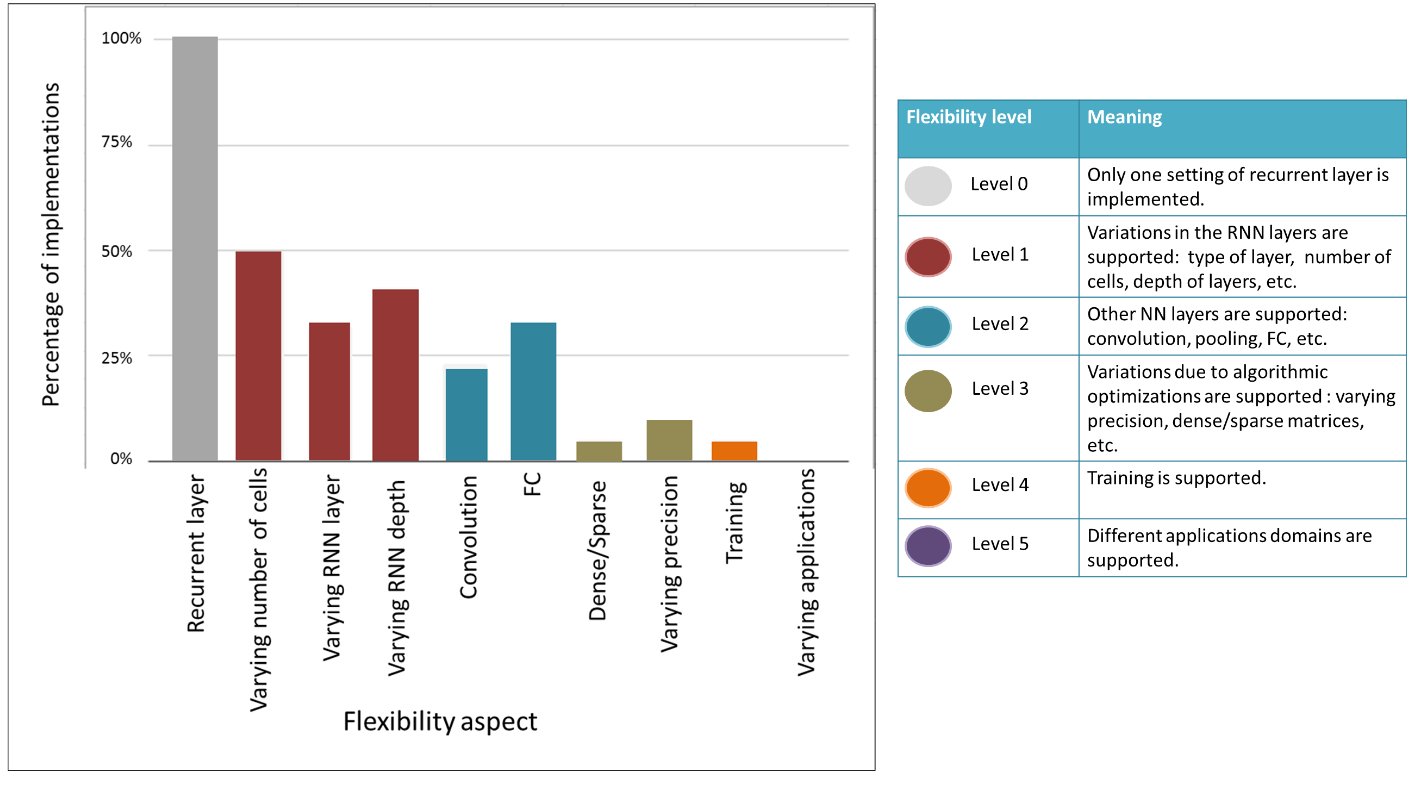}
    \caption{Percentage of implementations meeting flexibility aspects for different flexibility levels and the definition of flexibility levels.}
    \label{fig:levels}
\end{figure*}

To understand how far flexibility is met by the implementations, Figure~\ref{fig:levels} shows the percentage of implementations supporting each flexibility aspect. Flexibility is visualized as levels. Level 0 is used to indicate no flexibility. Level 0 requires the implementation to support only one recurrent layer configuration. All papers meet level 0 requirement but thereafter they vary in meeting other flexibility aspects.
The flexibility aspects and how they can be met are discussed below.
%\begin{itemize}
%\input{tables/flex.tex}

{\textbf{Supporting variations in RNN layers (level 1)}}
Recurrent layers can vary in the type of layers, the number of cells in each layer, and the number of layers (the depth of the RNN model). One optimization that might have a side effect on the flexibility of the solution is the choice of using on-chip or off-chip memory to store the weights. Being able to store all the weights in on-chip memory is very beneficial. It leads to better performance and less energy consumption by decreasing the cost of memory accesses. However, this solution may be unfeasible for larger problems. For example, in F8~\cite{HW-FPGA-lowSpeech}, the number of weights in the model and their precision are restricted by the on-chip memory size. It is not possible to run a model with an increased number of hidden cells or increased precision. A possible solution is to use an adaptable approach, where the location chosen to store the weights is dependent on the model size and thus can a wide range of models can be supported. Another solution was adopted in F12~\cite{HW-FPGA-baltimore}, where some of the weights are stored in internal memory, and the rest are stored in off-chip memory (Hybrid memory).

{\textbf{Supporting other NN layers (level 2)}}  Supporting other NN layers allows the solution to run a broader range of NN applications. Also, other NN layers may exist in the RNN model, such as convolutions used as a feature extractor.
 Supporting such a convolution in the implementation increases the flexibility of the solution, as it can run RNN models with visual inputs and run CNN independent applications. 

{\textbf{Supporting algorithmic
optimization variations (Level 3)}}
Variations in the optimizations applied are also considered as variations in the model. For example, variation due to applying or not applying pruning is related to the presence of sparse or dense matrices in the matrix to vector multiplications computations.
The design in A9~\cite{HW-ASIC-MAERI} employed a configurable interconnection network topology to increase the flexibility of the accelerator. The accelerator in A9~\cite{HW-ASIC-MAERI} supported both LSTM and CNN layers. The accelerators supported both sparse and dense matrices.
One other variation in the precision of the weights and activations. The design in A10~\cite{HW-ASIC-fusion} supported varying precision models by allowing dynamic precision per layer for both CNN and RNN models.
Similarly, the Microsoft NPU brainwave architecture~\cite{HW-FPGA-brainwave} supported varying precision using a narrow precision block floating-point format~\cite{varying-float}. To maximize the benefit of varying precision, F1~\cite{HW-FPGA-FINN-L-Xilinx} applied a parameterizable parallelization scheme. When lower precision is required, LSTM units are duplicated to exploit the unused resources to gain speedup. And, when higher precision is used SIMD folding is applied to save resources for the needed high precision.

{\textbf{Online training (Level 4)}}
Incremented online training was included in A4~\cite{HW-ASIC-ocean} to support retraining pre-trained networks to enhance accuracy. Changes in hardware design were applied to support both training and inference without affecting the quality of inference. For example, three modes of data transfer were applied. The first to load new weights; the second to load input sequences; and the third to update certain weights. Extra precision was only used for training.

{\textbf{Meeting different applications domains constraints (Level 5)}}
None of the implementations target variations in the application domain constraints. NetAdapt is a good example of an implementation that can adapt to different metric budgets~\cite{CNN-netadpat}. However, it only targets CNNs.

%\end{itemize}

%--------------------
\section{Discussions and opportunities}
\label{Sec:Discussion}
In the previous section, we studied the implementations of RNN on embedded platforms. In Section~\ref{Sec:Challanges}, we defined objectives for realizing RNN models on embedded platforms. In this section, we investigate how these objectives are being met by the implementations. 

     {\textbf{Throughput}} It is clear that throughput was the main objective for most of the implementations. As seen in Figure~\ref{fig:throughput}, high throughput was achieved by many of them. Algorithmic and memory optimizations are present in most of these high throughput implementations.
The algorithmic optimizations applied were effective because they decrease both the computation and the memory requirements of the RNN models. For example, if 4-bit precision is used instead of 32-bit for weights storage, the memory requirement is decreased to $1/8$. Multiple 4-bit weights can be concatenated during weights fetching. Thus, the number of memory accesses can decrease as well. Furthermore, the hardware required for 4-bit operations is simpler than the hardware required for 32-bit floating-point operations. %Having less precision would result in more benefits till reaching the extreme benefit when using full binarization as discussed in Section~\ref{subsub:quantize}. %Nevertheless, compression decreases the RNN model size. The size of the RNN model can decrease to 10$\%$ or 5$\%$ of its size. This decrease in the size is accompanied by a decrease in the memory space required, the number of memory accesses, and the number of computations. 

Memory-specific optimizations are effective because they decrease or hide the overhead of accessing the large number of weights used in RNN computations. Memory access time can be decreased by storing all weights in on-chip memory. However, this can bound the validity of the solution for larger models as on-chip memory may not be sufficient to store the weights. Overlapping the memory access time with computation and computation in memory are also considered to be memory optimizations.

{\textbf{Energy efficiency}} 
Applying algorithmic and memory access optimizations has a positive effect on energy efficiency as well. Algorithmic optimizations lead to a decrease in the number of computations, the complexity of computations, and the number of memory accesses, and so decrease the energy consumed by the implementation. Also, minimizing off-chip memory use by storing weights on on-chip memory is an effective way to enhance energy efficiency. Analog computing and processing in memory implementations showed superior energy efficiency in ASIC implementations. 

% HW-ASIC-laika  HW-FPGA-lowSpeech HW-FPGA-flight
{\textbf{Meeting real-time requirements}} was not an objective for many of the implementations. In a few of them, real-time deadlines were mentioned and followed in the design of the solution. 

{\textbf{Flexibilty}}
In Section~\ref{subsec:objectives}, flexibility is defined as a secondary objective. Thus, we do not expect flexibility to be fully met by the implementations. 
%\item{\textbf{Supporting variations in RNN models}} 
Variations in the RNN model was partially fulfilled by many implementations. However, the number of variations covered by each implementation is quite low. 
Few implementations included other NN layers and variations in algorithmic optimizations. Online-training was targeted by only one implementation. Embedded implementations do not usually support online-training. However, on the algorithmic side, researchers are carrying out interesting work based on online or continuous training~\cite{online-tracking,online-finetuning}. None of the RNN implementations support different applications, but this has been done by the CNN solution in~\cite{CNN-netadpat}. Following a similar method in RNNs, and in addition, also supporting model variations, could lead to interesting solutions.

\textbf{Opportunities for future research}

Based on the survey reported on in this article, we list some opportunities for future research.

    \textbf{QRNN and SRU}: 
    QRNN and SRU (Section~\ref{subsub:qrnnsru}) are two alternatives to LSTM where the matrix to vector computations for the current time-step are independent of previous time-step computations. Thus, using them in RNN models can make the parallelization more efficient and consequently lead to better performance.
   
     \textbf{DeltaRNN~\cite{RNN-delta-method} and DeltaCNN~\cite{CNN-Delta}}:
    We believe that applying the delta algorithm to both recurrent and convolution layers is a logical step because of the temporal relation between the input sequences. Adding a delta step to other algorithmic optimizations such as pruning and quantization would decrease memory access and computation requirements.
    
  \textbf{Block-circulant matrices}
    Using block-circulant matrices as an algorithmic optimization decreases the RNN size while avoiding irregularity of computation as introduced by pruning~\cite{HW-FPGA-CLSTM}. Applying circulant matrices can be accompanied by low precision parameters and activations, with a small effect on accuracy~\cite{HW-ASIC-circular}. With the addition of the delta algorithm, as mentioned earlier, RNN inference can achieve a promising throughput and energy efficiency.
    
     \textbf{Hybrid optimizations}:
    It has been shown that a mix of algorithmic optimizations can be applied to an RNN model with a loss in accuracy that is acceptable~\cite{HW-ASIC-circular}.
    Applying a mix of optimizations would enable the implementations to benefit from each optimization. For an RNN implementation, three classes of optimizations can be mixed with tuning. The first optimization is the delta algorithm, and the corresponding parameter is delta. The second is quantization and the corresponding parameters are the number of bits and the quantization method. The third optimization is compression. If the applied compression technique is block-circulant matrices, the parameter is the block size. Tuning the three parameters {delta, {number of bits, quantization method}, and block size}, the designer can achieve the highest performance while keeping the accuracy within an acceptable range (the range is dependent on the application).

     \textbf{Analog computing and processing in memory}: Analog computing~\cite{HW-ASIC-analog} and processing in memory~\cite{HW-ASIC-rram-xnor,HW-puma} have shown promising performance, especially in energy efficiency. Analog crossbar based matrix to vector multiplication units can provide low latency and computing in memory overcomes the memory access problems.

\textbf{Flexible neural networks and domain-specific architectures}
Domain-specific architectures (DSAs) have been highlighted as a future opportunity in the field of computer architecture~\cite{newera_Hennessy19}. DSAs (also called accelerators or custom hardware) for neural networks applications can reach high performance and energy efficiency. Designing an architecture for neural networks applications as a specific domain with known memory access patterns enhances parallelism and the use of the memory hierarchy. It is possible to use lower precision and benefit from domain-specific languages (DSLs). Google Edge TPU is an example of a DSA for neural networks inference using 8-bit precision~\cite{edge-tpu}. Based on the study in this article, we add that the neural networks DSA needs to support flexibility. For the flexibility aspects defined earlier in Section~\ref{Sec:Challanges} to be fulfilled, there are some features need to be supported in the underlying hardware.  
\begin{itemize}
    \item Variable bit-width operations as in A10~\cite{HW-ASIC-fusion} to support different quantization schemes.
    \item Some optimizations require pre/post-processing on input vectors and weights. Support for weights reordering, delta vectors computation, retaining circulant matrices from equivalent vectors, and other operations required by miscellaneous optimizations would be useful.
    \item Support for training that would imply the support of back-propagation and the allowance of weights modification.
    \end{itemize}
%--------------------
\section{Summary}
\label{Sec:Conclusions}
Today we see a trend towards more intelligent mobile devices that are processing applications with streamed data in the form of text, voice, and video. To process these applications, RNNs are important because of their efficiency in processing sequential data. In this article, we have studied the state-of-the-art in RNN implementations from the embedded systems perspective. The article includes all the aspects required for the efficient implementation of an RNN model on embedded platforms. We study the different components of RNN models, with an emphasis on implementation rather than on algorithms. Also, we define the objectives that are required to be met by the hardware solutions for RNN applications, and the challenges that make them difficult to implement. For an RNN model to run efficiently on an embedded platform, some optimizations need to be applied. Thus, we study both algorithmic and platform-specific optimizations. Then, we analyze existing implementations of RNN models on embedded systems.
Finally, we discuss how the objectives defined earlier in the article have been met and highlight possible directions for research in this field in the future.

We conclude from the analysis of the implementations that there are two optimizations that are used for most of the efficient implementations. The first is algorithmic optimizations. The second is to decrease the memory access time for weights retrieval, which can be implemented by relying on on-chip memory for storing the weights, applying an efficient overlap between weights loading and computations, or by computing in memory. However, using analog crossbar based multipliers can achieve high performance without relying too much on algorithmic optimizations. A study of the implementations in the literature shows performance high enough for many streaming applications while also showing a lack of flexibility. Finally, we deduce some opportunities for research to fill the gap between the defined objectives and the research work under study. We highlight some hardware efficient recurrent layers and algorithmic optimizations that can enhance implementations' efficiency. Additionally, we describe how custom embedded hardware implementations can support flexible RNNs solutions.
 
%We concluded from the analysis of the implementations that there exist two mandatory optimizations for efficient implementations. The first is the algorithmic optimizations.% These optimizations should target the matrix to vector multiplications with a significant impact on their computation complexity and their memory requirements. 
%The second is to decrease the memory access time for weights retrieval either by relying on the on-chip memory for storing the weights, applying an efficient overlap between weights loading and computations, or computing in memory.% However, relying on the on-chip memory to store all the weights can have a negative effect on the flexibility of the solution and its feasibility for larger models. Thus, we think that an adaptable solution that can use both on-chip and off-chip memories with different ratios depending on the size of the application can satisfy both implementation efficiency and flexibility. 

% \renewcommand


\begin{IEEEbiography}[{\includegraphics[width=1in,height=1.25in,clip,keepaspectratio]{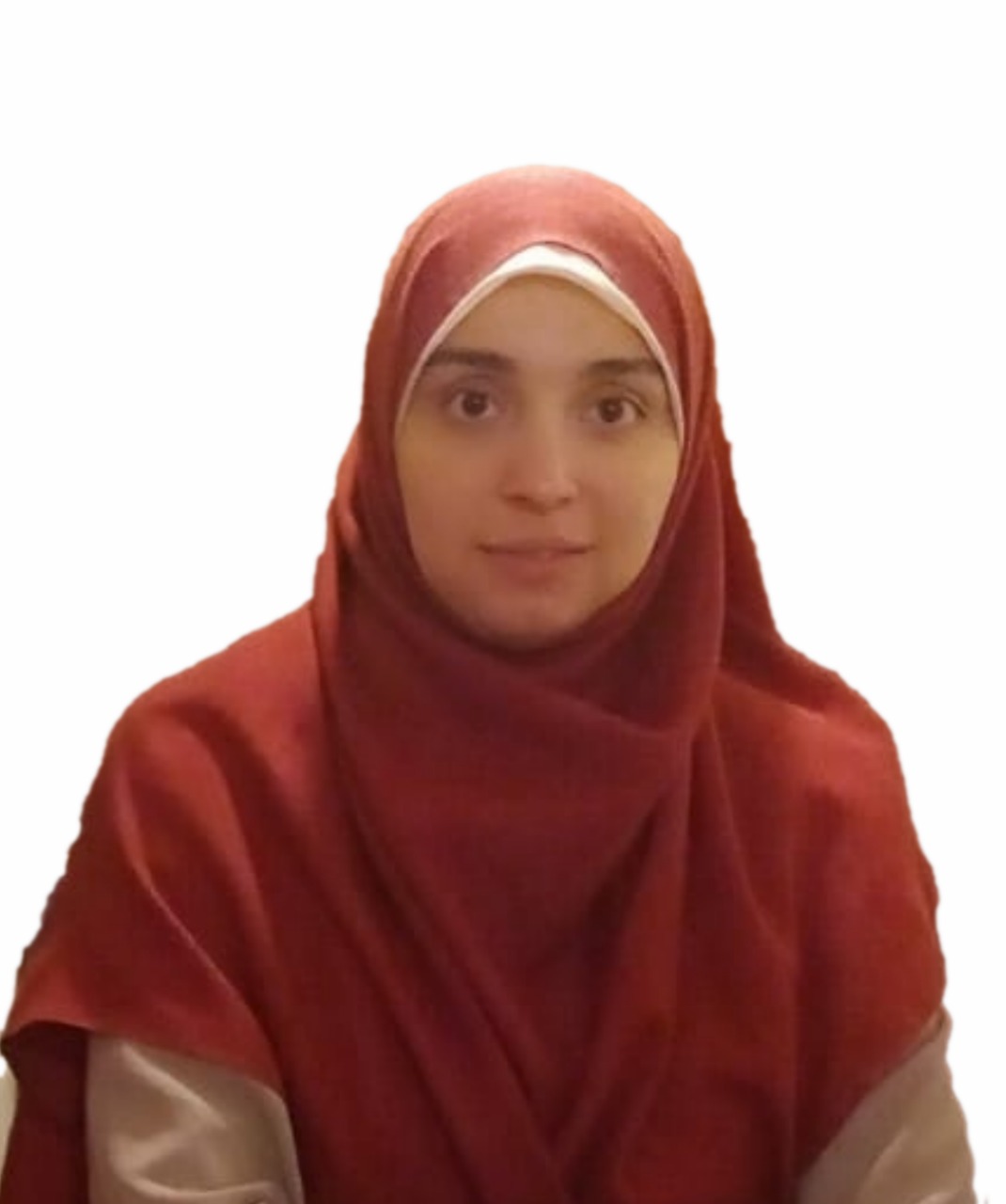}}]{Nesma M. Rezk}received her bachelor and masters degrees in computer and systems engineering from the faculty of engineering, Ain Shams University, Egypt in 2010 and 2015, respectively. 

From 2011 to 2016 she was a teaching and research assistant in the faculty of engineering, Ain Shams University. Since 2016, she has been a PhD student in the School of Information Technology at Halmstad University, Sweden. Her research interests include embedded systems, deep learning applications, and design of domain-specific architectures.  
\end{IEEEbiography}

\begin{IEEEbiography}[{\includegraphics[width=1in,height=1.5in,clip,keepaspectratio]{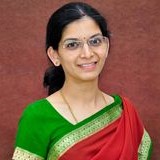}}]{ Madhura Purnaprajna} received her Bachelor's degree in Electronics and Communication Engineering from P.E.S Institute of Technology, Bengaluru in August 1998; her Master's in Electrical and Computer Engineering from University of Alberta, Canada in January 2005; and her PhD in Electrical Engineering from the Heinz Nixdorf Institute, University of Paderborn, Germany in December 2009.  

Madhura Purnaprajna was a post-doctoral fellow with an International Research Fellowship from the German Research Foundation (Deutsche Forschungsgemenischaft) and MHV Fellowship (SNSF), at the Processor Architecture Lab, EPFL, Switzerland and the High-performance Computing Lab, IISc., Bangalore. Currently, she serves as Associate Professor at the Department of Computer Science, School of Engineering, Bengaluru, since February 2015. Her research interests are in Reconfigurable Computing and Processor Architectures.  
\end{IEEEbiography}

\begin{IEEEbiography}[{\includegraphics[width=1.25in,height=1.25in,clip,keepaspectratio]{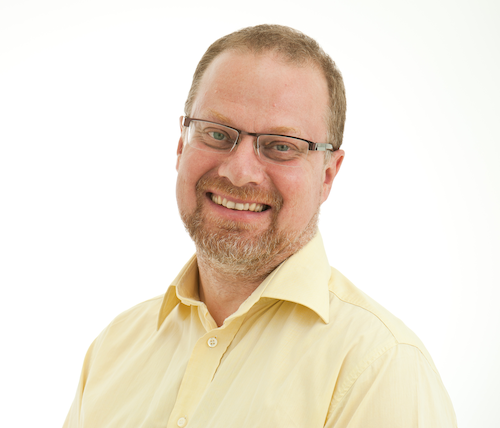}}]{Tomas Nordström} received his M.S.E.E. degree in 1988, his licentiate degree in 1991, and his Ph.D. in 1995, all from Luleå University of Technology, Sweden. His PhD Thesis "Highly Parallel Computers for Artificial Neural Networks" bridged the two fields of computer engineering and signal processing, between which he has been moving ever since.

Between 1996 and 1999, Tomas Nordström was with Telia Research (the research branch of the major Swedish telephone operator) where he developed broadband Internet communication over twisted copper pair. He also became Telia's leading expert in speaker verification during these years. In December 1999, he joined the FTW Telecommunications Research Center Vienna, Austria, where he has been working as a Key Researcher in the field of "broadband wireline access". During his years at FTW, he worked on various aspects of wireline communications such as the simulation of xDSL systems, cable characterization, RFI suppression, exploiting the common-mode signal in xDSL, and more recently, dynamic spectrum management. In 2009 was appointed Associate Professor in computer systems engineering at Halmstad University (HH), Sweden. At HH he has returned to the area of computer architecture and his current research interests include all aspects of energy-efficient embedded computers. In 2012 he became full Professor in Computer Engineering at HH and has built up a research group focusing on heterogeneous many-core architectures. Additionally, he has been working in the field of dependable wireless communication studying optimal usage of communication resources, dynamic spectrum management, and IoT reference architectures. In 2019 he became full Professor in Embedded and Intelligent Systems at Umeå University, Sweden, where his research focuses on edge computing, intelligent IoT systems, and high-performance embedded computing architectures and platforms.
 
\end{IEEEbiography}

\begin{IEEEbiography}[{\includegraphics[width=1in,height=1.5in,clip,keepaspectratio]{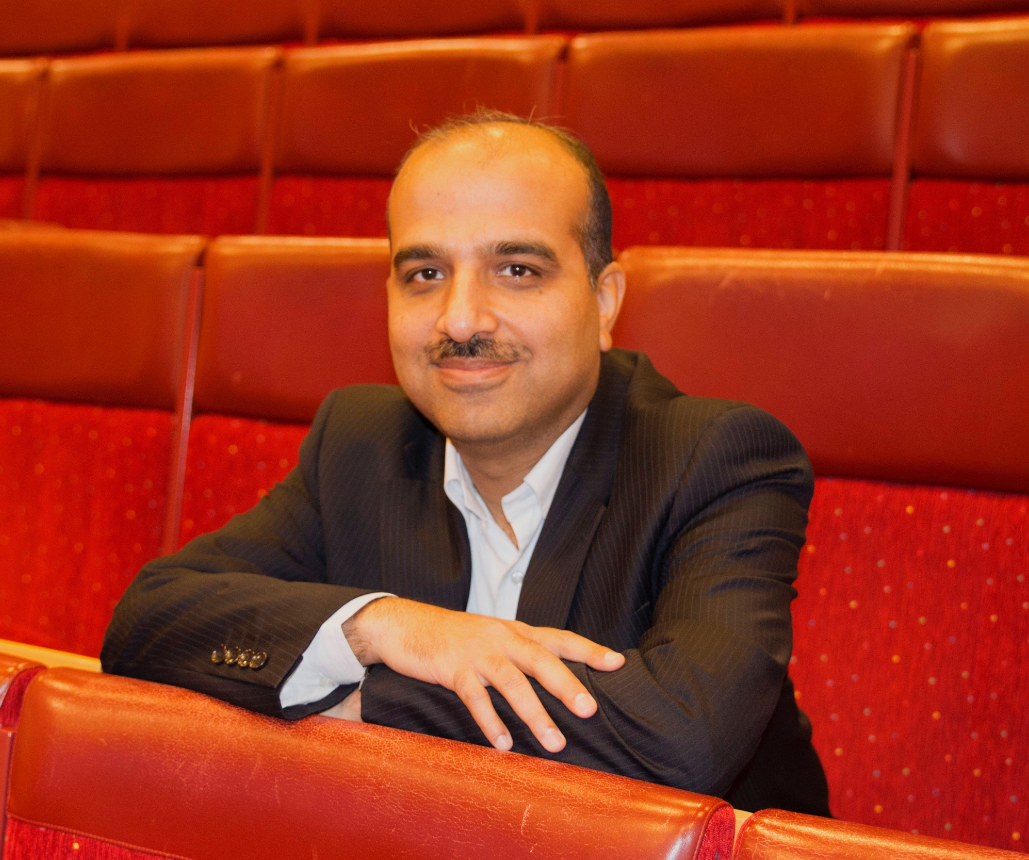}}]{ Zain-ul-Abdin} completed his PhD degree in Computer Science from Örebro University in 2011 and until recently held an appointment as Associate Professor at Halmstad University. 

He has played a key role in a number of research projects such as Smart Multicore Embedded Systems (SMECY), High-Performance Embedded Computing (HiPEC), and Towards Next Generation Embedded Systems (NGES). In these projects he has worked in close collaboration with the Swedish Defence Research Agency (FOI) and industrial partners such as ST- Microelectronics, Ericsson, Adapteva, and Saab. He has authored more than 42 journal and conference articles and has been awarded the best paper prize in the PhD forum of 19th Field-programmable Logic and Applications conference (FPL'09). He has sat on the technical program committee of several leading conferences and has served as an external reviewer for journals (IJRC, IEEE-TSP, IEEE-Micro, JSP). His main research interests are high-performance embedded computing and the parallel programming models. 
\end{IEEEbiography}

\EOD

\end{document}